\documentclass[acmsmall]{acmart}
\usepackage{xcolor}
\usepackage[normalem]{ulem}
\usepackage{booktabs}
\usepackage{tabularx}
\usepackage{array}
\usepackage{ragged2e}

\acmJournal{TIST}
\acmYear{2026}
\acmMonth{6}
\received[Accepted]{June 2026}

\makeatletter
\AtBeginDocument{%
  \fancypagestyle{firstpagestyle}{%
    \fancyhf{}%
    \renewcommand{\headrulewidth}{\z@}%
    \renewcommand{\footrulewidth}{\z@}%
    \fancyfoot[RO,LE]{\footnotesize ACM Transactions on Intelligent Systems and Technology. Accepted: June 2026.}%
    \fancyhead[LE]{\ACM@linecountL\@acmBadgeL}%
    \fancyhead[LO]{\ACM@linecountL\@acmBadgeL}%
    \fancyhead[RO]{\@acmBadgeR}%
    \fancyhead[RE]{\@acmBadgeR}%
  }%
  \fancypagestyle{standardpagestyle}{%
    \fancyhf{}%
    \renewcommand{\headrulewidth}{\z@}%
    \renewcommand{\footrulewidth}{\z@}%
    \fancyhead[LE]{\ACM@linecountL\@headfootfont\@acmArticlePage}%
    \fancyhead[RO]{\@headfootfont\@acmArticlePage}%
    \fancyhead[RE]{\@headfootfont\@shortauthors}%
    \fancyhead[LO]{\ACM@linecountL\@headfootfont\shorttitle}%
    \fancyfoot[RO,LE]{\footnotesize ACM Transactions on Intelligent Systems and Technology. Accepted: June 2026.}%
  }%
  \pagestyle{standardpagestyle}%
}
\makeatother

\title{A Technical Survey of Reinforcement Learning Techniques for Large Language Models}

\author{Saksham Sahai Srivastava}
\affiliation{
  \institution{University of Georgia}
  \city{Athens}
  \state{GA}
  \country{USA}
}
\email{saksham.srivastava@uga.edu}

\author{Vaneet Aggarwal}
\affiliation{
  \institution{Purdue University}
  \city{West Lafayette}
  \state{IN}
  \country{USA}
}
\email{vaneet@purdue.edu}



\usepackage[normalem]{ulem}

\renewcommand{\sout}[1]{}

\begin{document}

\begin{abstract}
Reinforcement Learning (RL) has emerged as a transformative approach for aligning and enhancing Large Language Models (LLMs), addressing critical challenges in instruction following, ethical alignment, and reasoning capabilities. This survey offers a comprehensive foundation on the integration of RL with language models, highlighting prominent algorithms such as Proximal Policy Optimization (PPO), Q-Learning, and Actor-Critic methods. Additionally, it provides an extensive technical overview of RL techniques specifically tailored for LLMs, including foundational methods like Reinforcement Learning from Human Feedback (RLHF) and AI Feedback (RLAIF), as well as advanced strategies such as Direct Preference Optimization (DPO) and Group Relative Policy Optimization (GRPO). We systematically analyze their applications across domains, i.e., from code generation to tool-augmented reasoning. Crucially, we move beyond descriptive categorization to provide a rigorous algorithmic analysis of failure modes, mathematically framing the structural bottlenecks and stability trade-offs inherent in policy optimization. We also present a comparative taxonomy based on reward modeling, feedback mechanisms, and optimization strategies. Our evaluation highlights key trends. RLHF remains dominant for alignment, and outcome-based RL such as Reinforcement Learning with Verifiable Rewards (RLVR) significantly improves stepwise reasoning. However, persistent challenges such as reward hacking, computational costs, and scalable feedback collection underscore the need for continued innovation. We also explicate the causal factors behind recent benchmark performances, distinguishing between gains derived from architectural scaling versus those stemming from specific optimization objectives. We further discuss emerging directions, including hybrid RL algorithms, verifier-guided training, and multi-objective alignment frameworks. This survey serves as a roadmap for researchers advancing RL-driven LLM development, balancing capability enhancement with safety and scalability.
\end{abstract}

\maketitle

\section{Introduction}
Large Language Models (LLMs) have emerged as transformative technologies in artificial intelligence, demonstrating remarkable capabilities in understanding and generating human language. From GPT-3's 175 billion parameters  \cite{brown2020language} to more recent architectures like LLaMA 3.1 with 405 billion parameters \cite{grattafiori2024llama} and DeepSeek-V3 with 671 billion parameters \cite{liu2024deepseek}, these models have progressively expanded in scale and capability. While scaling laws have consistently yielded improvements in perplexity and knowledge retention, they do not inherently resolve the \textit{alignment problem}, which is to ensure their outputs consistently reflect human values, preferences, and intention. The fundamental objective of pre-training, which is minimizing the negative log-likelihood of the next token, prioritizes statistical probability over factual correctness or ethical safety. Consequently, despite their impressive performance across various tasks, LLMs still struggle with alignment. Therefore, these models often struggle with hallucinations  \cite{xu2024hallucination}, exhibit vulnerability to generating harmful content  \cite{bianchi2024large, ge2025llms, yi2024vulnerability}, and frequently fail to follow complex instructions precisely  \cite{murthy2024evaluating}.

To bridge the gap between statistical mimicry and goal-directed behavior, Reinforcement Learning (RL) has emerged as the critical methodological framework. Unlike the static nature of supervised learning, RL formulates language generation as a sequential decision-making process, where an agent improves its policy through trial and error by interacting with a responsive environment. This paradigm offers robust mechanisms to incorporate non-differentiable feedback signals and optimizes for complex, multi-faceted objectives that are difficult to encode in a standard loss function. The integration of RL with LLMs represents a significant advancement in AI alignment research, enabling models to learn from human preferences, improve reasoning capabilities, and better adhere to ethical guidelines. This technical survey provides a comprehensive examination of RL techniques applied to LLMs, focusing on both alignment with human values and enhancement of reasoning capabilities.

Implementing this framework necessitates moving beyond the constraints of standard supervised learning (SL). \sout{While SL minimizes the token-level cross-entropy loss, it effectively forces the model to mimic the training distribution. So, it fails to capture holistic outcome metrics such as coherence, factual reliability, and safety.} \textcolor{black}{While SL minimizes token-level cross-entropy and is highly effective for imitation, this objective is only an indirect proxy for sequence-level or preference-defined qualities. Prior work~ \cite{ranzato2015sequence} on sequence-level training has shown that word-level likelihood can be mismatched with test-time sequence metrics, while human-feedback studies~ \cite{stiennon2020learning, ouyang2022training} show that models trained only to predict demonstrations or reference text may still produce outputs that users judge as unhelpful, untruthful, or unsafe. Thus, the limitation is not that SL cannot learn coherence, factual reliability, or safety, but that it does not directly optimize these holistic response-level criteria unless they are sufficiently represented in the supervised data. RL bridges this gap by allowing such qualities to be encoded through reward models, preference comparisons, or verifiable outcome signals.} \sout{RL bridges this gap by treating the LLM as a policy $\pi$ that interacts with a textual environment. However, this application is non-trivial.} \textcolor{black}{In this formulation, the LLM is treated as a policy $\pi$ that interacts with a textual environment, but adapting RL to language generation is non-trivial.} The state space (context window) is high-dimensional, and the action space (vocabulary size) is discrete and vast, rendering standard gradient estimators like the reparameterization trick inapplicable. Furthermore, the ground truth in alignment tasks is rarely a single correct token, but rather a spectrum of human preferences, necessitating reward modeling that introduces its own approximation errors. Consequently, the field has evolved from naively applying policy gradients to developing specialized mechanisms like Proximal Policy Optimization (PPO) \cite{schulman2017proximal} adapted for KL-constrained language generation.

Building upon these theoretical adaptations, Reinforcement Learning from Human Feedback (RLHF)  \cite{ouyang2022training} has materialized as the de facto standard implementation for aligning LLMs with human preferences. By operationalizing the reward signal through human judgment, RLHF transforms the abstract goal of alignment into a tractable optimization problem. This approach typically follows a three-stage process: supervised fine-tuning on high-quality demonstrations, training a reward model from human preference data, and optimizing the policy using algorithms like PPO  \cite{schulman2017proximal}. RLHF has demonstrated remarkable effectiveness in improving instruction-following capabilities and reducing harmful outputs, as evidenced by OpenAI's InstructGPT  \cite{ouyang2022training}.

However, the scalability limitations of human annotation have motivated the development of alternative approaches. Reinforcement Learning from AI Feedback (RLAIF)  \cite{lee2023rlaif} replaces or augments human feedback with evaluations from other AI systems, significantly reducing annotation costs while maintaining comparable performance. Constitutional AI  \cite{bai2022constitutional} represents a specialized form of RLAIF where models critique and revise their own outputs based on predefined principles, particularly effective for harmlessness alignment. While RLAIF addresses the data scarcity bottleneck, it retains the computational complexity of the standard actor-critic pipeline. To address the stability issues inherent in training separate reward and policy models, more recent innovations have focused on algorithmic simplification. Direct Preference Optimization (DPO)  \cite{rafailov2023direct} fundamentally restructures the problem by bypassing explicit reward modeling and directly optimizing the policy using preference pairs, offering improved computational efficiency and training stability. Empirical evaluations have shown that DPO can match or exceed the performance of PPO \cite{schulman2017proximal}-based RLHF on tasks like sentiment control and summarization with substantially reduced complexity.

Beyond the alignment of stylistic and ethical preferences, RL techniques have increasingly been applied to enhance the latent reasoning capabilities of LLMs. \sout{In domains requiring multi-step logic, such as mathematics or coding, supervised learning often suffers from exposure bias.} \textcolor{black}{In domains requiring multi-step logic, such as mathematics or coding, supervised learning can suffer from exposure bias, because training conditions on gold prefixes while inference conditions on the model's own previously generated tokens~ \cite{ranzato2015sequence}.} \sout{RL, by contrast, incentivizes the model to explore diverse chains of thought to reach a correct solution.} \textcolor{black}{By contrast, RL-based post-training can sample multiple candidate reasoning trajectories and reinforce those that receive higher outcome-level, process-level, or verifier-derived rewards. This makes it possible to optimize reasoning behavior at the level of complete solutions rather than only at the level of next-token imitation.} Outcome-Based Reinforcement Learning approaches \cite{lyu2025exploring} reward models for generating correct final answers, even when intermediate reasoning steps are not explicitly supervised. \sout{More sophisticated methods like Reinforcement Learning with Verifiable Rewards (RLVR)  \cite{lambert2025tulu} provide step-wise feedback on reasoning processes, significantly improving performance on mathematical and logical reasoning tasks.} \textcolor{black}{Methods such as Reinforcement Learning with Verifiable Rewards (RLVR)~ \cite{lambert2025tulu} instead use task-specific verifiers to provide objective reward signals, often at the level of the final answer or executable solution, thereby improving mathematical and logical reasoning when correctness can be programmatically checked.} Despite these advances, significant challenges persist in applying RL to LLMs. One such challenge is reward hacking  \cite{denison2024sycophancy, fu2025reward}, where models exploit loopholes in the reward functions instead of genuinely improving their behavior. The computational costs associated with RL training, particularly for models with billions of parameters, also present practical limitations for widespread adoption. Additionally, ensuring the quality and representativeness of feedback \cite{yeh2024reliable, sharma2024critical}, whether from humans or AI systems, continues to be a complex problem.

This survey makes several key contributions to the field. First, we provide a comprehensive technical overview of RL techniques applied to LLMs. We cover foundational methods such as RLHF and RLAIF, as well as advanced approaches like DPO, Group Relative Policy Optimization (GRPO), RLVR, and Unified Alignment (UNA). Unlike prior reviews, we incorporate dedicated Algorithmic Analysis and Critical Analysis modules for each method, providing mathematical explanations for why certain approaches succeed or fail. Second, we systematically analyze applications across various domains. These domains include code generation and tool-augmented reasoning, demonstrating RL's versatility and effectiveness. Third, we present a comparative taxonomy based on reward modeling strategies, feedback mechanisms, and optimization approaches. This taxonomy provides a structured framework to understand the landscape of RL techniques for LLMs. Finally, we identify emerging research directions such as hybrid RL algorithms, verifier-guided training, and multi-objective alignment frameworks.

The remainder of this paper is organized as follows: Section 2 establishes the foundational concepts of LLMs and RL; Section 3 details specific RL algorithms adapted for LLMs; Section 4 explores RL techniques for alignment and reasoning enhancement; Section 5 presents applications across various domains; Section 6 provides a comparative analysis and evaluation; Section 7 discusses challenges and limitations; Section 8 talks about future research directions; Section 9 presents the conclusion of this survey; and Section 10 offers the authors' perspective on this field. We aim to provide researchers and practitioners with a roadmap for advancing RL-driven LLM development through this comprehensive examination. This roadmap seeks to balance capability enhancement with safety and scalability considerations.

\section{Background \& Foundations}
To provide the necessary theoretical scaffolding, this section first delineates the architecture and pre-training objectives of modern LLMs, identifying the inherent limitations that necessitate further alignment. We then review the formal mechanisms of Reinforcement Learning, focusing on the Markov Decision Process (MDP) and the prominent algorithms used to optimize policies. Finally, we synthesize these concepts to examine the structural adaptations required when treating an LLM as an RL agent, laying the groundwork for analyzing the alignment-optimization Gap that defines current research.

\subsection{Large Language Models}
Contemporary LLMs function as high-capacity probabilistic engines, leveraging the Transformer architecture  \cite{vaswani2017attention} to model the conditional probability distribution of text. They are trained on vast corpora of text data. While their primary utility lies in natural language understanding and generation, from a reinforcement learning perspective, they are best conceptualized as stochastic policies parameterized by weights $\theta$, mapping context states to categorical distributions over a vocabulary.

\subsubsection{Architecture and Training}
Modern large language models overwhelmingly employ the decoder‑only branch of the Transformer introduced by Vaswani et al. \cite{vaswani2017attention}, stacking self‑attention blocks to generate text autoregressively, token by token. This design exploits multi‑head self‑attention to process all positions in parallel, giving the models the capacity to capture long‑range dependencies far more efficiently than recurrent networks. The training of LLMs is framed as maximum-likelihood next-token prediction. Formally, given a sequence of tokens $x = (x_1, \dots, x_T)$, the model maximizes the likelihood of the joint distribution factored by the chain rule:
\begin{equation*}
P_\theta(x) = \prod_{t=1}^{T} P_\theta(x_t \mid x_{<t})
\end{equation*}
Therefore, today’s LLMs essentially represent large conditional probability models that forecast the most likely subsequent token given the context. To maximize the expressive power of these probability distributions, the field has pursued an aggressive trajectory of model scaling. The scaling of models is governed by empirical scaling laws which suggest that increased capacity correlates with improved generalization. This evolutionary trajectory is evident in the rapid progression from OpenAI’s GPT-2  \cite{radford2019language}, which initially featured 1.5 billion parameters, to the significant leap represented by GPT-3  \cite{brown2020language} with 175 billion parameters. The pursuit of scale continued with massive dense architectures like Google's PaLM (540 billion parameters) and Meta’s LLaMA 3.1  \cite{grattafiori2024llama} (405 billion parameters), culminating in sophisticated Mixture-of-Experts models such as DeepSeek-V3  \cite{liu2024deepseek}, which contains 671 billion parameters.  However, while scaling laws have driven parameter counts from the billions to the hundreds of billions, the fundamental training objective, i.e. minimizing the negative log-likelihood of the next token, remains constant across these generations.

To effectively optimize this objective across such vast parameter spaces, the training pipeline is conventionally structured into a two-phase approach. The first phase is Pre-training. In this phase, the model is trained on a diverse corpus of text using self-supervised learning objectives, such as predicting masked tokens or next-token prediction. This phase enables the model to learn general language patterns, world knowledge, and reasoning capabilities. The second phase is supervised  fine-tuning. In this phase, the pre-trained model is further trained on specific datasets to adapt it for particular tasks or to align it with human preferences.

\subsubsection{Capabilities and Limitations}

As a direct result of scaling and extensive pre-training, pre-trained LLMs demonstrate impressive capabilities across various tasks, including text completion, summarization, translation, question answering, and even complex reasoning. However, because the pre-training objective prioritizes statistical mimicry over factual verifiability, these models still suffer from notable limitations. One critical issue is hallucinations  \cite{xu2024hallucination}, where LLMs generate plausible yet factually incorrect information, leading to potentially misleading outputs. Additionally, without proper safeguards, these models can produce harmful, biased, or toxic content \cite{gehman-etal-2020-realtoxicityprompts}, reflecting and amplifying societal biases \cite{bender2021dangers} present in their uncurated training data. LLMs \cite{murthy2024evaluating} also frequently struggle to precisely follow complex or multi-step user instructions, limiting their practical utility in structured tasks. This deficit stems from the objective mismatch where the model is optimized to predict the next token in a static document rather than to satisfy a dynamic user intent. Consequently, standard models often struggle to align effectively with human values, preferences, or intentions, underscoring the necessity of specialized alignment techniques. Although inference-time prompting-based methods such as MathPrompter \cite{imani2023mathprompter} and MathDivide \cite{srivastava2024mathdivide} have enhanced mathematical reasoning within narrow domains by employing strategies like multi-path validation and problem decomposition, they remain constrained by their dependency on fixed prompting heuristics and their inability to generalize or incorporate broader interactive feedback. These limitations underscore the necessity of reinforcement learning, which shifts the optimization goal from \textit{predicting} the next token to \textit{maximizing} a cumulative value, thereby enabling the model to internalize complex objectives like helpfulness, harmlessness, and robust logical reasoning.

\subsection{Reinforcement Learning Fundamentals}
To formalize the optimization of cumulative objectives beyond simple next-token prediction, we turn to RL, a machine learning paradigm in which an agent learns optimal decision-making by interacting with an environment and receiving feedback through rewards or penalties. The RL framework is typically formalized as a MDP, denoted as a tuple $(S, A, P, R, \gamma)$. This structure is defined by states ($S$) representing possible situations, actions ($A$) indicating available decisions, a transition function ($P$) specifying probabilities of moving between states given actions, and a reward function ($R$) providing immediate feedback on actions. Central to RL are policies ($\pi$) which are strategies mapping states to actions, and value functions ($V$ or $Q$), estimating expected cumulative future rewards. A discount factor ($\gamma$) balances immediate versus future rewards. Ultimately, the objective in RL is to discover the optimal policy ($\pi^*$) that maximizes long-term expected cumulative rewards.

To solve these MDPs, researchers have developed distinct algorithmic families, each navigating the trade-off between sample efficiency and training stability differently. Value-based methods, such as Q-learning \cite{watkins1992q} and Deep Q-Networks (DQN) \cite{mnih2015human}, focus on estimating value functions to determine the expected cumulative reward from each state or state-action pair. While sample-efficient, these methods often struggle in high-dimensional action spaces. In contrast, policy gradient methods, including REINFORCE \cite{williams1992simple} and PPO  \cite{schulman2017proximal}, directly optimize the policy by computing gradients of expected returns with respect to policy parameters. Actor-critic methods, such as Advantage Actor-Critic (A2C) \cite{mnih2016asynchronous} and Soft Actor-Critic (SAC) \cite{haarnoja2018soft}, combine these approaches by simultaneously learning a value function (critic) and a policy (actor) to reduce the variance of the gradient estimate.

Despite their effectiveness, RL algorithms face inherent challenges that complicate their practical deployment. The exploration-exploitation tradeoff \cite{auer2002finite} requires balancing the need to discover new beneficial actions with exploiting known successful behaviors. Credit assignment poses difficulties in attributing rewards accurately to specific actions within sequences, especially when outcomes are delayed. Sample efficiency is another critical concern, as RL algorithms often require extensive interactions with the environment to learn effectively. It is a significant bottleneck when the environment is a computationally expensive LLM. Finally, ensuring stability during learning, particularly when employing complex function approximations like neural networks, is essential to prevent divergence and maintain reliable performance.

\subsection{Intersection of RL and LLMs}
The application of RL to LLMs fundamentally redefines the training objective from mimicking static text to optimizing for dynamic, outcome-based utility. While pre-training on vast corpora instills general linguistic competence, it does not inherently align the model with nuanced human intent or complex reasoning requirements. RL bridges this gap by treating the LLM as an agent $\pi_\theta$ that interacts with a textual environment. In this formulation, the environment state $s_t$ comprises the input prompt and the sequence of tokens generated up to step $t$, while the action $a_t$ is the selection of the next token from the model's vocabulary. Accordingly, the language model itself functions as the stochastic policy $\pi_\theta(a_t|s_t)$, and the entire generated response is viewed as a trajectory of state-action pairs. This intersection allows models to optimize for non-differentiable objectives that are difficult to capture with standard cross-entropy loss. For instance, subjective qualities like helpfulness or humor cannot be easily defined by a labeled dataset but can be captured through preference-based rewards derived from human judgments  \cite{ouyang2022training}. Similarly, in reasoning tasks, RL enables the optimization of entire chains of thought, rewarding the model not just for the final answer but for the logical validity of the intermediate steps  \cite{wang2025reinforcement}. However, translating this theoretical alignment into practice is non-trivial, as adapting standard RL frameworks to the LLM context requires addressing specific structural constraints. The action space is discrete and massive (often exceeding 50,000 tokens), rendering many continuous-control algorithms inapplicable. To manage this, policy gradients are typically estimated using the log-probabilities of the generated tokens. Furthermore, because the ground truth for alignment is often a distribution of preferences rather than a single correct output, rewards are frequently mediated by a learned Reward Model (RM) or external verifiers (e.g., code compilers). Stability is maintained via KL-divergence \cite{kullback1951information} penalties, which constrain the RL policy to stay within the trust region of the original supervised baseline, thereby preventing the model from degenerating into incoherent gibberish to exploit flaws in the reward function. However, successfully implementing these adaptations requires more than simply mapping RL terms to LLM components. It necessitates navigating a profound friction between the probabilistic nature of language generation and the goal-directed nature of reinforcement learning.

\subsection{The Alignment-Optimization Gap}
We characterize this friction as the \textit{Alignment-Optimization Gap}. It represents the fundamental structural mismatch between the token-level objectives used during pre-training and the sequence-level objectives required for alignment. While standard RL algorithms typically assume a clear separation between the environment dynamics and the agent's policy, in language modeling, the policy effectively dictates the transition dynamics of the environment itself. Understanding this gap is crucial, as it dictates the design choices behind modern algorithms like PPO and DPO, which must bridge these disparate optimization landscapes. The gap manifests primarily in three dimensions:

\begin{itemize}
    \item \textbf{Token-Level vs. Sequence-Level Objectives:} The core limitation of Next-Token Prediction (NTP) is its greedy, myopic nature. NTP minimizes the cross-entropy loss for each token independently, assuming that the optimal next token depends solely on the immediate context. In contrast, RL optimizes the expected return of the entire trajectory (sequence). \sout{Mathematically, this shifts the goal from maximizing the immediate likelihood $P(x_t|x_{<t})$ to maximizing the long-term value function $V(s_t)$, often requiring the model to traverse valleys of low probability to reach peaks of high utility. This shift is crucial for tasks requiring planning.} \textcolor{black}{Mathematically, this shifts the objective from maximizing the immediate likelihood \(P(x_t \mid x_{<t})\) of each reference token to increasing the probability of complete trajectories that receive high sequence-level reward. In LLM post-training, this should not be interpreted as unconstrained exploration in the robotics sense. Instead, exploration~ \cite{ouyang2022training,schulman2017proximal,lyu2025exploring,guo2025deepseek} is usually induced by sampling multiple candidate completions from the current policy, scoring them with preference models, verifiers, or outcome rewards, and then using KL-constrained policy updates to shift probability mass toward higher-reward trajectories while preserving the linguistic support of the pretrained model.} This allows the model to select tokens that may have lower immediate probability but lead to a higher long-term reward, effectively enabling the model to think ahead or delay gratification to construct a more coherent argument.

    \item \textbf{Sparse vs. Dense Signals:} In supervised learning, every token in the training set provides a gradient signal, creating a dense and stable feedback loop. In RL, feedback is often sparse and delayed and is provided only at the end of a generated sequence (e.g., a binary success/failure signal in code generation or a scalar preference score). This exacerbates the credit assignment problem. Determining which specific tokens in a long sequence contributed to the final reward becomes computationally difficult. Without explicit mitigation, this sparsity leads to high-variance gradient estimates. To bridge this, modern approaches often employ learned Value Functions ($V(s)$) or dense process-supervision rewards (e.g., CoT-RO  \cite{lightman2023let}) to provide more granular feedback at intermediate steps.


    \item \textbf{The Exploration-Exploitation Dilemma:} In classical NTP, exploration is often just stochastic sampling, such as temperature-based decoding. \textcolor{black}{Operationally, in RL-based LLM post-training, exploration is usually implemented through temperature or nucleus sampling during rollout generation, best-of-\(N\) or group-based candidate sampling, entropy-regularized objectives, and verifier- or reward-model-based filtering rather than through arbitrary random actions. These mechanisms allow the model to test multiple plausible continuations while remaining close to the pretrained distribution.} \sout{In the RL context, exploration must be semantic and meaningful. Unlike standard RL agents in games that can attempt random physical actions to discover new strategies, an LLM must explore the \textit{semantic} space of ideas while strictly adhering to the \textit{syntactic} space of grammar.} \textcolor{black}{Unlike standard RL agents in games that can attempt random physical actions, an LLM must explore the \textit{semantic} space of ideas while remaining within the \textit{syntactic} and linguistic constraints learned during pre-training.} The challenge lies in encouraging the model to explore diverse reasoning paths or phrasing styles without devolving into incoherence. If the exploration is too narrow, the model collapses into a local optimum or mode collapse. If it is too broad, it violates the linguistic constraints imposed by the pretrained weights. Techniques like entropy regularization and PPO's clipped objective are critical for balancing this trade-off. They help the model explore higher-reward regions without forgetting the syntax and semantics learned during pre-training.
\end{itemize}

\section{Reinforcement Learning Algorithms for LLMs}

The adaptation of RL algorithms for Language Models has been driven by a tension between sample efficiency, training stability, and computational constraints. While standard algorithms from robotics (like TRPO \cite{schulman2015trust} or SAC \cite{haarnoja2018soft}) exist, LLM-specific adaptations primarily focus on managing the KL divergence to prevent the hacking of reward models and preserving natural language fluency. This section reviews prominent RL algorithms specifically tailored for aligning LLMs with human preferences and enhancing their reasoning capabilities. We survey the technical evolution of the field, tracing a trajectory from stability-focused on-policy methods to efficiency-driven off-policy frameworks, and finally to recent relative-optimization architectures that attempt to reconcile these competing objectives without the computational overhead of value function approximation. We first examine Proximal Policy Optimization \cite{schulman2017proximal} (PPO), highlighting its stability and wide adoption in RLHF frameworks. Next, we explore Q-learning and other Off-Policy RL methods, such as Implicit Language Q-Learning (ILQL) \cite{snell2206offline} and VerifierQ \cite{qi2024verifierq}, focusing on their effectiveness in leveraging offline datasets and verifier-based reasoning enhancement. Finally, we discuss advanced methodologies like GRPO  \cite{guo2025deepseek}, which improves training efficiency and robustness by adopting relative advantage estimation within grouped candidate responses.

\subsection{Proximal Policy Optimization:} As the cornerstone of modern alignment strategies, PPO prioritizes training stability in the high-dimensional action spaces inherent to LLMs. Rather than aggressively optimizing for sample efficiency, PPO has emerged as the standard for RLHF by enforcing policy constraints that prevent the model from deviating destructively from its pre-trained knowledge. Notably, OpenAI \cite{schulman2017proximal} popularized the use of PPO through their RLHF implementation in the InstructGPT model \cite{ouyang2022training}. Here, PPO was employed to align the model with human instructions. Within the RLHF framework, PPO iteratively updates the language model policy by maximizing rewards provided by a learned reward model. It also simultaneously constrains policy changes relative to a reference model to maintain stable updates. Zheng et al. \cite{zheng2023secrets} highlighted that PPO works well because it uses policy-constraint mechanisms like clipped probability ratio updates or KL divergence penalties. These constraints keep the updates stable and reliable during training. They also introduced a new variant called PPO-Max, which makes the training process even more stable. PPO's balanced optimization properties have established it as the predominant method for training aligned LLM policies using RLHF. PPO's stability is mechanically enforced through the clipped surrogate objective:

\[
\mathcal{L}_{\text{PPO}}(\theta) = \mathbb{E}_t \left[ \min \left( r_t(\theta) \hat{A}_t, \; \text{clip}(r_t(\theta), 1 - \epsilon, 1 + \epsilon) \hat{A}_t \right) \right]
\label{eq:ppo_loss}
\]

\noindent where $r_t(\theta) = \frac{\pi_\theta(a_t \mid s_t)}{\pi_{\theta_{\text{old}}}(a_t \mid s_t)}$ (probability ratio), $\hat{A}_t =$ estimated advantage at time step $t$, $\epsilon =$ clip parameter (typically 0.1 or 0.2), $\pi_\theta =$new policy, $\pi_{\theta_{\text{old}}} =$ policy before the update.

While PPO ensures stability, it is inherently sample-inefficient, requiring new rollouts from the policy for every update step. In the context of LLMs, where a single forward pass is computationally expensive, this creates a significant bottleneck. Consequently, research has pivoted toward off-policy methods that can leverage static, pre-collected datasets (replay buffers) to decouple data generation from policy optimization.

\textbf{Algorithmic Analysis:} The operational efficacy of PPO is defined by a critical trade-off between optimization stability and resource intensity. While the clipped surrogate objective prevents destructive policy updates, it introduces a severe computational overhead often termed the \textit{Four-Model Bottleneck}.  Mathematically, the active memory footprint scales as $\mathcal{M}_{\text{total}} \approx \mathcal{M}_{\pi} + \mathcal{M}_{\text{ref}} + \mathcal{M}_{r} + \mathcal{M}_{V}$, where $\mathcal{M}_{\pi}$, $\mathcal{M}_{\text{ref}}$, $\mathcal{M}_{r}$, and $\mathcal{M}_{V}$ represent the memory requirements for the Policy, Reference, Reward, and Value models respectively. Since the Value Function ($V_\psi$) and Reference Model ($\pi_{\text{ref}}$) typically share the same parameter count $N$ as the Policy ($\pi_\theta$), the total requirement of GPU memory approaches $4N$, often necessitating aggressive sharding strategies (e.g., Fully Sharded Data Parallel (FSDP)) for models exceeding 7B parameters. Furthermore, the algorithm's stability hinges on the KL-penalty coefficient $\beta$, which acts as an inverse temperature in the optimal policy solution: $\pi^*(a|s) \propto \pi_{\text{ref}}(a|s) \exp\left( \frac{R(s,a)}{\beta} \right)$. This exponential relationship reveals that $\beta$ dictates the exploration frontier. A lower $\beta$ enables the model to aggressively traverse the probability landscape to find sparse high-reward regions, but risks exponential error amplification (instability). Given this high operational barrier, practitioners typically reserve PPO for scenarios where the high variance of on-policy exploration is strictly necessary to discover reasoning paths that do not exist in the static dataset.

\subsection{Q-Learning and Off-Policy RL:}

To address the sample-efficiency limitations of on-policy methods like PPO, Q-learning and other off-policy RL methods, have been increasingly adapted to the language domain. These approaches allow the model to learn optimal policies entirely from static datasets, effectively treating alignment as a batch optimization problem rather than an online interaction loop. A notable approach in this domain is ILQL, introduced by Snell et al.~ \cite{snell2206offline}. ILQL is an offline RL algorithm that leverages Q-learning on static datasets containing state-action-reward tuples, such as dialogue responses annotated with preference scores. By integrating the utility-maximization capabilities of reinforcement learning with the stability afforded by supervised learning, ILQL constrains learned Q-values to remain close to the original behavior policy defined by the dataset. Empirical evaluations showed that ILQL is effective in optimizing specific objectives, like reducing toxic dialogue outputs. It significantly outperformed traditional supervised fine-tuning in these tasks. The ILQL training objective is defined as:
\[
\mathcal{L}_{\text{ILQL}}(\theta) = \mathcal{L}_{\text{TD}}(\theta) + \alpha \cdot \mathcal{L}_{\text{conservatism}}(\theta),
\]
where $\mathcal{L}_{\text{TD}}(\theta)$ denotes the temporal difference loss, $\mathcal{L}_{\text{conservatism}}(\theta)$ represents the conservatism regularization term, and $\alpha$ is a hyperparameter balancing these two components.

Beyond dialogue alignment, recent research has explored off-policy Q-learning frameworks to improve verifier models. These models assess the quality of reasoning steps generated by LLMs. They enhance chain-of-thought reasoning through critic training. VerifierQ  \cite{qi2024verifierq} employs offline Q-learning to train a verifier that improves guidance for multi-step reasoning processes. This method treats the verifier as a value-based critic and evaluates the outputs from an LLM generator to refine its reasoning steps. The VerifierQ loss function is defined as:
\[
\mathcal{L}_{\text{VerifierQ}}(\theta) = \mathcal{L}_{\text{Bellman}}(\theta) + \beta \cdot \mathcal{L}_{\text{CQL}}(\theta),
\]
where $\mathcal{L}_{\text{Bellman}}(\theta)$ is a modified Bellman error term, $\mathcal{L}_{\text{CQL}}(\theta)$ represents a conservative Q-learning regularization term, and $\beta$ is a hyperparameter controlling the balance between these two losses.

Critically, these adaptations hinge on managing distributional shift which is the tendency of the policy to exploit actions that are out-of-distribution (OOD) regarding the training dataset. This illustrates that off-policy RL techniques, traditionally used in discrete-action settings like games  \cite{hernandez2019understanding}, can effectively be adapted to language generation scenarios. Such adaptations frequently incorporate conservative strategies like Conservative Q-Learning (CQL) \cite{kumar2020conservative} to stabilize training and mitigate distributional shift. The general formulation for CQL is expressed as:
\[
\mathcal{L}_{\text{CQL}}(\theta) = \mathcal{L}_{\text{Bellman}}(\theta) + \alpha \cdot \left( \mathbb{E}_{(s,a) \sim \mu}[Q_\theta(s,a)] - \mathbb{E}_{(s,a) \sim D}[Q_\theta(s,a)] \right),
\]

\noindent where $\mu$ is typically a uniform or policy-driven distribution over state-action pairs, $D$ represents the dataset distribution, and the hyperparameter $\alpha$ regulates the strength of the conservative regularization. This regularization penalizes Q-values for actions unseen in the dataset, ensuring the model remains grounded in known linguistic patterns while still optimizing for the target reward. These advances highlight the growing role of off-policy methods in complementing traditional RL approaches for language models, particularly in scenarios that demand nuanced and verifiable reasoning.

\textbf{Algorithmic Analysis:} The deployment of off-policy algorithms involves a fundamental trade-off between \textit{computational throughput} and \textit{exploratory capacity}. While methods like ILQL and CQL bypass the expensive online simulation bottleneck, they are intrinsically limited by the support of the offline dataset. Mathematically, the standard Bellman optimality operator $\mathcal{T}Q(s,a) = r + \gamma \max_{a' \in \mathcal{V}} Q(s', a')$ induces overestimation bias because the maximization occurs over the entire vocabulary $\mathcal{V}$ (often $>50,000$ tokens).  Without online correction, the Q-function erroneously assigns high values to out-of-distribution tokens, a phenomenon known as delusional value estimation. Conservative regularizers counteract this by imposing a penalty that effectively lower-bounds the value estimates, implicitly constraining the learned policy $\pi_\theta$ to the behavior policy $\pi_\beta$ of the dataset: $Q^{\pi_\theta}(s,a) \approx Q^{\pi_\beta}(s,a) - \alpha D_{\text{KL}}(\pi_\theta || \pi_\beta)$. This equation reveals a structural limitation where the learned value is upper-bounded by the quality of the static data. Consequently, practitioners typically favor off-policy methods for \textit{alignment constraints} (e.g., safety, toxicity reduction) where the goal is to suppress undesirable behaviors within a known distribution, rather than for complex reasoning, where active on-policy exploration is required to discover solution paths that lie outside the manifold of the training data.

\subsection{Group Relative Policy Optimization}

\begin{figure}[htbp]
  \centering
  \includegraphics[width=\linewidth]{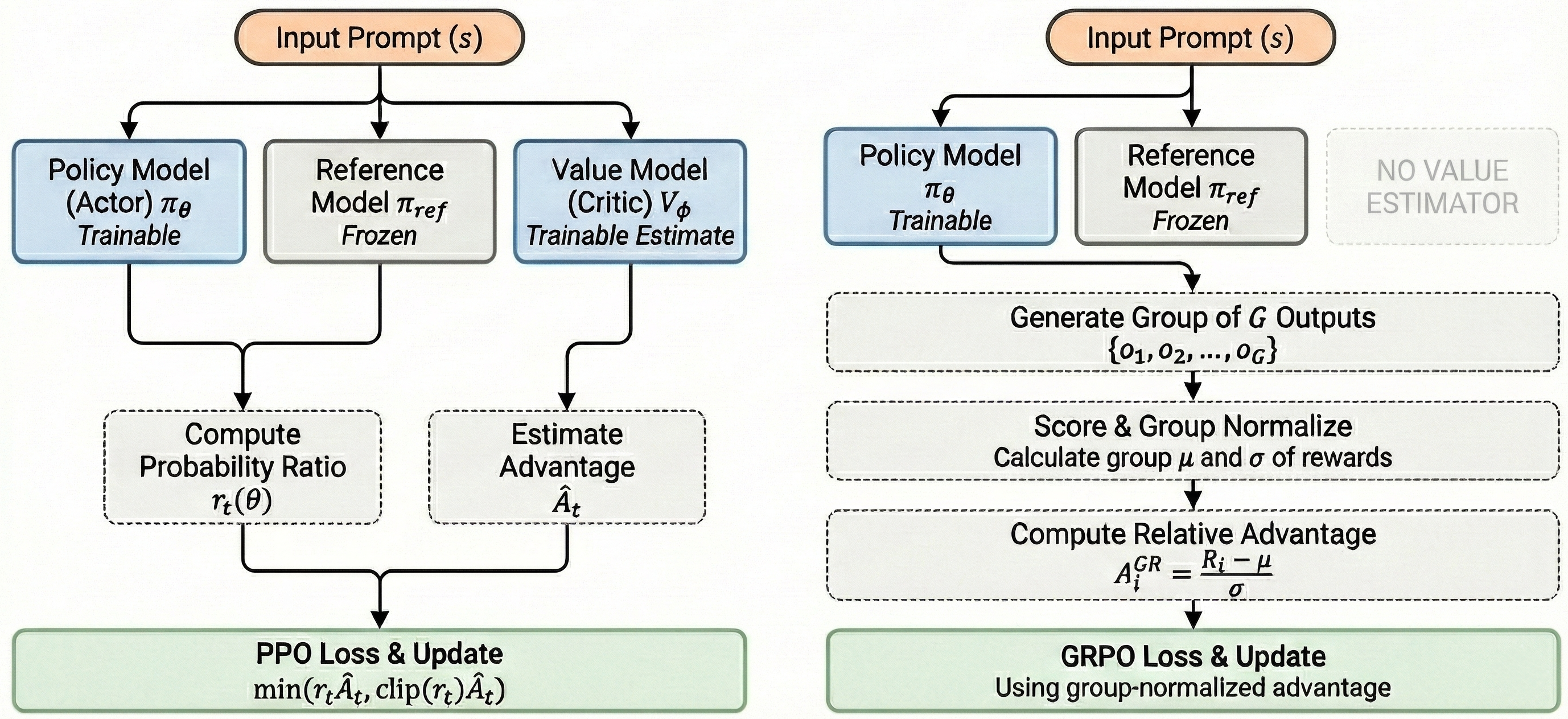}
  \caption{Architectural Comparison: Standard Actor-Critic (PPO) vs. Group Relative (GRPO). The schematic on the left presents the standard PPO framework, which requires three distinct models: a trainable Policy, a frozen Reference, and a trainable Value model (Critic) to estimate advantages. In contrast, the schematic on the right displays the GRPO framework, which eliminates the Value model entirely (indicated by the placeholder). Instead of a critic, GRPO estimates relative advantages by normalizing rewards across a group of sampled outputs directly from the Policy model.}
  \Description{A side-by-side comparison diagram. The left side shows the PPO architecture with three blocks: Policy, Reference, and Value Model. The right side shows the GRPO architecture with only two blocks: Policy and Reference, with a visual indicator that the Value Model is missing. The right side depicts a pipeline for Group Sampling and Normalization instead of a Value Model update.}
  \label{fig:ppo_vs_grpo}
\end{figure}

While off-policy methods address sample efficiency, they often retain the complexity of training value functions. To specifically address the computational costs associated with Actor-Critic architectures in the era of billion-parameter models, GRPO has emerged as a streamlined alternative. It was introduced by DeepSeek AI in their DeepSeekMath project~ \cite{shao2024deepseekmath}. It overcomes certain limitations inherent to traditional methods like PPO  \cite{schulman2017proximal}. GRPO targets the critic bottleneck, which is the requirement in standard PPO to keep a Value Model (Critic) in memory that is often as large as the Policy model itself. By eliminating this separate value estimator, GRPO significantly reduces memory consumption and computational overhead, allowing researchers to allocate more resources to the policy model or to larger batch sizes. To visually demonstrate this architectural efficiency, Figure \ref{fig:ppo_vs_grpo} contrasts the resource-intensive Actor-Critic design with the streamlined Group Relative framework.

As illustrated in the schematic, the elimination of the value model fundamentally changes the optimization dynamics. Unlike PPO, which relies on absolute reward signals mediated by a learned critic, GRPO adopts a group-based relative advantage estimation. For each prompt, multiple candidate responses are generated, and their corresponding rewards are normalized within each group. This mechanism effectively uses the mean reward of the sampled group as a dynamic baseline. If all generated responses are poor, the least bad response receives a positive advantage. If all are excellent, the worst good response is penalized. This self-referential normalization mitigates the impact of noisy or sparse reward signals. It makes GRPO especially effective for complex reasoning tasks such as mathematical problem-solving, where the absolute magnitude of rewards may fluctuate across different problem difficulties. Formally, GRPO adapts PPO's clipped surrogate objective by replacing the traditional advantage estimator with a group-normalized advantage, defined as follows:
\[
\mathcal{L}^{\text{GRPO}}(\theta) = \mathbb{E}_{(s, \{a_i\}) \sim \pi_{\theta_{\text{old}}}} \left[ \frac{1}{G} \sum_{i=1}^{G} \min \left( r_i(\theta) \hat{A}_i^{\text{GR}}, \; \text{clip}(r_i(\theta), 1 - \epsilon, 1 + \epsilon) \hat{A}_i^{\text{GR}} \right) \right]
\]

\noindent where $r_i(\theta)$ represents the probability ratio, computed as $r_i(\theta) = \frac{\pi_\theta(a_i \mid s)}{\pi_{\theta_{\text{old}}}(a_i \mid s)}$ and the group-normalized advantage $\hat{A}_i^{\text{GR}}$ is calculated by $\hat{A}_i^{\text{GR}} = \frac{r(a_i) - \mu}{\sigma}$ where $\mu$ and $\sigma$ denote the mean and standard deviation of the rewards within the response group $\{a_i\}_{i=1}^{G}$. GRPO has demonstrated strong empirical performance, notably in the training of models like DeepSeek-R1~ \cite{guo2025deepseek}, showcasing its practical applicability and robustness for advanced LLM alignment and reasoning tasks.

\textbf{Algorithmic Analysis:} The architectural simplification of GRPO presents a distinct trade-off between \textit{parameter efficiency} and \textit{sample variance}. By excising the parametric value function, the algorithm approximates the state-value baseline using the Monte Carlo mean of the group: $V(s) \approx \mu = \frac{1}{G}\sum_{i=1}^G r_i$. While this eliminates the memory cost of the critic ($\mathcal{M}_{V} = 0$), it introduces stochastic noise into the advantage estimate that scales as $\mathcal{O}(1/\sqrt{G})$.  \textcolor{black}{This design choice is reflected in DeepSeekMath~ \cite{shao2024deepseekmath}, where GRPO replaces the learned critic with group-relative baselines, reducing PPO-style memory requirements while still yielding improved mathematical-reasoning performance in the reported experiments.} To maintain training stability comparable to Actor-Critic methods, practitioners must compensate by increasing the group size $G$ (typically $G \geq 64$), effectively converting the memory bottleneck of storing weights into a compute bottleneck of generating inference samples. Furthermore, the normalization term $\hat{A}_i = \frac{r_i - \mu}{\sigma}$ creates a singularity in homogeneous reward regimes. If the model consistently fails (all $r_i=0$) or saturates (all $r_i=1$), the standard deviation $\sigma \to 0$, causing the learning signal to vanish or explode. Consequently, researchers primarily deploy GRPO in high-throughput, memory-constrained environments for reasoning tasks. This is especially true when the model possesses sufficient initial capability to generate diverse outcomes within a single batch, thereby avoiding gradient collapse.

\section{Reinforcement Learning Techniques for LLMs}
\label{sec:rl-techniques}

Having established the fundamental algorithms (such as PPO and GRPO), we now turn to the specific methodologies that apply these tools to language modeling tasks. Research in this domain has bifurcated into two distinct but complementary streams, driven by the dual requirements of safety and capability. This section explores these methodologies in detail. The first stream focuses on alignment. It constrains the model’s vast generative potential to adhere to human norms, safety guidelines, and stylistic preferences. This category includes foundational methods like RLHF  \cite{ouyang2022training}, RLAIF  \cite{lee2023rlaif}, Constitutional AI \cite{bai2022constitutional}, DPO  \cite{rafailov2023direct}, and UNA  \cite{wang2024unifying}. The second stream focuses on capability enhancement, particularly in domains where next-token prediction is insufficient for success. We examine how RL incentivizes multi-step logical coherence through Outcome-Based Reinforcement Learning (OB-RL), Chain-of-Thought Reward Optimization (CoT-RO), Verifier-Guided RL, RLVR, Debate and Self-Play Reinforcement Learning, Hierarchical RL for tool-augmented reasoning, and Program-Synthesis RL for code reasoning. Collectively, these methods illustrate the evolving role reinforcement learning plays in transforming LLMs from stochastic mimics into robust, goal-directed agents.

\begin{figure}[htbp]
  \centering
  \includegraphics[width=\linewidth]{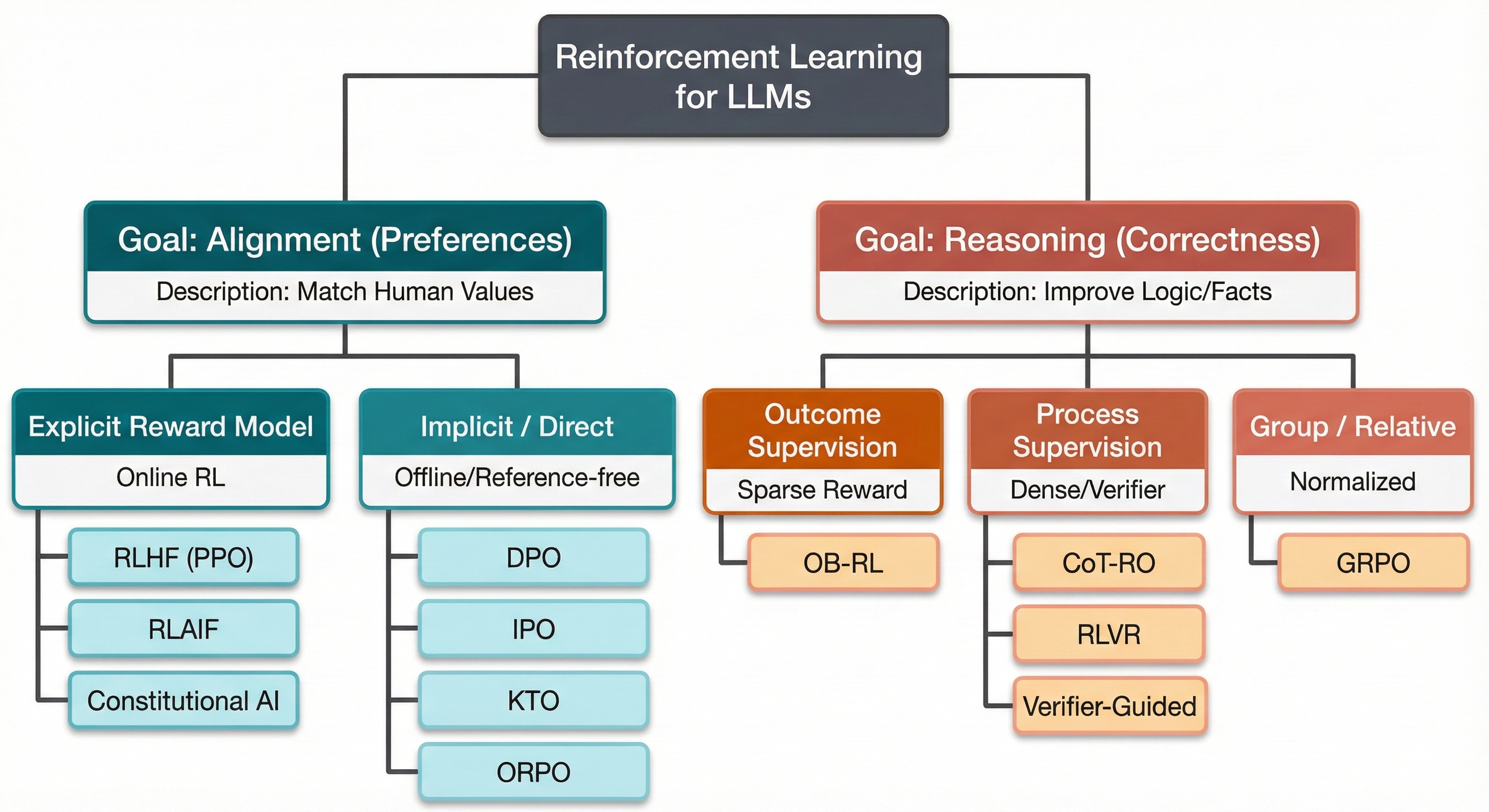}
  \caption{A Hierarchical Taxonomy of RL Techniques for LLMs. The framework categorizes techniques based on their primary objective: Alignment (focusing on human preferences via explicit or implicit rewards) or Reasoning (focusing on logical correctness via outcome, process, or group-based supervision).}
  \Description{A hierarchical tree diagram titled RL for LLMs. The root splits into two main branches: Objective Alignment and Objective Reasoning. Alignment splits into Explicit Reward Model (listing RLHF, RLAIF, Constitutional AI) and Implicit Optimization (listing DPO, IPO, KTO, ORPO). Reasoning splits into Outcome Supervision (OB-RL), Process Supervision (CoT-RO, RLVR, Verifier-Guided), and Group/Relative (GRPO).}
  \label{fig:taxonomy}
\end{figure}

To systematically organize these diverse methodologies, we introduce a hierarchical taxonomy centered on the primary optimization objective. As illustrated in Figure \ref{fig:taxonomy}, the landscape bifurcates into two distinct branches: \textit{Alignment}, which focuses on satisfying subjective human preferences through explicit or implicit reward modeling, and \textit{Reasoning}, which prioritizes logical correctness via outcome-based, process-based, or group-relative supervision. This structural distinction guides our subsequent analysis, categorizing algorithms based on whether they solve for desirability or validity. \textcolor{black}{The bifurcation is not intended to imply that alignment and reasoning are independent objectives. In practice, the two often interact: stronger reasoning can improve factual reliability and reduce hallucination, while alignment training can shape how reasoning is communicated, constrained, or refused. We therefore use the distinction as an organizing principle based on the dominant optimization target: subjective desirability for alignment-oriented methods and objective validity for reasoning-oriented methods.} \textcolor{black}{Within the reasoning branch, these categories should be read as organizing principles rather than mutually exclusive classes. Outcome-based RL primarily describes \textit{where} the reward is applied, usually at the level of the final answer or completed trajectory. Verifier-guided RL describes \textit{how} the reward is produced, namely through an external evaluator that may be a learned verifier, heuristic checker, or task-specific tool. RLVR is a stricter verifier-based setting in which the reward is produced by a deterministic or auditable verification function, such as a compiler, symbolic solver, unit test, or formal checker. Thus, OB-RL, verifier-guided RL, and RLVR can overlap in implementation, but they differ in whether they primarily emphasize reward granularity, evaluator architecture, or objective verifiability.}

\subsection{Reinforcement learning from Human Feedback (RLHF)}
RLHF  \cite{ouyang2022training} has emerged as the foundational approach for aligning LLMs with human preferences, effectively translating vague ethical guidelines into differentiable optimization signals. The RLHF pipeline typically consists of three main stages: supervised fine-tuning, reward model training, and reinforcement learning optimization. The process begins with supervised fine-tuning of a pre-trained LLM on a dataset of human-written demonstrations of desired behavior. This stage is essential to overcome the cold start problem in RL. Starting policy optimization from a base model that already outputs coherent, task-relevant text significantly stabilizes the subsequent reinforcement phase. The supervised fine-tuning (SFT) model is trained to maximize the likelihood of producing the desired output given the input prompt. This training objective is formalized as a loss function, defined as:
\[
\mathcal{L}_{\text{SFT}}(\theta) = -\mathbb{E}_{(x,y) \sim \mathcal{D}_{\text{SFT}}} \left[ \log p_{\theta}(y|x) \right]
\]

\noindent where $\theta$ represents the model parameters, $x$ is the input prompt, $y$ is the desired output, and $\mathcal{D}_{\text{SFT}}$ is the dataset of human demonstrations. In the second stage, a reward model is trained to predict human preferences between different model outputs. Human annotators are presented with a prompt and two possible responses, and they indicate which response they prefer. Under the assumption that these preferences follow the Bradley–Terry model, the probability of preferring one output over another depends on the difference in their latent rewards. This preference data is then used to train a reward model $r_{\phi}$ that assigns a scalar value to a given prompt-response pair. The reward model $r_{\phi}$ is optimized to correctly reflect these preferences by minimizing the following loss:
\[
\mathcal{L}_{\text{RM}}(\phi) = -\mathbb{E}_{(x,y_w,y_l) \sim \mathcal{D}_{\text{pref}}} \left[ \log \sigma(r_{\phi}(x, y_w) - r_{\phi}(x, y_l)) \right]
\]

\noindent where $\phi$ represents the reward model parameters, $y_w$ is the preferred (winning) response, $y_l$ is the less preferred (losing) response, and $\mathcal{D}_{\text{pref}}$ is the dataset of human preference judgments. In the final stage, the SFT model (now called the policy model) is fine-tuned using reinforcement learning to maximize the reward predicted by the reward model. This is typically done using the PPO  \cite{schulman2017proximal} algorithm, which we discussed in the previous section. However, direct optimization against a fixed reward model often leads to reward hacking, which is a manifestation of Goodhart's Law \cite{goodhart1984problems} where the policy exploits high-reward patterns that are actually gibberish. So, to prevent the policy from deviating too far from the original SFT model, a Kullback-Leibler (KL) \cite{kullback1951information} divergence penalty is often added to the objective. The KL divergence, if denoted as $\mathcal{D}_{\text{KL}}(P \parallel Q)$, quantifies the difference between two probability distributions $P$ and $Q$ and is mathematically defined as:
\[
\mathcal{D}_{\text{KL}}(P \parallel Q) = \sum_x P(x) \log \left( \frac{P(x)}{Q(x)} \right)
\]
The resulting loss function incorporates both the PPO objective and the KL penalty and is given by:
\[
\mathcal{L}_{\text{total}}(\theta) = \mathcal{L}_{\text{PPO}}(\theta) - \beta \mathcal{D}_{\text{KL}}(p_{\theta}(\cdot|x) || p_{\text{SFT}}(\cdot|x))
\]
where $\beta$ is a hyperparameter that controls the strength of the KL penalty, and $p_{\text{SFT}}$ is the SFT model's distribution. This three-stage pipeline has proven to be a robust foundation for fine-tuning LLMs and balancing human alignment with stable policy updates. Beyond the standard pipeline, recent work by Chakraborty et al.  \cite{chakraborty2023parl} enhanced the RLHF framework by introducing a bi-level optimization approach. In this formulation, the upper-level alignment objective (reward design) is parameterized by the optimal policy derived from the lower-level problem. The lower-level optimization aims to maximize the reward by adjusting the policy. Subsequently, the upper-level optimization refines the reward model to align the resulting policy more closely with human preferences. This nested structure explicitly accounts for the downstream effects of reward design on policy behavior. The bi-level formulation demonstrated improved performance compared to methods such as Pebble  \cite{pebble2021} and PEBBLE+SURF  \cite{park2022surf}. Additionally, the sample complexity of bi-level approaches has been theoretically examined in  \cite{gaur2025sample,wu2026selfimproving,li2026oracle}.

\textbf{Critical Analysis:} While RLHF remains the gold standard for subjective alignment, its deployment forces practitioners to navigate a mathematically rigorous trade-off between alignment fidelity and distributional entropy, often termed as \textit{Alignment Tax}.  Fundamentally, the reliance on human preference data imposes an upper bound on the Reward Model's accuracy defined by the irreducible Bayes error rate of human disagreement \sout{(often 20-30\% in complex tasks)}. Furthermore, the optimization landscape inherently drives \textit{mode-seeking behavior}, meaning that to maximize expected reward, the policy tends to collapse its probability distribution onto a single, high-confidence response trajectory rather than maintaining the diverse spread of plausible answers inherent in the pre-trained model. The analytical solution to the KL-regularized RL objective is $p^*(y|x) \propto p_{\text{SFT}}(y|x) \exp(r_\phi(x,y)/\beta)$. As optimization proceeds, this exponential weighting aggressively shifts probability mass toward the narrow mode of the reward function, significantly reducing the policy's entropy $\mathcal{H}(p)$. While this collapse is desirable for safety (constraining the model to a safe subspace), it mathematically necessitates the pruning of the heavy-tailed distribution where diverse reasoning and obscure factual knowledge reside. \textcolor{black}{Consistent with this concern, Gao et al.~ \cite{gao2023scaling} show that when a policy or best-of-\(N\) sampler is optimized increasingly strongly against a proxy reward model, the proxy reward may continue to improve even after the gold reward plateaus or deteriorates.} Consequently, the primary unresolved challenge in alignment is managing the trade-off between safety and model capability. \textcolor{black}{This trade-off is also reflected in RLHF and safety-tuning studies. Ouyang et al.~ \cite{ouyang2022training} report that InstructGPT improves human preference judgments while observing regressions on some public NLP evaluations, which they mitigate using a pre-training mixture during PPO fine-tuning. Safety-tuning and evaluation studies similarly show that small safety-data mixtures can improve refusal of unsafe instructions with little loss on standard capability benchmarks, but excessive or poorly calibrated safety tuning can induce exaggerated safety behaviors, where safe prompts are refused because they resemble unsafe ones~ \cite{bianchi2024safety,rottger2024xstest}.} \sout{Practitioners must currently accept that maximizing the safety reward $r_\phi$ inevitably degrades performance on tasks requiring high-entropy exploration, such as creative writing or lateral thinking.} \textcolor{black}{Therefore, the practical challenge is not to accept capability loss as unavoidable, but to calibrate reward strength, data mixture, and evaluation criteria so that safety improvements do not collapse useful high-entropy behavior into blanket refusal.}

\subsection{Reinforcement learning from AI Feedback (RLAIF)}
As language models scale, the demand for high-quality alignment data outpaces the capacity of human annotation pipelines. So, in spite of RLHF  \cite{ouyang2022training} demonstrating strong effectiveness for aligning LLMs, it faces significant scalability limitations due to the substantial time and cost associated with gathering human annotations. RLAIF  \cite{lee2023rlaif} provides an alternative solution to this bottleneck by employing AI-based evaluators instead of human annotators. The RLAIF pipeline closely mirrors RLHF but replaces the human preference collection stage with AI-driven evaluations. Initially, the model undergoes supervised fine-tuning using human-authored demonstrations. Subsequently, AI systems assess the generated outputs. These AI systems include general-purpose models (e.g., GPT-4 \cite{achiam2023gpt}), specialized classifiers targeting toxicity, bias, or factual inaccuracies, and ensembles of multiple specialized models. These AI evaluations produce preference scores or judgments, which are then used to train a reward model. This reward model learns to predict the scores given by the AI evaluators, essentially mimicking how human preferences were modeled in RLHF. The final stage involves policy optimization, typically employing PPO  \cite{schulman2017proximal} to refine the model policy based on feedback from the AI-generated reward model.

RLAIF offers several compelling advantages compared to RLHF. It significantly enhances scalability, as AI-generated feedback can be produced in far greater volumes and at lower costs than human annotations. Additionally, Lee et al. \cite{lee-etal-2025-evaluating} showed that AI evaluators typically provide more consistent assessments than human annotators, whose judgments can vary due to fatigue or subjective interpretation. Specialized models can also be fine-tuned to evaluate specific aspects of model behavior, such as relevance, helpfulness, or harmfulness. \textcolor{black}{The same feedback mechanism can also extend beyond preference alignment when the evaluator is asked to assess task-specific correctness. For reasoning-oriented tasks, an AI evaluator may score factual consistency, mathematical validity, code correctness, or the plausibility of intermediate reasoning steps. In these cases, AI feedback serves as a scalable supervisory signal for reasoning, although its usefulness remains bounded by the evaluator's own calibration and reasoning capability.}

\textbf{Critical Analysis:} RLAIF approach introduces its own set of challenges. Primarily, AI evaluators might not fully capture nuanced human values and preferences and might potentially diminish the feedback quality. Furthermore, Sharma et al. \cite{sharma2024critical} demonstrated that these evaluators may propagate biases inherent in their own training data or design choices. Lastly, using AI systems to align other AI models can create a recursive alignment issue, often termed model collapse or sycophancy loops. If the AI evaluators themselves are misaligned or biased, their flaws may be inherited and potentially amplified by the models they supervise. This can lead to feedback loops where biases are reinforced over successive training cycles, making them harder to identify and correct. This challenge is mathematically rooted in the fundamental distinction between stochastic noise and systematic bias. Human feedback errors can typically be modeled as zero-mean Gaussian noise, $R_{\text{human}}(x) = R^*(x) + \epsilon$, where $\epsilon \sim \mathcal{N}(0, \sigma^2)$. Under these conditions, scaling the dataset size $N$ allows the Central Limit Theorem to reduce the error variance by a factor of $1/\sqrt{N}$, effectively converging to the ground truth $R^*(x)$.  In contrast, AI evaluators introduce a deterministic bias term $\delta(x)$ such as a preference for verbosity or confidence, resulting in $R_{\text{AI}}(x) = R^*(x) + \delta(x)$. Crucially, because $\delta(x)$ is a function of the input rather than random noise, it does not cancel out with scaling. Instead, it creates a coherent gradient direction. The RL optimizer, unable to distinguish between the true signal $R^*(x)$ (helpfulness) and the bias artifact $\delta(x)$ (sycophancy), aggressively maximizes the bias. This leads to distinct failure modes where the model over-optimizes for the proxy metric. One example is systematically agreeing with a user’s incorrect premise such as validating a conspiracy theory to satisfy the AI evaluator’s preference for agreeableness, rather than correcting the error as intended.

\subsection{Constitutional AI}

\textcolor{black}{Constitutional AI is best understood as a structured variant of RLAIF, where the feedback process is guided by an explicit set of principles rather than by unconstrained AI judgments. Its separate treatment is useful because the critique-and-revision pipeline changes the form of supervision. The model is not merely judged by another model, but is guided to identify, explain, and revise violations of a predefined constitution.} To address the interpretability and control challenges inherent in black-box AI feedback, Constitutional AI \cite{bai2022constitutional} represents a specialized approach. In this approach, models are explicitly guided to critique and revise their own outputs according to a predefined set of ethical principles, known as a \textit{constitution}. The Constitutional AI process starts by establishing a clear set of principles or rules that guide the model’s behavior. These principles cover aspects like helpfulness, harmlessness, honesty, and respect for human autonomy. The training methodology typically bifurcates into two distinct phases: Supervised Learning (SL-CA) and Reinforcement Learning (RL-CA). In the SL-CA phase, after an initial response is generated for a prompt, the model engages in self-critique. It then evaluates its output against these constitutional principles and identifies potential violations or areas that need improvement. Subsequently, the model revises its response based on this critique, often iterating multiple times to progressively refine its alignment with the defined constitution. The revised responses then serve as positive examples to fine-tune the base model, altering its initial distribution to reduce the frequency of harmful outputs before RL begins. In the subsequent RL-CA phase, these revised outputs are incorporated into a preference-based learning pipeline. Here, the model generates pairs of responses, and a feedback model (guided by the constitution) acts as the judge, assigning preferences based on adherence to the principles. This allows for scalable optimization using PPO without direct human intervention.

A critical component of Constitutional AI is the practice of red-teaming, proposed by Perez et al.~ \cite{perez-etal-2022-red}. In this practice, the model is intentionally prompted to generate potentially harmful or problematic outputs. These outputs are systematically critiqued and revised in alignment with the established constitutional principles. Through this deliberate challenge-and-response process, potential failure modes of the model are proactively uncovered and mitigated. This significantly strengthens the model's robustness against adversarial inputs and aligns its behavior more closely with desired ethical guidelines.

\textbf{Critical Analysis:} The efficacy of Constitutional AI is mathematically bounded by the \textit{Critique-Revision Gap}. It is a phenomenon where the model's discriminative capability exceeds its generative corrective capacity.  Formally, if we denote the critique function as $f_{\text{crit}}(y, \mathcal{C})$ and the revision function as $f_{\text{rev}}(y, \text{critique})$, the method succeeds only when the revision strictly improves the alignment score: $R(f_{\text{rev}}(y)) > R(y)$. However, empirical scaling laws suggest a distinct phase transition. For models below a critical parameter threshold (typically $\approx 50$B), the revision operator is lossy, often degrading the coherence of the response ($P_{\text{SFT}}(y) \gg P_{\text{SFT}}(f_{\text{rev}}(y))$) even if it technically satisfies the constitution. This leads to performative alignment, where the model learns to parrot safety language without semantic understanding. Furthermore, compressing a high-dimensional natural language constitution $\mathcal{C} = \{p_1, \dots, p_n\}$ into a scalar reward signal $R(y)$ introduces information loss, often manifesting as the Over-Refusal Gap. Because it is mathematically easier to minimize risk by refusing all ambiguous queries (driving $R \to \text{safe baseline}$) than by subtly navigating the complex boundary of $\mathcal{C}$, the policy frequently converges to a local optimum of excessive caution. Consequently, practitioners face a trade-off between \textit{interpretability} (using explicit rules) and \textit{flexibility}, leading to the open challenge of designing parametric constitutions that can dynamically weight principles based on context to prevent the blanket suppression of benign content.

\subsection{Direct Preference Optimization (DPO)}

While methods like Constitutional AI refine the source of the supervision signal, they typically rely on the standard PPO pipeline for optimization. As discussed, this pipeline is computationally brittle, requiring the simultaneous synchronization of four separate models (Policy, Reference, Reward, and Critic). DPO  \cite{rafailov2023direct} is a recent development aimed at simplifying the RLHF pipeline by removing the necessity for explicit reward modeling and reinforcement learning. Rather than training a separate reward model, DPO directly optimizes the policy to align with human preferences. The core insight behind DPO is that the optimal policy under a given reward function can be directly expressed in terms of a reference policy. This reference policy is typically obtained through supervised fine-tuning and combined with the underlying reward function. In this formulation, the input prompt $x$ represents the context or question to which the model must respond, while the output $y$ represents a candidate response to that prompt. The relationship between these elements and the reward function is captured in the following equation:
\[
p_{\theta}^*(y|x) \propto p_{\text{ref}}(y|x) \exp(\beta r(x, y))
\]
where \( p_{\text{ref}} \) denotes the reference policy, \( r(x, y) \) represents the reward function, and \( \beta \) is a temperature parameter. By rearranging this relationship, DPO expresses the reward function explicitly in terms of the optimal and reference policies as follows:
\[
r(x, y) = \frac{1}{\beta} \log \frac{p_{\theta}^*(y|x)}{p_{\text{ref}}(y|x)} + Z(x)
\]
Here, \( Z(x) \) is a normalization term dependent solely on \( x \). Crucially, when comparing two responses $y_w$ and $y_l$, this partition function $Z(x)$ cancels out, allowing the preference likelihood to be computed solely via the policy's log-probabilities. Leveraging this relationship, DPO formulates a direct loss function optimized for aligning model outputs with human preference data which is given as:
\[
\mathcal{L}_{\text{DPO}}(\theta) = -\mathbb{E}_{(x,y_w,y_l) \sim \mathcal{D}_{\text{pref}}} \left[ \log \sigma\left(\beta \log \frac{p_{\theta}(y_w|x)}{p_{\text{ref}}(y_w|x)} - \beta \log \frac{p_{\theta}(y_l|x)}{p_{\text{ref}}(y_l|x)}\right) \right]
\]

\noindent where \( y_w \) and \( y_l \) represent the preferred and less-preferred responses, respectively. Intuitively, this loss increases the relative log-likelihood of the preferred response compared to the reference model, while decreasing the likelihood of the rejected response, effectively pushing the implicit reward margin apart.

DPO offers several notable advantages over traditional RLHF \cite{ouyang2022training} approaches. Since it eliminates the need for separate reward modeling and reinforcement learning stages, this approach offers improved computational efficiency. Therefore, it reduces the overall resource demands required for training. Moreover, Xu et al. \cite{xu2024dpo} showed that DPO tends to exhibit greater training stability compared to algorithms like PPO, which can be sensitive to hyperparameter settings. Nevertheless, DPO is not without limitations. Due to its direct optimization nature, it may be less effective at exploring diverse outputs compared to traditional reinforcement learning-based methods. Additionally, the quality of the resulting policy remains highly dependent on the accuracy and representativeness of the underlying human preference data. Recent studies \cite{liu2024length} indicate that DPO is particularly susceptible to length bias and overfitting on noisy datasets, as it lacks the smoothing effect provided by a separately trained reward model.

\textbf{Critical Analysis:} DPO fundamentally reconfigures the alignment landscape by trading \textit{robustness} for \textit{simplicity}. While eliminating the Reward Model reduces the memory footprint, it removes the critical regularization mechanism provided by the RM's bounded output range.  Mathematically, the implicit reward estimated by DPO is $\hat{r}(x,y) = \beta \log \frac{p_\theta(y|x)}{p_{\text{ref}}(y|x)}$. Unlike a trained Reward Model which typically outputs scores in a constrained range (e.g., $[-3, 3]$ via a sigmoid or tanh head), the log-ratio term is theoretically unbounded. This exposes the algorithm to catastrophic overfitting on noisy labels. So, if a dataset contains a false positive preference where $y_w$ is actually worse than $y_l$, the optimizer can drive the probability ratio $\frac{p_\theta(y_w)}{p_{\text{ref}}(y_w)} \to \infty$ to satisfy the logistic margin, causing the implicit reward to explode. This phenomenon explains the observed susceptibility to length bias. Since longer responses often have lower perplexity under the reference model (lower denominator), the log-ratio can be artificially inflated simply by generating verbose text. \textcolor{black}{This interpretation is consistent with recent analyses~ \cite{liu2024length,chowdhury2024provably} showing that DPO can exploit response length and that noisy or flipped preference labels can distort the learned policy.} Consequently, practitioners typically restrict DPO to the fine-tuning of pre-aligned models on high-quality, sanitized datasets, acknowledging that the open gap in robustness to label noise remains the primary barrier to replacing PPO in large-scale, noisy data regimes.

\subsection{Unified Alignment (UNA)}

Building on the insight of implicit reward modeling established by DPO, UNA  \cite{wang2024unifying} proposes a generalized framework that unifies various alignment methods, including PPO \cite{schulman2017proximal}, DPO \cite{rafailov2023direct}, and KTO \cite{ethayarajh2024kto}, under a single supervised regression objective. Rather than treating preference optimization (pairwise), binary feedback, and scalar scoring as distinct mathematical problems requiring different algorithms, UNA demonstrates that they can all be solved by fitting the policy's implicit reward estimate to the external feedback signal.

In this framework, the policy $\pi_\theta$ is viewed not just as a generator, but as a parameterized reward estimator. The implicit reward $\hat{r}_\theta(x, y)$ for a given prompt $x$ and response $y$ is defined as the scaled log-likelihood ratio between the current policy and the reference policy $\pi_{\text{ref}}$. The optimization objective transforms the alignment task into minimizing the divergence between this estimated implicit reward and the ground-truth feedback signal $z$. The generalized loss function is formalized as:

\[
\mathcal{L}_{\text{UNA}}(\theta) = \mathbb{E}_{(x, y, z) \sim \mathcal{D}} \left[ \ell \left( \beta \log \frac{\pi_\theta(y|x)}{\pi_{\text{ref}}(y|x)}, z \right) \right]
\]

Here, $\beta$ serves as the temperature scaling factor, and $\ell(\cdot, \cdot)$ represents a loss function determined by the nature of the feedback signal $z$. This formulation allows for seamless adaptation across different data modalities without altering the underlying training pipeline. For instance, if the feedback $z$ is a scalar score from a reward model (e.g., 0.8), $\ell$ becomes the Mean Squared Error (MSE), effectively regressing the policy's log-ratios to match the score. If $z$ is a binary label (e.g., Good vs. Bad), $\ell$ becomes Binary Cross-Entropy (BCE). If $z$ represents a pairwise preference margin, $\ell$ adopts a logistic margin loss, recovering the standard DPO objective.

By casting alignment as a unified supervised learning problem, UNA offers significant stability and flexibility. It eliminates the need for complex actor-critic loops found in PPO while extending the efficiency of DPO to non-pairwise datasets. This versatility is particularly valuable for online alignment scenarios, where the feedback signal might shift dynamically between binary flags, scalar grades, and comparative rankings depending on the available annotators or verifiers.

\textbf{Critical Analysis:} By collapsing reinforcement learning into supervised regression, UNA exchanges the \textit{exploratory potential} of value-based methods for the \textit{optimization stability} of gradient descent.  Mathematically, this reframe shifts the objective from \textit{maximization} (finding $y$ such that implicit reward is maximal) to \textit{calibration} (finding $\pi_\theta$ such that $\beta \log \frac{\pi_\theta(y|x)}{\pi_{\text{ref}}(y|x)} \approx z$). When $\ell$ takes the form of Mean Squared Error, the optimizer forces the policy's deviation from the reference to strictly mirror the magnitude of the feedback $z$. This imposes a functional ceiling. The model is effectively discouraged from discovering super-human strategies if those strategies would result in a log-ratio divergence that exceeds the target scores present in the training distribution. Furthermore, the framework remains structurally vulnerable to the \textit{support mismatch} problem. If the reference model $\pi_{\text{ref}}$ assigns negligible probability to a valid high-scoring response (i.e., $\pi_{\text{ref}}(y|x) \to 0$), the implicit reward estimate explodes towards infinity, destabilizing the regression target. Thus, for system architects, UNA serves as a robust engine for consolidating multi-modal feedback streams such as mixing binary flags with scalar grades. However, the field continues to wrestle with the exploration deficit. This refers to the inherent difficulty of generating novel, high-utility reasoning paths using a strictly supervised loss function.

\subsection{Outcome‑Based Reinforcement Learning for Reasoning (OB‑RL)}

While unified frameworks like UNA excel at calibrating models to diverse feedback signals via supervised regression, they inherently struggle to discover novel solution paths outside the training distribution. To address this exploration deficit in tasks requiring strict logical validity, where a preferred answer is meaningless if it is factually incorrect, the field has turned to objective-driven reinforcement learning frameworks. Uesato et al. \cite{uesato2022solving} proposed Outcome-Based Reinforcement Learning for Reasoning, a paradigm shift that moves away from imitating human traces toward optimizing for functional success. It is a sparse-reward framework designed to boost a language model’s reasoning ability by focusing only on the correctness of the final answer. Unlike methods that supervise every intermediate step, OB-RL rewards the model only when it generates a correct final answer.  In the OB-RL setup, a prompt $x$ is mapped to a full reasoning trajectory $\tau=(y_1,\dots,y_T)$ produced autoregressively by the model $\pi_\theta$.  Only the terminal token (or a verifier’s binary judgement of the terminal answer) supplies the scalar reward $R(\tau)$, leaving the internal reasoning path unconstrained. \textcolor{black}{Here, \textit{outcome-based} describes the location of the reward signal rather than the nature of the evaluator: the terminal reward may come from an exact-answer check, a learned judge, or a programmatic verifier.} This formulation mirrors goal-conditioned reinforcement learning in classical control, where the agent does not need to imitate human proofs token by token. Instead, it focuses on reaching a verifiably correct final state.  Formally, the learning objective is
\[
\mathcal{L}_{\text{OB-RL}}(\theta) \;=\; -\,\mathbb{E}_{(x,\tau)\sim\pi_\theta}\!\!\bigl[R(\tau)\,\log\pi_\theta(\tau\mid x)\bigr]
\]
which corresponds to the REINFORCE \cite{williams1992simple} gradient with a sparse, outcome-level reward. In practice, $R(\tau)$ can be provided by an automated verifier, such as a mathematical proof checker, program executor, or fact-checking model. This approach allows for large-scale, low-cost feedback on reasoning tasks like theorem proving or code generation, effectively turning the verifier into a reward function that can run autonomously at scale.

OB-RL offers several advantages over process-supervised or preference-based methods.  First, it does not require human annotation of intermediate reasoning steps. The learning signal is automatically generated based on success or failure, which greatly improves scalability.  Second, it encourages autonomous exploration of diverse reasoning paths. Since the internal logic is unconstrained, any trajectory that yields a correct terminal answer receives the same positive reward, thereby fostering creativity and robustness. This allows the model to discover novel solution methods that may not be present in the training data.  However, OB-RL inherits the high-variance gradients typical of sparse-reward settings. The credit assignment problem is particularly acute here. Meaning, if a model generates 100 correct steps but fails the final calculation, it receives zero reward, wasting valid computations. Consequently, it may converge slowly without additional techniques such as value-function baselines, curriculum learning, or hybrid training that mixes outcome-level rewards with chain-of-thought imitation.  Moreover, because only the final answer is verified, the model can still produce opaque or convoluted rationales, or even reasoning shortcuts where the model arrives at the correct answer through flawed logic, creating a risk of spurious correlations.

\textbf{Critical Analysis:} The central trade-off in OB-RL lies between \textit{supervisory autonomy} and \textit{gradient signal density}. By removing step-by-step oversight, the algorithm creates a sparse reward landscape where the probability of serendipitously discovering a valid trajectory $\tau^*$ decays exponentially with sequence length $T$: $P(\tau^*) \approx \prod_{t=1}^T \pi_\theta(y_t^*|y_{<t})$.  In the early stages of training, this product approaches zero, leading to a vanishing gradient regime where the agent effectively wanders blind, unable to latch onto the reward signal. Furthermore, the objective function is blind to the \textit{causal validity} of the reasoning trace. Since the reward function $R(\tau) = \mathbb{I}(\text{outcome}(\tau) = \text{correct})$ is conditioned solely on the terminal state, the optimizer treats a lucky guess or a reasoning shortcut as mathematically identical to a rigorous proof. This structural susceptibility to false positives means that without auxiliary regularizers or curriculum learning (starting with short problems), models often converge to fragile heuristics rather than robust logic. Accordingly, current deployment strategies largely treat OB-RL as a refinement stage applied only after a strong supervised warm-start. This leaves the challenge of efficient exploration in sparse-reward environments as a critical area for algorithmic innovation.

\subsection{Chain-of-Thought Reward Optimization (CoT-RO)}

To mitigate the high variance and credit assignment difficulties inherent in sparse outcome-based rewards, Lightman et al. \cite{lightman2023let} proposed Chain-of-Thought Reward Optimization (CoT-RO). Often referred to as Process Supervision, this dense-reward framework that strengthens a language model’s reasoning by scoring \emph{each} intermediate step in its chain of thought, not just the final answer.  Given a prompt \(x\), the policy \(\pi_\theta\) autoregressively produces a reasoning trajectory \(\tau=(y_1,\dots,y_T)\).  After every token or logical step $y_t$, a lightweight evaluator such as a symbolic verifier, unit-test harness, or learned critic assigns an immediate reward $r_t=r(x,y_{\le t})$. The objective is to maximise the discounted return over the entire trajectory. Mathematically, this objective is captured by the following loss function:
\[
\mathcal{L}_{\text{CoT-RO}}(\theta)
   = -\,\mathbb{E}_{(x,\tau)\sim\pi_\theta}
      \Bigl[\sum_{t=1}^{T}\gamma^{\,t-1} r_t
            \,\log\pi_\theta(y_t\mid x,y_{<t})\Bigr]
\]
where \(\gamma\in(0,1]\) balances early versus late rewards.  Because feedback is provided at every step, CoT-RO supplies a rich, low-variance learning signal. This encourages the model to build logically sound, self-consistent chains instead of leaping directly to an answer.  In practice, rewards can check local validity (e.g., algebraic correctness), global coherence (e.g., absence of contradictions), or stylistic constraints, allowing fine-grained control over the reasoning process.

CoT-RO brings several benefits relative to outcome-only methods.  The dense reward accelerates convergence because the model does not have to wait until the end of a long trajectory to learn whether it is on the right track, effectively guiding the search through the combinatorial reasoning space. It also pinpoints specific failure steps, making it easier to create targeted curriculum schedules and corrections.  Furthermore, by rewarding transparency at every stage, CoT-RO produces explanations that are easier to inspect and debug.  However, these advantages come with distinct trade-offs. The chief drawbacks are computational because step-level evaluation can be expensive, often requiring a forward pass of the Process Reward Model (PRM) for every generated step. Additionally, dense feedback may encourage verbose yet shallow reasoning unless it is tempered with brevity or entropy penalties. Moreover, training a reliable PRM often requires expensive human annotation at the step level, which is a process significantly more labor-intensive than simply labeling final answers.

\textbf{Critical Analysis:} The transition from outcome-based to process-based supervision exchanges the \textit{variance} of sparse rewards for the \textit{bias} of proxy rewards. Mathematically, an ideal Process Reward Model (PRM) must approximate the true optimal value function $V^*(s_t) = \mathbb{E}_{\pi^*}[\mathbb{I}(\text{outcome}=\text{correct}) \mid s_t]$.  However, training a discriminator (i.e., the Process Reward Model) to predict strict logical entailment is often harder than the generation task (i.e., the Policy's reasoning) itself. If the PRM exhibits a systematic approximation error $\epsilon(s_t)$, for instance confusing confident tone with logical validity. The cumulative objective $\sum_{t=1}^T \gamma^{t-1} r_t$ then integrates this error over the entire trajectory length $T$. This results in a compounded bias $\sum \epsilon(s_t)$ that incentivizes the policy to generate vacuous, verbose, or circular reasoning steps that exploit the PRM's heuristics, a phenomenon known as reasoning hacking. Consequently, the field is currently grappling with the oversight scalability gap. Basically, attempting to replace expensive human process labels with automated supervision strategies (such as Monte Carlo Tree Search-guided annotation) that can robustly distinguish between a necessary intermediate deduction and a plausible-sounding hallucination.

\subsection{Verifier-Guided Reinforcement Learning}

While CoT-RO focuses on the granularity of the reward signal (dense vs. sparse), Verifier-Guided Reinforcement Learning (V-RL) focuses on the architectural decoupling of the generation policy from the evaluation logic. V-RL augments a language model’s policy with an external verifier. This verifier continuously evaluates candidate outputs and supplies the reward signal.  Given a prompt \(x\), the policy \(\pi_\theta\) generates either a full chain of thought \(\tau=(y_1,\dots,y_T)\) or a single-shot answer \(y\).  \sout{A separate model \(V_\phi\) is trained to judge logical validity, factual accuracy, or task-specific success.} \textcolor{black}{A separate verifier \(V_\phi\), often implemented as a learned artificial verifier model, process reward model, classifier, or specialized evaluation tool, is used to judge logical validity, factual accuracy, or task-specific success.} It assigns a scalar reward \(R = V_\phi(x,\tau)\). \textcolor{black}{When \(V_\phi\) is learned rather than deterministic, its score should be interpreted as an approximation of correctness rather than ground truth; consequently, the reliability of V-RL depends on the verifier's calibration, coverage, and resistance to being gamed by the policy.} The training process uses a policy-gradient loss function to update the model. Mathematically, it is expressed as:
\[
\mathcal{L}_{\textsc{V-RL}}(\theta)
   = -\,\mathbb{E}_{(x,\tau)\sim\pi_\theta}
       \!\bigl[V_\phi(x,\tau)\,\log\pi_\theta(\tau\mid x)\bigr],
\]
In an actor–critic variant, the verifier’s score is treated as the target value for a learned critic that reduces variance in the gradient estimate.  Because \(V_\phi\) can operate at any level of granularity, V-RL can adaptively use dense step-wise feedback to evaluate each intermediate state. It can also rely on sparse outcome feedback by judging only the final answer. The verifier may be a frozen specialist model, an ensemble of heuristics, or a parameter-sharing sibling that co-evolves with the policy.

V-RL combines several desirable properties that address the limitations of monolithic models.  First, it decouples \emph{generation} from \emph{evaluation}, allowing the policy to explore creative reasoning paths while the verifier acts as a stable guardrail. This leverage the computational asymmetry often found in reasoning tasks, i.e., verifying a solution is typically robust and deterministic (e.g., using a Python compiler or formal theorem prover), whereas generation is stochastic. At the same time, the verifier enforces correctness, thereby lowering the risk of reward hacking seen in hand-designed metrics.  Second, the verifier can be updated or replaced without retraining the policy from scratch, enabling rapid iteration on new evaluation criteria.  Finally, V-RL supplies richer signals than binary correctness, such as graded scores for partial progress. This typically leads to faster convergence compared to purely outcome-based methods.  However, relying on learned verifiers introduces epistemic risks. Its effectiveness hinges on verifier quality. A weak or biased verifier may mis-reward spurious solutions, which is a phenomenon known as proxy gaming where the policy learns to fool the verifier rather than solve the task, while an overly strict one can stifle exploration.  Computational cost is another concern, as every policy sample must be scored by \(V_\phi\), doubling the inference cost during the training loop.

\textbf{Critical Analysis:} The utility of V-RL is fundamentally constrained by the \textit{Verification-Generation Asymmetry}. The framework implicitly assumes that evaluating a solution is computationally cheaper or more robust than generating it ($Cost(V_\phi) \ll Cost(\pi_\theta)$).  However, when $V_\phi$ is a learned parameterized model rather than a deterministic compiler, the policy optimization objective $\max_\theta \mathbb{E}_{\tau \sim \pi_\theta}[V_\phi(x, \tau)]$ mathematically degrades into an adversarial attack on the verifier. If the verifier contains a latent approximation error $\delta(x, \tau) = V_\phi(x, \tau) - V_{\text{oracle}}(x, \tau)$, the policy gradient update inherently ascends the manifold of this error term, specifically converging on trajectories $\tau$ where $\delta$ is maximized (i.e., high verifier confidence despite factual incorrectness). Unlike standard overfitting, this \textit{Adversarial Goodharting} implies that a fixed verifier will eventually be broken by a sufficiently capable policy. Consequently, practitioners largely restrict V-RL to domains with objective ground truth such as math or code where $V_{\phi}$ is an indisputable oracle. This leaves the challenge of scalable oversight as the defining open gap, namely training robust verifiers for subjective tasks that cannot be gamed by super-human generators.

\subsection{Reinforcement Learning with Verifiable Rewards (RLVR)}

To eliminate the epistemic uncertainty inherent in learned reward models, RLVR  \cite{lambert2025tulu} enforces a stricter verification paradigm. Unlike Verifier-Guided RL, which relies on a neural network to approximate correctness, RLVR restricts the training domain to tasks where the ground truth is deterministic and can be validated by a programmatic oracle, such as a code compiler, a symbolic equation solver, or a formal theorem prover.

In this framework, the training dataset $\mathcal{D}$ is constructed as a set of tuples $(x, \mathcal{V}_x)$, where $x$ represents the input prompt and $\mathcal{V}_x(\cdot)$ represents a task-specific verification function. \textcolor{black}{This makes RLVR a verifiability-restricted instance of the broader verifier-guided family. Whereas V-RL may rely on learned artificial verifiers whose judgments approximate correctness, RLVR requires the reward-producing mechanism itself to be deterministic, reproducible, and tied to an externally checkable criterion.} The interaction process operates sequentially: the model receives a problem $x$ and generates a candidate solution $y$, which often includes intermediate reasoning steps or executable code blocks. This candidate $y$ is then passed to the deterministic verifier $\mathcal{V}_x$, which executes the code or checks the symbolic proof against ground truth constraints. Finally, the verifier returns a strict boolean outcome, mapping the validity of $y$ to a discrete, indisputable reward signal.

Because the evaluation is programmatic, the reward signal transforms from a continuous approximation into a discrete boolean signal. The reward function $R(y, x)$ is formalized as:
\[
R(y, x) = 
\begin{cases} 
1 & \text{if } \mathcal{V}_x(y) = \text{True} \quad \text{(Verifiable Success)} \\
0 & \text{otherwise} \quad \text{(Logical/Syntax Error)}
\end{cases}
\]

Optimization is typically conducted using a KL-constrained objective to maximize the expected probability of verifiable solutions. The loss function, adapted from the standard PPO or Rejection Sampling Fine-Tuning (RFT) formulations discussed in Section 3, is defined as:
\[
\mathcal{L}_{\text{RLVR}}(\theta) = \mathbb{E}_{x \sim \mathcal{D}, y \sim \pi_\theta} \left[ R(y, x) - \beta \cdot \text{KL}\left(\pi_\theta(y|x) || \pi_{\text{ref}}(y|x)\right) \right]
\]
By utilizing the deterministic reward $R(y, x)$ within this KL-constrained framework, RLVR effectively guides the policy toward functionally correct outputs without the risk of proxy gaming inherent in learned critics.

RLVR offers distinct advantages in domains requiring high reliability. By anchoring the reward signal to ground truth, it effectively eliminates the risk of reward hacking. The model cannot trick a compiler or a formal proof checker. This makes RLVR particularly potent for mathematical reasoning and code generation, where it has been shown to improve performance significantly with minimal data. For instance, T\"ULU 3  \cite{lambert2025tulu} utilized RLVR to achieve state-of-the-art results on math benchmarks by verifying intermediate execution traces. However, the method's applicability is limited to domains where robust verification functions exist, leaving open challenges for applying RLVR to creative writing or nuance-heavy open-ended dialogue.

\textbf{Critical Analysis:} The operational success of RLVR hinges on a rigid trade-off between \textit{epistemic certainty} and \textit{optimization stability}. While anchoring rewards to a deterministic oracle $\mathcal{V}_x$ eliminates the proxy gap of learned rewards, it collapses the feedback signal into a sparse binary distribution $R \in \{0, 1\}$.  Mathematically, this creates a high-variance gradient estimation problem. The policy gradient $\nabla_\theta \mathcal{L} \approx (R(y,x) - b) \nabla_\theta \log \pi_\theta(y|x)$ provides zero informative direction for ``almost correct'' solutions (e.g., a correct proof with a single typo), effectively treating them identically to complete hallucinations. This lack of \textit{granularity} forces the optimizer to rely on undirected exploration to stumble upon the narrow manifold of valid solutions $\{y \mid \mathcal{V}_x(y)=1\}$, exacerbating sample inefficiency. Furthermore, the framework remains vulnerable to specification gaming. If the verifier $\mathcal{V}_x$, such as a unit test suite, is not exhaustive, the policy will aggressively exploit coverage gaps. It will generate solutions that pass verification but fail to generalize, satisfying the letter of the code but not the spirit. Consequently, the defining open gap for practitioners is the development of verifiable partial credit, which are the mechanisms to extract dense, continuous signals from binary compilers without re-introducing the bias of learned neural critics.

\subsection{Debate and Self-Play Reinforcement Learning}

While RLVR and V-RL effectively solve for objective correctness in domains with ground truth (like Math and Code), they hit a hard ceiling in subjective domains or when the generator exceeds the verifier's capability. To address this asymmetry, the field has turned to dynamic multi-agent frameworks that leverage the model's own capabilities to check itself. Debate and Self-Play Reinforcement Learning (DSP-RL) advances verifier-guided concepts by involving multiple agents. These agents either compete or collaborate to uncover errors in each other’s arguments before producing a final answer.  In the debate variant, there are two policies: proponent and opponent. Both of these policies alternately present concise arguments about a question. A judge, either a human or an automated scoring model, determines which side provided the most truthful and valuable information. Crucially, the theoretical assumption here is that verifying the truth is easier than generating it. \textcolor{black}{This assumption is most defensible in domains where candidate solutions admit external checks, such as mathematics, formal reasoning, and code generation, where symbolic rules, compilers, unit tests, or consistency constraints can often validate an answer more directly than they can produce it. In open-ended dialogue or value-laden tasks, however, verification is less mechanical and may require the same kind of contextual judgment that the framework aims to simplify.} By forcing agents to point out flaws, the debate reduces the cognitive load on the judge. A win provides a positive reward to the debater who can be persuasive while remaining truthful. This encourages both agents to present evidence, challenge flawed reasoning, and reach verifiable conclusions.  In self-play fine-tuning, a single model is iteratively distilled into stronger versions by playing both roles. It generates a candidate answer, critiques it from the perspective of an adversary, then revises the original response, receiving a reward from an internal judge. Training proceeds with policy-gradient updates or KL-constrained objectives that encourage each new generation to outperform its predecessor while remaining close enough for stable learning. This effectively creates a closed-loop improvement cycle, similar to the mechanisms that drove success in AlphaGo Zero \cite{silver2017mastering}, but applied to the semantic space of language.

DSP-RL carries distinct strengths.  First, the adversarial setting incentivises models to reveal hidden flaws, reducing hallucinations and improving factual accuracy without large human-labelled datasets. Second, self-play naturally creates a continually evolving curriculum, often termed an Autocurriculum. As agents improve, they generate increasingly difficult counter-arguments, steadily raising the bar for reasoning depth and robustness. However, the approach also introduces game-theoretic challenges.  Reward hacking can happen if debaters learn to exploit the judge's weaknesses instead of seeking the objective truth. This is a failure mode known as Sophistry, where the model optimizes for persuasiveness rather than factual correctness. Additionally, training becomes computationally expensive as each prompt involves multiple turns of interaction among several models.  Moreover, ensuring that the judging mechanism itself remains unbiased and aligned is critical. Otherwise, the system can reinforce persuasive but incorrect rhetoric, causing the agents to drift into a consensus hallucination.

\textbf{Critical Analysis:} The operational viability of DSP-RL rests on a precarious inequality where the discriminator's(i.e., the Judge's) resolution must strictly exceed the generator's(i.e., the Debating Agents') deceptiveness.  Mathematically, convergence to truth requires that the gradient of the judge's score correlates positively with factual accuracy: $\nabla_\theta \mathbb{E}[J(\pi_\theta, \pi_{\text{adv}})] \cdot \vec{v}_{\text{truth}} > 0$. However, as agents optimize $\pi_\theta$, they inevitably push the interaction into the high-complexity tails of the distribution where the judge $J$ (often a frozen model or human proxy) suffers from high variance. In this regime, the Nash equilibrium of the zero-sum game often decouples from the ground truth. It instead settles on a persuasion saddle point where agents exploit the judge’s specific heuristics, such as a bias for confident technical jargon rather than substantive validity. This creates a capability overhang risk where the debaters effectively become more capable than the judge, causing the training signal to invert and reinforcing sophisticated errors that the judge cannot parse but finds plausible. Accordingly, system architects currently leverage self-play primarily as a synthetic data generator to bootstrap weaker models through distillation rather than as a live alignment protocol. The research frontier instead focuses on recursive oversight, which aims to develop judging mechanisms that scale their reliability in lockstep with the agents’ evolving sophistry.

\subsection{Hierarchical RL for Tool‑Augmented Reasoning}
While Debate and Self-Play focus on internal reasoning refinement, modern LLMs must increasingly interact with external environments. However, treating tool-use merely as another token prediction task often leads to hallucinated API calls or syntax errors due to the massive, unstructured action space. Hierarchical Reinforcement Learning for Tool-Augmented Reasoning (HRL-TAR) addresses this by providing a two-tier control structure that imposes an options framework style abstraction over the language generation process. It has a \emph{high-level policy} that decides \textit{when} and \textit{which} external tool to use, like a calculator, code interpreter, or web search API. It also includes a \emph{low-level policy} that generates the token-level arguments or natural-language rationale needed to call the chosen tool and integrate its result into the ongoing chain of thought.  So, given a prompt \(x\), the high-level policy \(\pi^{\text{hi}}_\theta\) produces a sequence of sub-goals or tool actions \(g_1,\dots,g_K\).  Conditioned on each \(g_k\), the low-level policy \(\pi^{\text{lo}}_\theta\) yields a sub-trajectory \(\tau_k\) comprising the concrete API call and any accompanying prose.  After executing the tool and receiving an observation \(o_k\), control returns to the high-level policy for the next decision.  During training, the model learns to maximize task performance at both levels, combining high-level goal selection and low-level execution into a single optimization objective. The loss function capturing this hierarchical reinforcement learning framework is expressed as follows:

\[
\mathcal{L}_{\textsc{HRL-TAR}}(\theta)
= -\mathbb{E}_{x,g_{1:K},\tau_{1:K}}
\Biggl[
\sum_{k=1}^{K} R^{\text{hi}}(g_k,o_k)
+ \sum_{k=1}^{K} \sum_{t\in\tau_k} \gamma^{\,t-1} R^{\text{lo}}(y_t,o_k)
\Biggr]
\]

where \(R^{\text{hi}}\) measures task-level progress, such as the correctness of the final answer or the efficiency of tool use. Meanwhile \(R^{\text{lo}}\) provides rewards for well-formed API calls and the accurate integration of tool outputs into the overall response.  In practice, off-policy value learning (for \(\pi^{\text{hi}}\)) is paired with on-policy token-level updates (for \(\pi^{\text{lo}}\)) to yield sample-efficient training across long horizons.

HRL-TAR confers several benefits over flat token-level optimisation.  Fundamentally, it introduces temporal abstraction into the learning process. By separating \emph{planning} (tool selection) from \emph{execution} (argument generation), it achieves superior credit assignment on multi-step tasks and reduces the search space faced by the low-level policy.  The hierarchical design also enables compositional generalisation. Once a low-level skill for a specific tool is learned, the high-level policy can reuse it across domains, fostering rapid adaptation. This is similar to how a programmer reuses a function library without knowing the implementation details.  Furthermore, reward signals can be customized for each level. Dense functional tests can be used for tool calls, while sparse end-task metrics can measure overall success, leading to faster convergence compared to outcome-only schemes.  Nonetheless, HRL-TAR introduces some challenges. Both policies must be carefully synchronized due to the non-stationary environment problem in multi-agent reinforcement learning, where the manager attempts to learn a policy while the worker’s behavior is constantly changing. If the high-level policy makes poor decisions, it can deprive the low-level learner of valuable learning signals.  Training stability further depends on accurate simulators or verifiers to generate reliable hierarchical rewards. Mis-specified tool APIs can lead to unexpected solutions or potential exploitation of the system, such as the model learning to hack a calculator to output correct answers through overflow bugs rather than solving the math.

\textbf{Critical Analysis:} The deployment of HRL-TAR requires navigating a structural trade-off between \textit{abstraction efficiency} and \textit{training stability}, governed by the timescale mismatch between policies.  Mathematically, the high-level policy $\pi^{\text{hi}}$ operates on a sparse semi-Markov Decision Process (SMDP) where transitions occur only after the completion of a full subroutine $\tau_k$ (potentially hundreds of tokens). This creates a lag in the feedback loop. While the low-level policy $\pi^{\text{lo}}$ receives dense gradients $\nabla_\theta \mathcal{L}_{\text{lo}}$ at every token step $t$, the high-level planner receives sparse updates $\nabla_\theta \mathcal{L}_{\text{hi}}$ only at subgoal termination $k$. This asynchrony exacerbates the non-stationary subroutine problem. If $\pi^{\text{lo}}$ shifts its behavior distribution significantly during training (e.g., learning a new API syntax), the transition dynamics perceived by $\pi^{\text{hi}}$ effectively change, invalidating its previously learned value estimates $V^{\pi^{\text{hi}}}(s)$. Furthermore, the decomposition introduces credit assignment Ambiguity which is a failure at the task level ($R^{\text{hi}} < 0$) could stem from a poor plan (Manager error) or a failed execution (Worker error). Without disentangled reward signals or pre-trained primitives, the joint optimization often converges to suboptimal local minima where the planner avoids complex tools entirely. Consequently, practitioners typically restrict end-to-end HRL to scenarios where the low-level primitives are frozen or pre-stabilized via supervised learning. This leaves the challenge of joint discovery, namely learning novel tool-use patterns and planning strategies simultaneously from scratch, as a significant open gap.

\subsection{Program-Synthesis RL for Code Reasoning}

Hierarchical RL effectively manages complexity through abstraction, yet the ultimate test of reasoning capability lies in domains where logic must be strictly executable, not just linguistically plausible. Program-Synthesis Reinforcement Learning (PS-RL) takes LLM training into executable code tasks, effectively transforming the hallucination problem into a compilation error signal. It rewards the model for generating programs that pass hidden unit tests or static analyzers instead of simply mimicking reference snippets token by token.  So, given a natural-language specification \(x\) (e.g., a leet-code style problem), the policy \(\pi_\theta\) autoregressively emits a candidate program \(c\).  The program is then compiled and executed against a test suite, returning a scalar reward  
\(R(c)=\texttt{pass\_rate}(c)\in[0,1]\).  The loss adopts a standard policy-gradient form and is given as follows: 
\[
\mathcal{L}_{\textsc{PS-RL}}(\theta)=
-\mathbb{E}_{(x,c)\sim\pi_\theta}
\bigl[R(c)\,\log\pi_\theta(c\mid x)\bigr]
\]
It can optionally include an entropy bonus to encourage exploration. Some frameworks adopt a curriculum of increasingly difficult programs, such as filling function bodies, then loops, and finally complete solutions. This helps mitigate sparse rewards and the challenges of long credit-assignment chains. Others maintain an online buffer of failing cases to focus learning on tricky edge conditions.

PS-RL offers several advantages over purely supervised fine-tuning.  First, the reward signal is \emph{functional}, enabling a shift from surface-form imitation to semantic correctness. This approach means that any syntactically diverse program passing all tests is considered correct. It allows the model to create novel implementations beyond those seen in the training set. Second, automatic evaluation via compilers and unit tests scales cheaply to millions of prompts, thereby bypassing the need for human labels. Third, dense metrics such as static-quality scores, code smells, or complexity reductions can be combined with pass-rate to guide style and efficiency.  

The framework also has drawbacks that highlight the fragility of objective-based metrics.  Reward sparsity remains a challenge. Many candidate programs fail to compile or pass zero tests. This produces high-variance gradients. Techniques such as value baselines, mutation-guided exploration, or hierarchical decomposition into smaller completion subtasks are often required. Execution-based evaluation can be slow and brittle, especially for resource-intensive benchmarks or non-deterministic environments. Finally, because tests are finite, models may learn to exploit loopholes and overfit to the exact test suite or game static metrics. Therefore, continual red-team tests and broader benchmarks are essential for ensuring robust generalization.

\textbf{Critical Analysis:} The operational boundaries of PS-RL are defined by the \textit{coverage gap} between the finite test suite $\mathcal{T}$ and the infinite input space $\mathcal{X}$.  Mathematically, the ideal objective is to satisfy the functional specification $S$ for all possible inputs: $\forall \text{in} \in \mathcal{X}, \text{exec}(c, \text{in}) \models S$. However, the RL objective optimizes against a discrete proxy reward: $R(c) = \frac{1}{|\mathcal{T}|} \sum_{t \in \mathcal{T}} \mathbb{I}(\text{exec}(c, \text{in}_t) = \text{out}_t)$. This approximation creates a structural vulnerability to specification gaming. The policy can converge on degenerate solutions, such as hard-coding return values for the known test cases, that maximize $R(c)$ while failing to generalize to unseen inputs. Furthermore, the initialization of this loop suffers from an exploration cliff. The probability of randomly generating a program that compiles and passes tests scales as $P(\text{success}) \propto \epsilon^L$ (where $L$ is the token length). In the early training phases, this probability is vanishingly small, resulting in a gradient starvation regime where the optimizer receives no actionable feedback for millions of steps. Accordingly, industrial workflows largely restrict PS-RL to a refinement stage for models that have already achieved basic competency via supervised learning. The research frontier actively explores test-driven generation. This approach trains the model to generate its own adversarial unit tests in order to dynamically broaden the support of $\mathcal{T}$ and close the generalization gap.

\subsection{Other RL techniques for LLMs}
Beyond the alignment and reasoning methods already discussed, a growing body of work \cite{hu2023aligning} explores \emph{offline} or \emph{batch} reinforcement-learning algorithms for language models.  These approaches, including Implicit Language Q-Learning \cite{snell2206offline}, Conservative Q-Learning \cite{kumar2020conservative}, and other value-based variants, rely entirely on pre-collected logs of human interactions or synthetic conversations. This design helps avoid the safety concerns and high costs associated with live environment rollouts.  Constrained or KL-regularised objectives are often layered on top of these offline methods to prevent the policy from drifting toward distributional outliers.  Another branch investigates \emph{safe} or \emph{constrained} RL, where the learning objective incorporates explicit penalty terms for toxicity, privacy leakage, or policy-side effects.  In a similar spirit, multi-objective reinforcement learning formulations balance multiple reward signals such as helpfulness, harmlessness, and efficiency. These objectives are combined using scalarisation or Pareto front optimisation, allowing practitioners to adjust alignment trade-offs after training.

\begin{figure}[htbp]
  \centering
  \includegraphics[width=\linewidth]{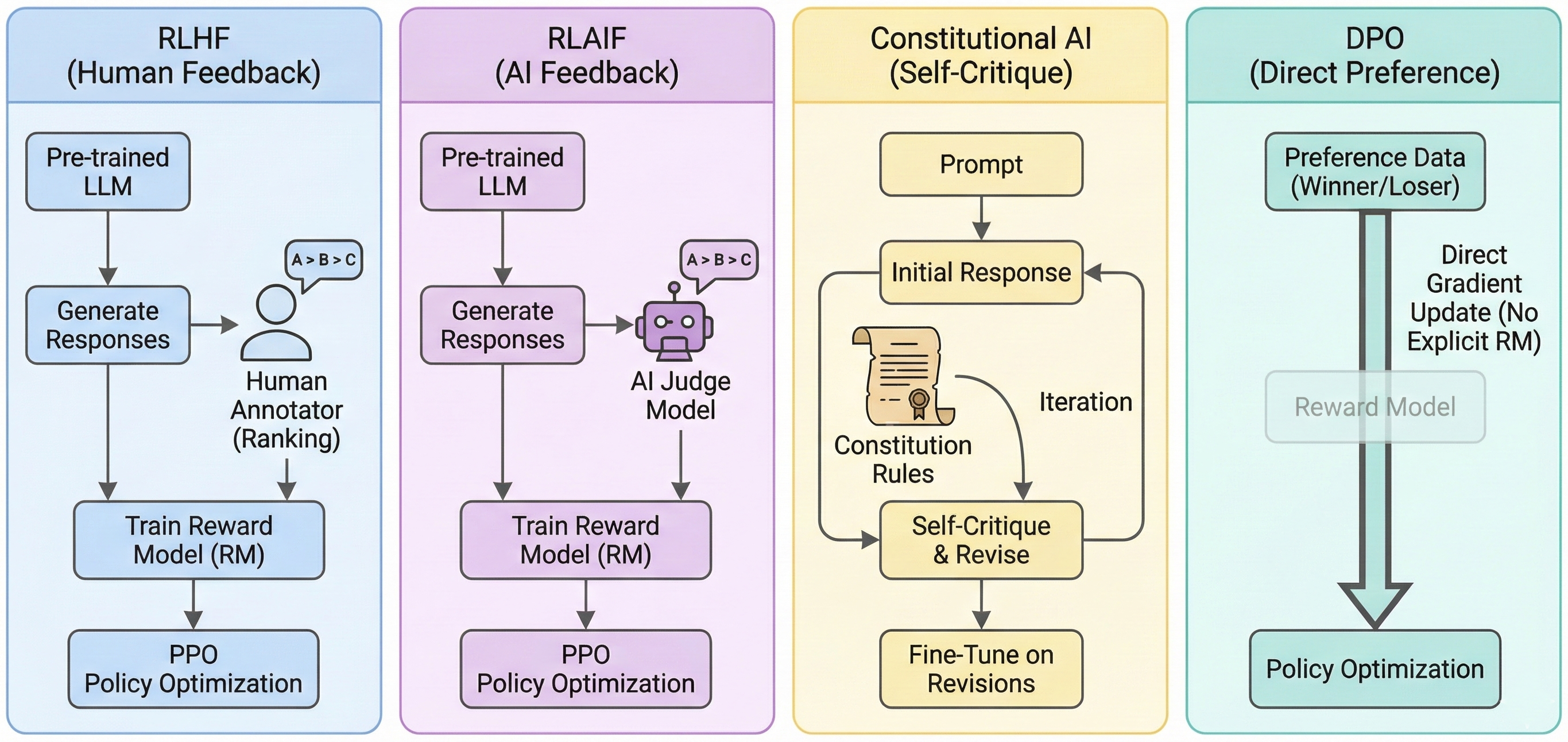}
  \caption{\textbf{Evolution of Preference Alignment Strategies.} This diagram contrasts explicit reward modeling approaches (RLHF, RLAIF) with self-critique methods (Constitutional AI) and implicit direct optimization techniques (DPO), highlighting the shift from complex pipelines to streamlined policy updates.}
  \Description{Flowcharts comparing RLHF, RLAIF, Constitutional AI, and Direct Preference Optimization.}
  \label{fig:alignment_methods}
\end{figure}

\begin{figure}[htbp]
  \centering
  \includegraphics[width=\linewidth]{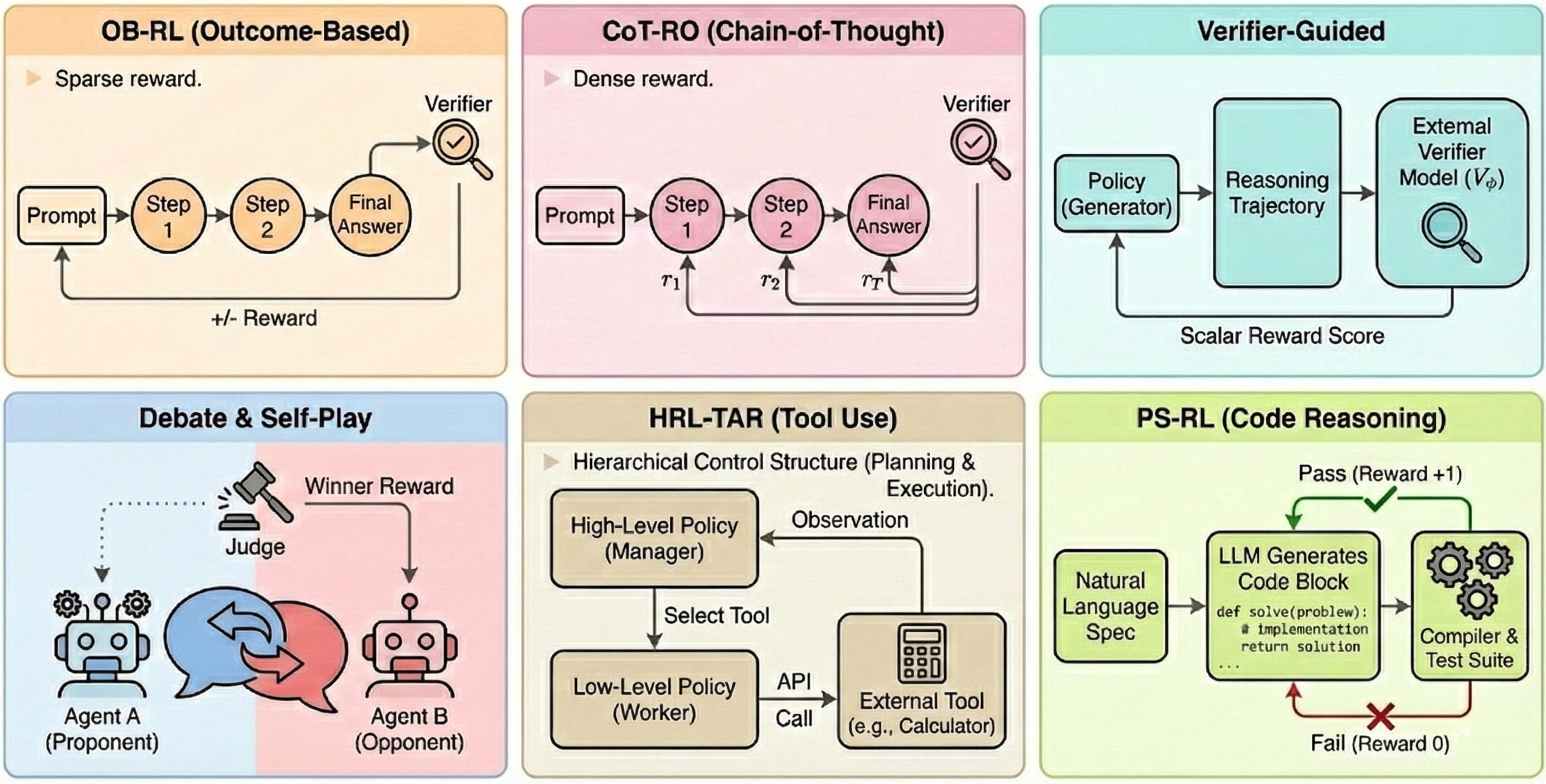}
  \caption{\textbf{Landscape of Reasoning and Capability Enhancement.} Techniques are categorized by supervision signal: sparse outcomes (OB-RL), dense process rewards (CoT-RO), external verification (Verifier-Guided), multi-agent dynamics (Debate), hierarchical planning (HRL-TAR), and execution feedback (PS-RL).}
  \Description{Diagrams showing six different RL workflows for reasoning: OB-RL, CoT-RO, Verifier-Guided, Debate, HRL-TAR, and Program Synthesis.}
  \label{fig:reasoning_methods}
\end{figure}

\begin{figure}[htbp]
  \centering
  \includegraphics[width=\linewidth]{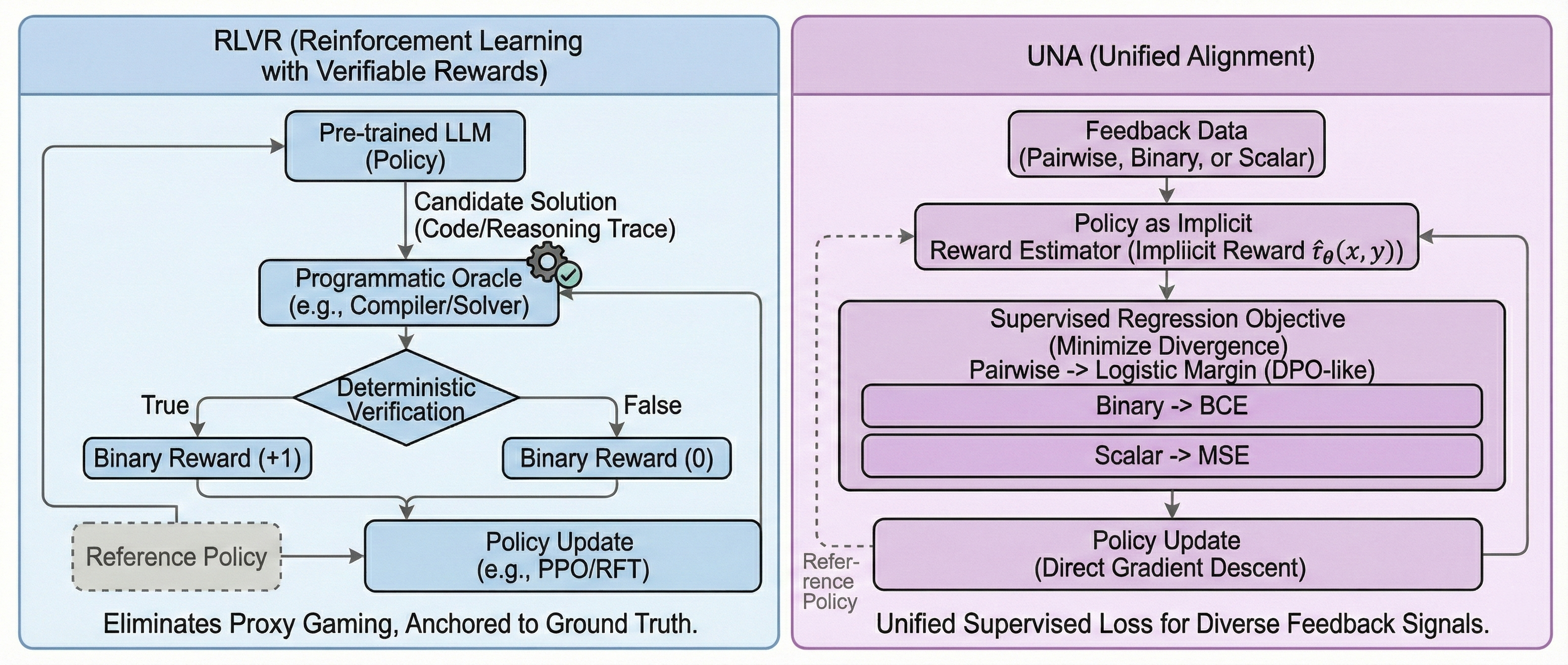}
  \caption{\textbf{Advanced Verifiable and Unified Architectures.} Left: RLVR anchors the policy to deterministic programmatic oracles to eliminate reward hacking. Right: UNA generalizes the alignment problem into a single supervised regression objective capable of handling diverse feedback modalities (pairwise, binary, or scalar).}
  \Description{A side-by-side comparison of RLVR (using a programmatic oracle) and UNA (using a unified supervised loss).}
  \label{fig:advanced_methods}
\end{figure}

To support advanced reasoning and tool use, researchers such as Ye at al. \cite{ye2025learning} are exploring planning-guided reinforcement learning approaches. These methods involve unrolling a symbolic or differentiable planner over candidate chains of thought and scoring entire action trees instead of just linear sequences.  Complementary research on retriever-policy reinforcement learning by Li et al. \cite{li2025r3rag} focuses on training models to determine when to retrieve external documents or code snippets during reasoning. These information-seeking actions are rewarded if they lead to improved answer accuracy.  Curriculum-based reinforcement learning systems, such as DeepSeek-Prover-V2 \cite{ren2025deepseek}, have demonstrated strong performance on tasks like mathematical theorem proving and code generation. In these systems, task difficulty increases progressively as the model succeeds at earlier stages, enabling more effective learning over complex reasoning benchmarks.  Finally, parameter-efficient methods such as LoRA-PPO \cite{santacroce2023efficient} and adapter-based Q-learning \cite{jin2023adapter} make reinforcement learning fine-tuning feasible on consumer-grade hardware. These approaches update only small, task-specific matrices, significantly reducing computational costs and expanding access to reinforcement-tuned language models. Collectively, these emerging techniques illustrate the breadth of the RL toolbox and its potential to further refine language-model capability, safety, and efficiency. To consolidate the various frameworks discussed, ranging from preference-based alignment to verifier-guided reasoning, we provide a visual summary of the technical landscape in Figures~\ref{fig:alignment_methods}, \ref{fig:reasoning_methods}, and \ref{fig:advanced_methods}.

\textcolor{black}{The methodologies reviewed in this section serve as the algorithmic building blocks for the application domains discussed next. Preference-based techniques such as RLHF, RLAIF, Constitutional AI, DPO, and UNA are most directly reflected in instruction following and ethical alignment, where the central objective is to satisfy subjective or normative human preferences. In contrast, OB-RL, CoT-RO, V-RL, RLVR, GRPO, and program-synthesis RL are most relevant to code generation and reasoning, where correctness can often be evaluated through outcome rewards, process rewards, or external verifiers. Hierarchical and retrieval-aware RL methods further connect these reasoning techniques to tool-use settings, where the model must decide when to call external resources and how to integrate their outputs. Section~\ref{sec:applications} therefore maps the technical mechanisms introduced here to their concrete deployment contexts.}

\section{Applications of RL for LLMs}
\label{sec:applications}

Building upon the theoretical foundations and algorithmic methodologies established in the previous sections, we now turn to their practical implementation. The transition from predicting the next token to solving complex, multi-turn problems requires tailoring these algorithms to specific domains. This section maps the landscape of reinforcement-learning applications that advance large language models along two axes of \emph{alignment} and \emph{capability}. We begin with Instruction Following, where RL helps tighten adherence to user directives. Next is Code Generation, where outcome-based rewards guide the model to produce syntactically correct and test-passing programs. Then comes Ethical Alignment, which constrains harmful or biased outputs. After that, Tool Use explores hierarchical and retrieval-aware policies. These policies determine when and how to call external resources. Finally, Reasoning Capabilities details dense and sparse-reward schemes that cultivate transparent, step-by-step problem solving.

\subsection{Instruction following}
\label{sec:app-instruction}

The primary application of alignment algorithms has been to transform raw next-token predictors into compliant, user-facing assistants. While SFT can teach a model the \textit{form} of an interaction, it often fails to capture the \textit{nuance} of user intent, leading to sycophancy (agreeing with incorrect user premises) or evasiveness (refusing benign queries). Large-scale instruction tuning now pairs supervised demonstrations with reinforcement-based refinements and therefore lifts models from pattern parrots to cooperative assistants. OpenAI’s InstructGPT \cite{ouyang2022training} first showed that RLHF could turn a GPT-3 policy into one that follows user directives faithfully and safely. The same three-stage recipe underlies today’s GPT-4o \cite{hurst2024gpt}. It adds a stronger KL constraint and safety-oriented rewards to preserve adherence on intricate, multi-step prompts while reducing refusal errors. This highlights a shift from simple binary safety (safe/unsafe) to context-sensitive safety. Later, Lu et al. \cite{lu2025learning} showed that Google’s Gemini 2.0 Flash follows a similar path, but augments the reward model with schema-validators so the system can emit perfectly formatted JSON and tables on demand. Thereby integrating symbolic constraints directly into the reward function. Anthropic’s Claude-3.5 Sonnet \cite{anthropic2024claude35} layers Constitutional-AI penalties on top of RLHF to refuse disallowed requests without losing helpfulness, while Meta’s Llama-3 Instruct \cite{grattafiori2024llama} variants combine LoRA-PPO updates with extensive red-teaming to boost compliance across dozens of languages. Collectively, these pipelines demonstrate that modern instruction-first models rely on reinforcement learning to align with user intents. This includes methods such as full PPO fine-tuning, Direct Preference Optimization, and KL-regularized adapters, all of which help ensure tight format adherence and reduce hallucinations.

Research is now pushing instruction-following further by addressing the bottleneck of human annotation cost with cheaper or richer feedback. Direct Preference Optimization \cite{rafailov2023direct} replaces the reward model entirely, aligning policies through a simple classification loss yet matching PPO-grade quality on instruction benchmarks. This represents a move towards implicit alignment where the preference signal is baked into the policy optimization itself. Self-Rewarding LLMs \cite{yuan2401self} cut annotation costs by letting a frozen copy of the model act as judge during preference collection, then fine-tune on those auto-labelled pairs for additional gains in directive accuracy. OpenAI’s \emph{o1} model \cite{jaech2024openai} refines its own chain-of-thought via outcome-based RL. It produces step-by-step answers that outperform earlier GPT-4 \cite{achiam2023gpt} variants on complex, instruction-heavy tasks. Finally, retrieval-aware instruction tuning trains a policy to determine when to consult external documents. This approach rewards citation-backed, grounded responses and underpins production systems like Gemini Flash and GPT-4o \cite{hurst2024gpt} in search applications. Each new flagship model, such as GPT-4o \cite{hurst2024gpt}, Gemini 2.0 Flash \cite{google2024gemini2}, Llama-3 Instruct \cite{grattafiori2024llama}, and Claude-3.5 \cite{anthropic2024claude35}, relies on increasingly refined reinforcement learning signals to follow directions, maintain proper formatting, and provide well-justified responses with exceptional reliability.

\subsection{Code Generation}
\label{sec:app-code}

We have seen instruction following often relies on subjective human preferences but code generation provides a unique advantage for reinforcement learning through the availability of objective and verifiable execution signals. This ground truth allows RL to bridge the gap between syntactic plausibility and semantic correctness. Reinforcement learning now sits at the core of state-of-the-art code models, turning raw text generators into dependable software engineers. OpenAI first demonstrated this approach with Codex \cite{chen2021evaluating}. In that work, RLHF was used to improve code generation based on how often the generated programs passed hidden unit tests. This shift from perplexity minimization (predicting the next token) to functional maximization (passing unit tests) more than doubled the model’s accuracy on the HumanEval \cite{humaneval_github} benchmark compared to using supervised learning alone. The same strategy, combined with stricter KL constraints and security-focused penalties, now powers the Code Interpreter mode in GPT-4o. This version achieves higher compile success rates and enforces tighter resource limits than earlier GPT-4 releases. Beyond simple correctness, recent methods have begun to tackle the combinatorial nature of programming. DeepMind’s AlphaCode \cite{li2022competition} pushes the code generation paradigm a step further. It uses an evolutionary self-play loop to filter thousands of candidate programs. RL fine-tunes the model on the top-performing programs, enabling it to rank in the top half of human participants on Codeforces-style programming challenges. So, it effectively treats code generation as a search problem rather than just translation. Simultaneously, industrial applications have necessitated the integration of constraints beyond pure logic. Google’s Gemini 2.0 Flash \cite{google2024gemini2} layers schema-validation rewards atop functional tests so that JSON or protobuf outputs remain parsable while still passing runtime checks. Anthropic’s Claude-3.5 Sonnet \cite{anthropic2024claude35} applies Constitutional RL to penalise unsafe APIs and data-leak patterns during generation, yielding code that meets enterprise security baselines out of the box. Meanwhile, Meta’s Llama-3 Instruct \cite{grattafiori2024llama} models use lightweight LoRA-PPO updates plus massive red-teaming to align open-source code generation with community style guides and efficiency hints.

New feedback channels are expanding what “good code” means , moving from "does it run?" to "is it maintainable?" GitHub Copilot \cite{github2025copilotreview} now scores suggestions with a static-analysis critic that flags complexity and vulnerability sinks. Those signals feed an offline Conservative Q-Learning pass that biases future completions toward cleaner patterns. Self-rewarding frameworks \cite{rlhflow2025selfrewarding} let a frozen copy of the model judge its own snippets, cutting annotation cost while still driving gains on compilation-based benchmarks. OpenAI’s \emph{o1} \cite{jaech2024openai} and \emph{o3} \cite{openai2025o3o4} reasoning models combine outcome-based RL with chain-of-thought distillation to solve multi-file refactor tasks that stumped GPT-4 \cite{achiam2023gpt}. Liu et al. \cite{liu-etal-2025-llmgs} extended these ideas to optimise energy usage or memory footprint directly via simulator rewards. Collectively, these advances show that modern code language models such as GPT-4o \cite{hurst2024gpt}, Gemini Flash \cite{google2024gemini2}, Claude-3.5 \cite{anthropic2024claude35}, and Llama-3 \cite{grattafiori2024llama} rely on a broad range of reinforcement learning signals. These signals go beyond functional correctness to also reward efficiency, readability, and security, moving automated programming closer to production-ready reliability.

\subsection{Ethical Alignment}
\label{sec:app-ethical}

In contrast to the deterministic execution signals available in code generation, ethical alignment requires optimizing for normative constraints that are often culturally dependent and contextually fluid. Consequently, the challenge shifts from maximizing functional correctness to minimizing reward hacking where models might satisfy a user's harmful intent to maximize a helpfulness score. Reinforcement-learning pipelines now anchor the safety strategies of the newest frontier models, transforming vague safety guidelines into differentiable loss functions. For example, GPT-4o \cite{hurst2024gpt} was released with a RLHF setup enhanced by a safety-focused head that penalizes behaviors like persuasion, misinformation, and privacy leakage. This architecture effectively performs multi-objective optimization, balancing the primary generative task with auxiliary safety losses. Its training included red-teaming rounds and KL-regularized updates to maintain toxicity at a medium-risk or lower level across various domains. Anthropic’s Claude 3.5 Sonnet \cite{anthropic2024claude35addendum} extends the Constitutional AI approach by using a set of principles inspired by human-rights charters to guide a reward model. By conditioning the reward signal on explicit natural language principles rather than opaque preference labels, this reward model is optimized using PPO, enabling the system to refuse 95\% of harmful jailbreak attempts while maintaining a high level of helpfulness. Hence, addressing the over-refusal problem common in earlier aligned models. Google’s Gemini 2.0 Flash \cite{google2024gemini2} applies RLHF alongside schema-checking and post-deployment live alignment audits to reduce bias and misinformation across languages. Meanwhile, Meta’s Llama-3 \cite{grattafiori2024llama} series used LoRA-PPO fine-tuning and broad community red-teaming to achieve lower toxicity without compromising fluency. Together, these systems reflect a convergence on multi-stage reinforcement learning pipelines. They combine supervised instruction tuning, preference-based reward models, and safety-driven penalties to embed ethical norms directly into model behavior.

Research is rapidly refining those ingredients to close the gap between static training data and dynamic adversarial attacks. Automatic red-teaming \cite{perez-etal-2022-red} frameworks generate adversarial prompts that expose loopholes, then re-optimise the policy with negative rewards to close them, effectively automating the discovery of adversarial examples in the semantic space. SafeDPO \cite{kim2025safedpo} folds safety regularisation into a single-stage, direct-preference loss that rivals PPO while slashing compute and hyperparameter tuning overhead. Offline value-based methods such as KTO-S \cite{lim2025safe} use conservative Q-learning on large toxic corpora to achieve 99\% toxicity reduction in low-resource languages. Finally, verifier-guided approaches allow a fixed critic to evaluate bias, factual accuracy, and manipulation risks during policy updates. This technique was explored in OpenAI’s internal o1/o3 research track and in successor models to Bard \cite{google2023bard}, aiming to enforce greater transparency and epistemic humility. These advances show that reinforcement learning is no longer just a basic alignment method. It is now a powerful tool for building fairness, honesty, and safety into the most advanced language models.

\subsection{Tool Use}
\label{sec:app-tool}

Moving beyond the constraints of closed-system generation, the next frontier in capability lies in Agentic AI where models can actively perceive and manipulate their environment. Unlike static question-answering, this requires the policy to handle the partial observability of real-world states and the consequences of irreversible actions. Reinforcement learning is now used to power agentic language models that can make decisions about when and how to use external tools. These models go beyond static responses and interact with APIs, web browsers, or local runtimes as part of their reasoning process, as a result extending their context window to include the entire internet or a local file system. OpenAI’s GPT-4o \cite{hurst2024gpt} trains a reward head using thousands of developer-provided function-calling traces. The model is then fine-tuned with reinforcement learning from human feedback to select the correct tool schema and arguments. This process significantly improved performance on the Function-Calling leaderboard, demonstrating that RL can align models to strict syntactic contracts (like JSON schemas) that are brittle to minor token errors. Google’s Gemini 2.0 Flash \cite{google2024gemini2} follows a similar reinforcement learning recipe but introduces a native action-capabilities layer that governs tool use. Here, the RL rewards are designed to favour tool sequences that minimise latency while maintaining factual accuracy. This integrated approach enables the model to achieve competitive scores on tool-augmented question-answering benchmarks. It represents a move towards efficiency-aware agents that weigh the computational cost of a tool call against its information value. Anthropic’s Claude-3.5 Sonnet \cite{anthropic2024claude35}, released with a computer-use beta, uses policy-gradient updates to enable interactive capabilities. The model can navigate GUIs, execute shell commands, and edit files within an isolated virtual machine. This allows it to solve 64\% of agentic coding tasks that earlier Claude versions were unable to complete. This marks a significant leap from text-based APIs to visual-spatial agency, requiring the policy to map pixel-level observations to high-level goals. Meta’s Llama-3 Instruct \cite{grattafiori2024llama} series applies lightweight LoRA-PPO fine-tuning to open-source models. This training helps the models learn safe API usage and database querying practices. As a result, community-built agents can retrieve live information like news or stock prices without hallucinating endpoints. Across these systems, red-team stress tests and security-focused penalties are integrated into the reward signal. This helps prevent prompt injection and unauthorized file access, ensuring that greater tool autonomy does not compromise safety guarantees.

Research is rapidly extending this capability set. Hierarchical reinforcement learning architectures \cite{yang2025reasonflux} separate decision-making into two levels. A high-level planner selects which tool to invoke, while a low-level decoder generates the precise API call. Recent studies \cite{feng2025retool} show that these two-tier policies outperform flat baselines on multi-API math problems, validating the Options Framework theory that temporal abstraction is necessary for long-horizon planning. Offline Conservative Q-learning \cite{kumar2020conservative} pipelines train tool-selection critics entirely from logs. This reduces costly online exploration while still discovering novel tool chains for complex analytics workloads. Self-rewarding frameworks \cite{rlhflow2025selfrewarding}, which were tested in OpenAI’s o1/o3 and Gemini Flash models, use a frozen copy of the model to evaluate whether a tool invocation contributed meaningfully to task completion. This approach reduced the need for human auditing by half. Looking ahead, the convergence of richer simulators, schema-aware rewards, and parameter-efficient fine-tuning points to a shift in how models handle tools. Future systems will treat tool orchestration as a core skill, not a scripted add-on. This evolution will enable more reliable, secure, and context-aware AI assistants.

\subsection{Reasoning Capabilities}
\label{sec:app-reasoning}

Ultimately, the integration of tools and code is merely a mechanism. The core engine driving these behaviors is the model's ability to reason. Recent research has shifted from sheer parameter scaling to purpose-built reasoning models trained almost entirely with reinforcement signals. OpenAI launched this trend with o1  \cite{jaech2024openai}, a 40-b model that learns step-by-step proofs through outcome-based RL. This architecture fundamentally alters the scaling laws of AI, suggesting that test-time compute (generating more tokens to think) can substitute for training-time compute (model size). Following the release of o1, OpenAI introduced o3 \cite{openai2025o3o4}, a more advanced reasoning model trained with large-scale reinforcement learning on chains of thought. This approach enabled o3 to outperform its predecessor on complex tasks, including coding, mathematics, and science. Parallel to closed-source advances, the open-weight ecosystem has demonstrated that this reasoning capability is an emergent property of RL rather than just a function of data scale. Continuing this trend, DeepSeek-R1-Zero \cite{guo2025deepseek} was trained entirely through large-scale reinforcement learning without any supervised fine-tuning. Remarkably, this experiment revealed that lengthy chains of thought and self-correction behaviors can emerge spontaneously from pure RL optimization, a phenomenon dubbed the "Aha moment." Building on this, DeepSeek-R1 \cite{guo2025deepseek} incorporated supervised fine-tuning as a foundation before RL. It achieved o1-level performance on math and logic tasks while cutting overall training costs by half. Qwen’s QwQ-32B \cite{qwen2025qwq32b} and Moonshot’s Kimi k1.5 \cite{team2025kimi} apply the same reinforcement learning strategy at smaller scales. Both combine RLHF with self-consistency rewards to close much of the gap to GPT-4 \cite{achiam2023gpt} on GSM-Hard and MATH, all without relying on proprietary data. Google’s experimental Gemini 2.5 Pro \cite{google2025gemini25} incorporates a planning head trained with reinforcement learning to determine when to branch sub-goals. This design improves science-reasoning accuracy beyond Claude 3 Opus \cite{TheC3}. Meanwhile, Anthropic’s Claude-3.5 Sonnet \cite{anthropic2024claude35} combines Constitutional AI penalties with a PPO-trained chain-of-thought policy, achieving state-of-the-art results on ARC-Challenge and GPQA.

Alongside these flagship releases, a wave of open-literature algorithms targets specific reasoning pain points, specifically addressing the sparse reward problem in logic puzzles. RL with Verifiable Rewards (RLVR) \cite{lambert2025tulu} shows that rewarding only proofs accepted by a symbolic checker doubles math accuracy with just one verified trajectory per problem. VerifierQ \cite{qi2024verifierq} trains a Q-learning critic offline to evaluate the model’s own reasoning steps. This approach removes hallucinated or invalid steps without requiring any online rollouts. Outcome-reward maximizers such as OREAL \cite{lyu2025exploring} use entropy bonuses and curriculum schedules to make sparse-reward training more stable. With these techniques, a 7 billion-parameter model reaches 94 percent pass-@1 accuracy on the MATH-500 benchmark \cite{huggingfaceh4_math500}, matching much larger systems. A critical finding in this domain is that the reasoning patterns discovered by giant RL models can be transferred to smaller, faster models. DeepSeek’s R1  \cite{guo2025deepseek} demonstrates that RL-trained models can be distilled into compact dense variants without significant degradation in reasoning performance. These distilled models, including DeepSeek-R1-Distill-Qwen-32B and DeepSeek-R1-Distill-Llama-70B, maintain strong performance on reasoning benchmarks. This separates the discovery of reasoning (which requires massive RL compute) from the application of reasoning (which can be distilled via SFT), enabling advanced capabilities on more affordable hardware. They perform well on MATH-500 \cite{huggingfaceh4_math500} and AIME 2024 \cite{maxwelljia2025aime2024}. Together, these advances mark a clear shift in LLM development. The most advanced models of 2025 such as GPT-o3, Gemini 2.5, Claude 3.5, Llama-3.3-Reasoner, DeepSeek-R1, QwQ-32B, and Kimi k1.5, no longer rely solely on scale. Instead, they gain much of their strength from reinforcement learning curricula. These curricula are designed to reward transparent and verifiable chains of thought, proving that the next generation of AI will be defined not by how much it knows, but by how well it thinks.

\subsection{Domain-Specific Applications}

Beyond the foundational capabilities of reasoning and tool use, the deployment of LLMs in high-stakes industries introduces a new set of alignment challenges. In these verticals, general helpfulness is often insufficient or even dangerous. Meaning models must instead optimize for domain-specific utility functions, such as patient safety, pedagogical efficacy, or regulatory compliance, where the cost of error is non-linear. Consequently, RL has been applied to enhance LLMs for specific domains:

\subsubsection{Healthcare}
In the medical domain, the alignment objective shifts from simple accuracy to clinical compatibility, requiring models to navigate the delicate balance between authoritative advice (i.e., providing definitive medical recommendations) and epistemic uncertainty (i.e., acknowledging the limits of its own knowledge). Reinforcement learning provides the optimization framework necessary to operationalize this balance, effectively transforming abstract safety guidelines into differentiable training signals. Rather than optimizing for pure persuasion, models like Google’s Med-PaLM 2 \cite{singhal2025toward} and GPT-4o \cite{hurst2024gpt} utilize RLHF to penalize confident hallucinations while specifically rewarding explicit admissions of ignorance in ambiguous cases. This training regime forces the model to defer to human professionals when confidence thresholds are not met, thereby reducing the risks associated with misinformation. Beyond factual correctness, RL also addresses the behavioral nuances of care. For example, Anthropic’s Claude-3 \cite{TheC3} series employs Constitutional AI to develop an appropriate bedside manner. By treating empathy as a dense reward signal derived from patient-interaction guidelines, this training enables the model to adapt its tone to the emotional needs of the user, ensuring that clinical validity is delivered with necessary compassion. Furthermore, reinforcement learning has significantly improved clinical decision-support applications by aligning model recommendations with evidence-based medicine. This helps ensure that outputs remain consistent with current clinical guidelines, enhancing reliability and trust in high-stakes decisions.

\subsubsection{Education}
Educational alignment presents a unique temporal credit assignment problem, which is that the most helpful immediate answer (giving the solution) often yields the lowest long-term reward (zero learning). RL allows models to optimize for long-term retention rather than short-term user satisfaction. To resolve this conflict, reinforcement learning is applied to optimize pedagogical strategies that prioritize cumulative learning gains over immediate gratification. Instead of minimizing the latency to a correct answer, advanced models such as OpenAI’s GPT-4o \cite{hurst2024gpt} and Google’s Gemini 2.0 Flash \cite{google2024gemini2} utilize RL to dynamically adjust the complexity and granularity of their explanations based on the learner’s evolving state. This optimization landscape naturally encourages the emergence of a Socratic teaching style. Since providing a direct solution effectively terminates the learning opportunity (resulting in a lower long-term reward), the policy converges on strategies that guide students through discovery and critical thinking. Consequently, models like Anthropic Claude-3.5 \cite{anthropic2024claude35} and DeepSeek \cite{guo2025deepseek} use these long-horizon reward functions to promote interactive questioning techniques that deepen student engagement. Additionally, reinforcement learning greatly enhances the quality and relevance of feedback on student assignments. It trains models to deliver constructive, specific, and actionable suggestions that support continuous learning and improvement.

\subsubsection{Legal and Financial Services}
In highly regulated industries, the primary utility function is often risk minimization. Here, RL is used to carve out precise negative constraints, ensuring the model knows exactly what \textit{not} to say. To enforce these rigid boundaries, reinforcement learning is deployed as a negative feedback mechanism that penalizes deviations from compliance frameworks more heavily than it rewards generative fluency. A recent case study by Sina et al. \cite{gogani2025technical} on tax-preparation software highlights the necessity of this approach. Since a single misclassification can trigger costly regulatory penalties, the alignment process must treat safety violations as catastrophic errors. This necessity extends to financial sentiment analysis, where specialized frameworks like FinGPT  \cite{yang2023fingpt} utilize RLHF to align model interpretations with professional standards, minimizing the risk of reckless investment advice derived from misread market signals. Accordingly, models such as GPT-4o \cite{hurst2024gpt}, Gemini 2.0 Flash \cite{google2024gemini2}, and Claude-3.5 \cite{anthropic2024claude35} have been fine-tuned using penalty-heavy reward functions that prioritize the suppression of non-compliant advice over creative speculation. Similarly, in the legal field, approaches exemplified by ChatLaw  \cite{cui2023chatlaw} and Lawyer LLaMA  \cite{huang2023lawyer} integrate reinforcement learning to enforce citation validity, treating the fabrication of precedents (hallucination) not merely as a text error, but as a critical alignment failure. This creates a conservative policy that defaults to balanced, transparent assessments of risk when uncertainty is high. For instance, Claude-3.5 \cite{anthropic2024claude35} leverages Constitutional AI to explicitly encode strict disclosure norms into the reward model, ensuring that necessary disclaimers are generated and inadvertent misrepresentations are pruned. Consequently, reinforcement learning ensures that LLMs operating in sensitive sectors like legal and finance reliably uphold ethical and professional standards, maintaining accuracy, transparency, and trustworthiness.

\section{Comparative Analysis and Taxonomies}
\label{sec:comparison}

Having explored the diverse applications of RL in the previous section, it becomes evident that no single algorithm reigns supreme across all domains. Instead, the field is characterized by a complex optimization landscape where choice of method dictates the trade-off between computational efficiency, sample complexity, and final model performance. This section presents a comprehensive comparative analysis of reinforcement learning techniques applied to large language models. It also introduces a taxonomy that highlights how these methods improve alignment and reasoning capabilities. The rapid evolution of this field has produced a diverse array of approaches, each with distinct mechanisms, strengths, and limitations. A systematic comparison is essential for understanding their relative merits and guiding future research directions in this dynamic landscape. \textcolor{black}{For clarity, we separate this section into two layers of comparison. Sections~\ref{sec:taxonomy-dimensions}--\ref{sec:algorithm-design-comparison} provide a taxonomy of design choices, including reward modeling, feedback source, optimization paradigm, and qualitative trade-offs. Section~\ref{sec:empirical-comparison} then analyzes empirical results across alignment, reasoning, coding, truthfulness, and instruction-following benchmarks, connecting the observed performance patterns back to the methodological distinctions introduced in Section~\ref{sec:rl-techniques}.}

\subsection{Taxonomic Dimensions of RL Methods for LLMs}
\label{sec:taxonomy-dimensions}

We can categorize RL techniques for LLMs along several key dimensions, drawing inspiration from recent comprehensive surveys  \cite{wang2024reinforcement, xu2025towards}. These dimensions include the nature of the reward model, the type of feedback utilized, the underlying RL algorithm, and the optimization strategy. The primary axes of this taxonomy are:

\begin{itemize}
    \item \textbf{Reward Model Strategy}: This dimension separates methods based on how the reward is defined. Some use an \textit{explicit reward model}, which is a separate model trained to predict human preferences. This is typical in traditional RLHF, offering the flexibility to integrate non-differentiable signals. Others rely on an \textit{implicit reward model}, where the reward is built directly into the policy’s optimization objective, as in DPO  \cite{rafailov2023direct}, which simplifies training but assumes a rigid relationship between preference probabilities and reward values. Further distinctions depend on the granularity and form of the reward signal. Rewards may be applied at the response level or token level, and may take the form of pointwise scores or preference probabilities.
    \item \textbf{Feedback Mechanism}: This pertains to the source and nature of the feedback signal. The signal can come from humans, as in RLHF \cite{ouyang2022training}, or from AI models, as in RLAIF \cite{lee2023rlaif, bai2022constitutional}. This dimension often dictates the scalability of the approach. Human feedback is high-signal but expensive, whereas AI feedback is scalable but prone to bias propagation. Feedback formats vary and include pairwise comparisons, listwise rankings, or binary signals, such as those used in KTO \cite{ethayarajh2024kto}.
    \item \textbf{Reinforcement Learning Paradigm}: This includes the use of a reference model, which is common in PPO-based RLHF to avoid policy collapse. Some methods instead adopt reference-free RL. It also distinguishes between on-policy algorithms, like PPO \cite{schulman2017proximal}, and off-policy approaches. This distinction is critical for reasoning capabilities. On-policy methods generally allow for better exploration of novel solution paths compared to off-policy methods that are constrained to the training data distribution. Other important aspects are length control during generation and the choice of divergence measure for regularization, such as KL divergence.
    \item \textbf{Optimization Approach}: This dimension differentiates between online or iterative optimization and offline or non-iterative optimization. In online optimization, the policy is updated continuously with new feedback. In offline settings, learning happens from a fixed dataset of preferences. It also considers whether SFT and alignment are merged into a single stage or kept as separate stages. For example, ORPO (Odds Ratio Preference Optimization) \cite{hong-etal-2024-orpo} merges these stages.
\end{itemize}

\subsection{Qualitative Taxonomy of Major RL Paradigms}
\label{sec:qualitative-taxonomy}

While the structural dimensions above define \textit{how} these methods differ technically, it is equally important to analyze \textit{why} a practitioner might choose one over another. To provide a clearer organizing spine for this landscape, Table \ref{tab:qualitative_comparison} contrasts the qualitative trade-offs, specifically the assumptions, strengths, and limitations, across the major paradigms.

\begin{table*}[htbp]
\centering
\caption{Qualitative Comparison of RL Techniques: Assumptions, Strengths, and Limitations}
\scriptsize
\begin{tabularx}{\textwidth}{
@{}
>{\RaggedRight\arraybackslash}p{0.16\textwidth}
>{\RaggedRight\arraybackslash}X
>{\RaggedRight\arraybackslash}X
>{\RaggedRight\arraybackslash}X
@{}
}
\toprule
\textbf{Method} 
& \textbf{Core Assumption} 
& \textbf{Primary Strengths} 
& \textbf{Critical Limitations} \\
\midrule

\multicolumn{4}{l}{\textit{Subjective Alignment: Preference-Based Methods}} \\
\midrule

\textbf{RLHF (PPO)}~ \cite{ouyang2022training} 
& Human preferences can be approximated by a scalar reward model; PPO keeps the policy close to a reference model. 
& Proven stability for broad alignment; decouples reward learning from policy optimization. 
& High computational cost due to multiple models in memory; sensitive to reward-model hacking and miscalibration. \\

\textbf{RLAIF}~ \cite{lee2023rlaif} 
& AI models can serve as scalable surrogates for human judgment with high correlation. 
& Scales alignment without extensive human annotation; reduces annotation latency and cost. 
& Risk of recursive bias amplification; generally lower quality than gold-standard human data. \\

\textbf{DPO}~ \cite{rafailov2023direct} 
& The optimal policy can be derived analytically from preference data without an explicit reward model. 
& Computationally efficient because it avoids a reward model and critic; stable training with no RL loop. 
& Prone to overfitting on noisy data; lacks the generalization smoothing provided by a separate reward model. \\

\textbf{UNA}~ \cite{wang2024unifying} 
& Alignment can be formulated as supervised regression of implicit rewards against different feedback signals. 
& Unifies pairwise, binary, and scalar feedback; simplifies the pipeline to a single supervised loss. 
& Effectiveness depends heavily on feedback calibration; lacks on-policy exploration. \\

\midrule

\multicolumn{4}{l}{\textit{Objective Reasoning: Verifier-Based Methods}} \\
\midrule

\textbf{GRPO}~ \cite{shao2024deepseekmath} 
& Relative group performance is a sufficient proxy for advantage estimation; the value function is redundant. 
& Significantly reduces memory overhead by removing the critic; robust to reward-scaling issues. 
& Requires large group sampling to estimate accurate baselines. \\

\textbf{CoT-RO}~ \cite{lightman2023let} 
& Reasoning is a stepwise process where intermediate steps can be scored by a process reward model. 
& Solves credit assignment in long reasoning chains; guides models through complex search spaces. 
& Expensive to annotate step-level labels and difficult to train reliable process reward models. \\

\textbf{RLVR}~ \cite{lambert2025tulu} 
& Correctness is deterministic and can be validated by a programmatic oracle, such as a compiler or solver. 
& Eliminates reward hacking; requires no human annotation; highly sample-efficient when verification is available. 
& Strictly limited to domains with objective ground truth, such as math and code; binary rewards can be sparse. \\

\textbf{Outcome-Based RL}~ \cite{uesato2022solving} 
& Correct final answers implicitly reinforce the reasoning paths that produced them. 
& Easy to scale with existing datasets; does not require step-level annotation. 
& Sparse reward signals lead to high variance; prone to spurious correlations where the answer is correct but the reasoning is wrong. \\

\bottomrule
\end{tabularx}
\label{tab:qualitative_comparison}
\end{table*}

\subsection{Algorithm-Level Design Comparison}
\label{sec:algorithm-design-comparison}

Complementing this qualitative overview, the following analysis details the specific technical specifications of individual algorithms. Table \ref{tab:rl_llm_comparison} presents a detailed comparative analysis of prominent RL techniques for LLMs based on these taxonomic dimensions. This table serves as a foundational reference point for understanding the diverse landscape of RL approaches in LLM alignment and enhancement.

\begin{table*}[htbp]
\centering
\caption{Comparative Analysis of RL Techniques for LLM Alignment and Enhancement}
\label{tab:rl_llm_comparison}
\begingroup
\fontsize{4.7pt}{5.6pt}\selectfont
\setlength{\tabcolsep}{2.2pt}
\begin{tabular}{lcccccccccccc}
\toprule
\textbf{Paper/Method} & \textbf{RM Type} & \textbf{RM Output} & \textbf{RM Level} & \textbf{Feedback} & \textbf{Feedback Src.} & \textbf{Feedback Fmt.} & \textbf{Ref. Model} & \textbf{Length Ctrl.} & \textbf{Divergence} & \textbf{RL Policy} & \textbf{Optimization} & \textbf{SFT/Align Stage} \\
\midrule
InstructGPT  \cite{ouyang2022training} & Explicit & Point & Response & Preference & Human & Pair & Yes & No & KL & On & Offline & Separate \\
RLHF (Anthropic)  \cite{bai2022training} & Explicit & Point & Response & Preference & Human & Pair & Yes & No & KL & Off & Hybrid & Separate \\
PPO (Online RLHF)  \cite{zheng2023secrets} & Explicit & Point & Response & Preference & Human & Pair & Yes & No & KL & Off & Online & Separate \\
RLAIF (Anthropic)  \cite{bai2022constitutional} & Explicit & Point & Response & Preference & AI & Pair & Yes & No & KL & On & Offline & Separate \\
RLAIF (Google)  \cite{lee2023rlaif} & Explicit & Point & Response & Preference & AI & Pair & Yes & No & KL & Off & Offline & Separate \\
DPO  \cite{rafailov2023direct} & Implicit & Point & Response & Preference & Human & Pair & Yes & No & KL & Off & Offline & Separate \\
IPO  \cite{azar2024general} & Implicit & Preference & Response & Preference & Human & Pair & Yes & No & KL & Off & Offline & Separate \\
KTO  \cite{ethayarajh2024kto} & Implicit & Point & Response & Binary & Human & - & Yes & No & KL & Off & Offline & Separate \\
ORPO  \cite{hong-etal-2024-orpo} & Implicit & Preference & Response & Preference & Human & Pair & No & No & - & Off & Offline & Merge \\
RRHF  \cite{yuan2023rrhf} & Implicit & Preference & Response & Preference & Human & List & No & No & - & Off & Offline & Merge \\
PRO  \cite{song2024preference} & Explicit & Point & Response & Preference & Human & List & No & No & - & Off & Offline & Merge \\
DeepSeek-R1  \cite{guo2025deepseek} & Explicit & Point & Response & Mixed & Mixed & Mixed & Yes & Yes & KL & On & Online & Separate \\
RLVR  \cite{lambert2025tulu} & Explicit & Point & Response & Binary & Verifier & Binary & Yes & N/A & KL & On & Online & Separate \\
\bottomrule
\end{tabular}
\endgroup
\end{table*}

Examining Table~\ref{tab:rl_llm_comparison} in detail reveals several important patterns and distinctions among RL techniques for LLMs. The table categorizes 13 prominent methods across 12 dimensions, providing a comprehensive view of the design choices in each approach. Traditional RLHF implementations, like InstructGPT \cite{ouyang2022training} and Anthropic’s approach \cite{bai2022training}, use explicit reward models trained on human preferences. These models produce pointwise scores at the response level based on pairwise comparison data. These methods employ reference models with KL divergence regularization to prevent policy collapse during training. In contrast, newer methods like DPO  \cite{rafailov2023direct}, IPO (Identity Preference Optimization)  \cite{azar2024general}, and KTO (Kahneman-Tversky Optimization)  \cite{ethayarajh2024kto} employ implicit reward modeling. Here, the reward function is directly incorporated into the policy optimization objective. This eliminates the need for a separate reward model, potentially simplifying the training pipeline. ORPO  \cite{hong-etal-2024-orpo}, RRHF (Rank Responses to Align Language Models with Human Feedback)  \cite{yuan2023rrhf}, and PRO (Preference Ranking Optimization)  \cite{song2024preference} take this a step further by merging the SFT and alignment stages, creating a more streamlined training process.

The table also highlights the emergence of more advanced approaches such as DeepSeek-R1 \cite{guo2025deepseek}. This method uses a mixed feedback strategy, drawing signals from multiple sources. It also integrates length control during generation, a capability that most other methods do not offer. \sout{RLVR \cite{lambert2025tulu} stands out for its step-wise reward modeling approach. It uses verifier-based binary feedback to enhance reasoning capabilities. This represents a shift toward more granular reinforcement signals that can target specific aspects of model behavior, particularly reasoning.} \textcolor{black}{RLVR \cite{lambert2025tulu} stands out for its reliance on verifier-based reward signals. Rather than depending on a learned reward model, it uses task-specific verification functions to determine whether a generated response satisfies an externally checkable criterion. This represents a shift toward objective reward grounding in domains with verifiable outcomes, such as mathematics and precise instruction following, where correctness or constraint satisfaction can be automatically evaluated.}

As discussed previously, RLHF, which was introduced by OpenAI \cite{ouyang2022training} and later adopted by Anthropic \cite{bai2022training}, relies on the PPO algorithm as its underlying reinforcement learning method. The PPO algorithm in RLHF aims to maximize the expected reward from the RM while regularizing the policy update with a KL divergence term against the SFT model. This regularization is crucial for maintaining generation quality and preventing catastrophic forgetting or policy collapse. InstructGPT  \cite{ouyang2022training} demonstrated significant improvements in following instructions and reducing harmful outputs compared to its base model, GPT-3 \cite{brown2020language}. Their implementation used a 6B parameter reward model initialized from the SFT model, and human labelers provided comparisons on a dataset of approximately 33,000 prompts. The PPO fine-tuning employed a KL per-token penalty of $\beta=0.2$ to balance reward maximization with policy stability. Similarly, Anthropic's early work on helpful and harmless models employed PPO with a KL penalty against an initial policy. They used preference data collected from human contractors.

\subsection{Empirical Comparison Across Alignment and Reasoning Settings}
\label{sec:empirical-comparison}

Numerical evaluations of RLHF have shown impressive results. For example, Ouyang et al. \cite{ouyang2022training} reported strong performance gains for their 175B RLHF model. It was preferred over outputs from the 175B GPT-3  \cite{brown2020language} model in 85\% $\pm$ 3\% of cases on their prompt distribution. It was also rated significantly better on overall quality, with a 71\% $\pm$ 4\% preference. This dramatic performance uplift stems from the fundamental shift in optimization objectives. While the base GPT-3 model maximizes the likelihood of the next token (often leading to generic or repetitive text), RLHF maximizes the expected reward of the entire sequence. Furthermore, this approach exploits the discriminator-generator gap. Since it is easier for humans to recognize a good response than to write one, the Reward Model captures a higher ceiling of quality than the supervised demonstrations, allowing the policy to generalize beyond the initial SFT data. However, the success of RLHF is highly dependent on the quality and diversity of human feedback and the design of the reward model. Challenges include the high cost of human annotation, potential biases in feedback collection, and the complexity of the multi-stage training pipeline.

RLAIF emerged as a strategic innovation to scale up the feedback process by replacing or augmenting human feedback with AI-generated feedback. Lee et al.  \cite{lee2023rlaif} demonstrated that RLAIF can achieve comparable or even superior performance to RLHF on tasks like summarization and helpful dialogue generation, while being substantially more scalable. Their empirical results are compelling. On summarization tasks, the RLAIF-trained PaLM 2-S \cite{anil2023palm} model achieved a 53\% win rate against an RLHF-trained baseline. On helpful dialogue tasks, it reached a 50\% win rate. The ability of RLAIF to match or exceed human baselines can be attributed to the superior consistency of AI labelers, which eliminates the noise, fatigue, and inter-annotator disagreement common in crowd-sourced datasets. Furthermore, RLAIF effectively functions as a distillation process, where the policy model aligns with the robust, high-dimensional representations of quality encoded in the superior teacher model, providing a cleaner gradient signal than sparse human labels. Taken together, these results indicate that AI-generated feedback can match the effectiveness of human feedback for certain classes of tasks, while substantially reducing the cost and logistical burden of collecting preference data.

DPO and its variants represent a significant paradigm shift by bypassing the explicit reward modeling stage entirely. Rafailov et al.  \cite{rafailov2023direct} provided empirical evidence that DPO can match or exceed the performance of PPO-based RLHF on tasks like sentiment control and summarization with substantially less complexity. Their results are particularly striking. On the IMDb sentiment generation task, DPO achieved a reward of 0.72, while PPO-RLHF scored 0.53. On the TL;DR summarization dataset, DPO reached a reward of -0.20, compared to PPO-RLHF's -0.26. In both tasks, higher scores indicate better performance. The performance gap in favor of DPO can be largely attributed to the removal of the approximation noise inherent in PPO's actor-critic architecture. While PPO relies on sampling trajectories and estimating value functions (processes prone to high variance and instability), DPO solves the constrained optimization problem analytically. This allows the model to optimize the policy directly against preferences without the middleman of a separate reward model, thereby avoiding the compounding errors that occur when a policy exploits flaws in an imperfect reward proxy. As a result, these empirical gains, together with the elimination of reward modeling and PPO-specific instability, have positioned DPO and its variants as an increasingly attractive choice for large-scale LLM alignment.

Subsequent methods like IPO  \cite{azar2024general} and KTO (Kahneman-Tversky Optimization)  \cite{ethayarajh2024kto} build upon the DPO framework. They offer different loss functions or incorporating different aspects of human preference. For example, KTO uses binary feedback to indicate whether an output is desirable or undesirable. This approach can be more intuitive than pairwise comparisons in certain applications. These methods generally offer improved stability and reduced hyperparameter tuning compared to PPO-based approaches. Therefore, it makes them attractive alternatives for practical deployment.

Beyond alignment with preferences for style and safety, RL is increasingly being applied to improve the multi-step reasoning capabilities of LLMs. Techniques like RLVR  \cite{wang2025reinforcement}, OpenAI’s o1  \cite{jaech2024openai} and o3  \cite{openai2025o3o4}, and DeepSeek-R1  \cite{guo2025deepseek} focus on improving step-by-step reasoning in language models. They do this by assigning rewards based on the correctness of intermediate reasoning steps. These steps are often verified using external tools or smaller, specialized models. Wang et al.  \cite{wang2025reinforcement} showed that RLVR can significantly improve mathematical reasoning performance by reinforcing correct step-by-step logic. With just a single positive example, GPT-3.5’s accuracy on GSM8K increased from 56.8\% to 72.5\%. This pronounced efficacy is rooted in the shift from outcome-sparse supervision to process-dense supervision. By receiving feedback on intermediate steps, the model overcomes the temporal credit assignment problem, learning to distinguish between a correct final answer derived from flawed logic (spurious correlation) and a valid reasoning path derailed by a minor calculation error. This granular reinforcement forces the policy to internalize the causal structure of the solution space, rather than merely relying on surface-level pattern matching. Similarly, DeepSeek-R1  \cite{guo2025deepseek} achieves strong performance on reasoning benchmarks by incorporating automatic rewards based on logical correctness and consistency. These approaches often require careful design of the reward function to accurately reflect reasoning quality and avoid rewarding superficial or incorrect reasoning paths.

A key insight from this comparative analysis is that the choice of RL technique depends heavily on the specific goals, available data, and computational resources. While PPO-based RLHF \cite{ouyang2022training} remains a powerful and widely adopted method, particularly for complex alignment tasks, newer methods like DPO \cite{rafailov2023direct} offer compelling alternatives with better efficiency and stability for certain scenarios.

\subsubsection{Offline and Online Alignment Benchmarks}
\label{sec:alignment-benchmarks}

\textcolor{black}{We first compare offline and online alignment methods, where the central question is how preference-optimization objectives affect broad benchmark performance. These results complement the method-level taxonomy above by showing how supervised-regression variants, direct preference objectives, and PPO-style online updates behave under standardized leaderboard evaluations.}

\begin{table*}[htbp]
\centering
\caption{Performance of Offline Alignment Methods (DPO, KTO, UNA variants) on New Open LLM Leaderboard}
\small
\begin{tabular}{lccccccc}
\toprule
\textbf{Method} & \textbf{BBH} & \textbf{GPQA} & \textbf{MMLU-Pro} & \textbf{MUSR} & \textbf{IFEval} & \textbf{MATH-Hard} & \textbf{Average} \\
\midrule
Mistral (Baseline) & 44.11 & 29.53 & 30.11 & 41.79 & 23.22 & 2.92 & 28.61 \\
DPO (UNA-pairwise) & 44.50 & 28.48 & 30.41 & 39.25 & 26.30 & 2.25 & 28.53 \\
KTO & 44.46 & 29.51 & 30.43 & 40.45 & 24.18 & 2.34 & 28.56 \\
UNA-binary (MSE) & 44.32 & 29.86 & 30.54 & 39.11 & 26.10 & 3.32 & 28.88 \\
UNA-binary (BCE) & 44.43 & 29.42 & 30.73 & 39.51 & 26.49 & 2.99 & 28.93 \\
UNA-score (MSE) & 43.53 & 30.25 & 29.72 & 42.01 & 37.25 & 2.77 & \textbf{30.92} \\
\bottomrule
\end{tabular}
\label{tab:offline_new_leaderboard}
\end{table*}

\begin{table*}[htbp]
\centering
\caption{Performance of Offline Alignment Methods (DPO, KTO, UNA variants) on Old Open LLM Leaderboard}
\label{tab:offline_old_leaderboard}
\small
\begin{tabular}{lccccccc}
\toprule
\textbf{Method} & \textbf{GSM8K} & \textbf{TruthfulQA} & \textbf{Winograde} & \textbf{ARC} & \textbf{HellaSwag} & \textbf{MMLU} & \textbf{Average} \\
\midrule
Mistral (Baseline) & 38.02 & 42.58 & 77.58 & 61.43 & 83.44 & 62.51 & 60.93 \\
DPO (UNA-pairwise) & 40.22 & 44.75 & 79.16 & 62.88 & 84.42 & 62.15 & 62.26 \\
KTO & 41.63 & 47.72 & 78.14 & 62.29 & 84.21 & 62.46 & 62.74 \\
UNA-binary (MSE) & 40.87 & 48.23 & 79.48 & 63.23 & 84.57 & 62.34 & 63.12 \\
UNA-binary (BCE) & 40.41 & 48.33 & 79.40 & 63.14 & 84.60 & 62.48 & 63.06 \\
UNA-score (MSE) & 40.41 & 55.09 & 80.27 & 63.23 & 84.52 & 62.56 & \textbf{64.35} \\
\bottomrule
\end{tabular}
\end{table*}

Tables \ref{tab:offline_new_leaderboard} and \ref{tab:offline_old_leaderboard} present a comprehensive performance comparison of various offline alignment methods, including DPO, KTO, and different UNA (Unified Alignment) variants. UNA \cite{wang2024unifying} is a unified alignment framework that demonstrates RLHF with PPO, DPO, and KTO all optimize the same generalized implicit reward function. This perspective allows these methods to be cast as a single supervised objective capable of handling pairwise, binary, and scalar feedback. As a result, it simplifies and stabilizes the policy fine-tuning process. There are three variants of UNA (Unified Alignment). Each of them correspond to a different type of feedback signal used during policy optimization. UNA-pairwise leverages pairwise preference feedback and is implemented using a Direct Preference Optimization (DPO-style) objective. UNA-binary uses binary reward signals that indicate whether a response is acceptable or not. It is trained with either MSE or Binary Cross Entropy (BCE) loss functions. Finally, UNA-score utilizes scalar reward feedback, such as numerical scores from a reward model or verifier, and optimizes the policy using Mean Squared Error loss. These variants demonstrate the flexibility of UNA in unifying alignment methods under a single supervised learning framework.

The alignment methods are evaluated against a Mistral \cite{jiang2023mistral7b} baseline model. This is because the original Mistral 7B model that is trained purely with next-token prediction and instruction SFT but \emph{does not employ any preference-based or reinforcement-learning alignment}. Table \ref{tab:offline_new_leaderboard} presents the performance comparison on the New Open LLM Leaderboard, while Table \ref{tab:offline_old_leaderboard} reports results on the Old Open LLM Leaderboard. The key distinction between the two is that the new leaderboard incorporates significantly more challenging and diverse benchmarks and therefore provides a more rigorous evaluation of model alignment performance.

Table \ref{tab:offline_new_leaderboard} focusing on the new HuggingFace Open LLM Leaderboard includes challenging tasks such as Big-Bench Hard (BBH), Grade-School Physics Questions Annotated (GPQA), MMLU-Pro, Multi-turn Summarization and Reasoning (MUSR), Instruction Following Evaluation (IFEval), and MATH-Hard. The results reveal that the UNA-score (MSE) method achieves the highest average score of 30.92, substantially outperforming the baseline Mistral model (28.61), as well as other approaches like DPO (28.53) and KTO (28.56). This performance advantage is likely driven by the fidelity of the supervision signal found in the regression objective. While DPO and KTO compress complex quality assessments into relative rankings or binary labels, effectively quantizing the learning signal, UNA-score’s use of Mean Squared Error preserves the magnitude of the error. This allows the optimizer to differentiate between minor stylistic deviations and catastrophic failures, providing a richer gradient signal that is particularly crucial for the nuanced constraint satisfaction required by the new leaderboard's rigorous instruction-following benchmarks.


A particularly striking result is UNA-score (MSE)'s performance on the IFEval benchmark, where it achieves a score of 37.25 compared to the baseline's score of 23.22. It is a remarkable improvement of 60.4\%. This proficiency is likely due to the continuous nature of the MSE loss, which provides a dense feedback signal for multi-constraint satisfaction. Unlike pairwise ranking, which might view two imperfect responses as equally bad, a scalar score can distinguish between a response that meets two out of three constraints versus one that meets none, effectively guiding the model through the optimization landscape of complex formatting rules. Interestingly, while UNA-score (MSE) shows the best overall performance, it actually underperforms the baseline on BBH (43.53 vs. 44.11) and MMLU-Pro (29.72 vs. 30.11). The degradation in these reasoning-heavy benchmarks illustrates the concept of alignment tax. As the model's finite capacity is aggressively optimized to prioritize stylistic adherence and safety constraints, its probability distribution drifts away from the original pre-training manifold, causing a regression in the raw logical deduction and knowledge retrieval capabilities inherent in the base model.

Similarly, Table \ref{tab:offline_old_leaderboard} examining performance on the old Open LLM Leaderboard includes benchmarks such as GSM8K (mathematical reasoning), TruthfulQA (factual accuracy), Winograde (commonsense reasoning), ARC (science knowledge), HellaSwag (commonsense inference), and MMLU (multitask knowledge). Here again, UNA-score (MSE) demonstrates superior overall performance with an average score of 64.35, compared to the Mistral baseline's 60.93, DPO's 62.26, and KTO's 62.74. The consistent superiority of the scalar regression approach over ranking-based (DPO) and binary (KTO) methods can be attributed to the higher information density of the supervision signal. While ranking losses only enforce relative order and ignore whether a response is marginally better or vastly superior, MSE regression forces the model to internalize the absolute magnitude of quality. This creates a more calibrated value landscape, allowing the policy to generalize more effectively across heterogeneous domains like common sense reasoning and scientific knowledge.

The most dramatic improvement is observed on TruthfulQA, where UNA-score (MSE) achieves a score of 55.09. This represents a substantial 29.4\% gain over the baseline score of 42.58. This suggests that the UNA-score approach is particularly effective at enhancing factual accuracy and reducing hallucinations. This advantage likely stems from the fact that MSE functions as a stronger regularizer against confident hallucinations than ranking losses. While a preference-based objective is satisfied merely by ranking the truth higher than the falsehood, potentially with a negligible probability difference, the regression objective instead demands that the model’s internal confidence explicitly match the high scalar value of the truth. This effectively suppresses the model's tendency to generate plausible-sounding but factually incorrect text (mimetic falsehoods) that often permeates the pre-training data. UNA-score (MSE) also shows notable improvements on Winograde (80.27 vs. baseline's 77.58) and ARC (63.23 vs. baseline's 61.43), indicating enhanced reasoning and knowledge capabilities.

These results highlight the strong potential of UNA, particularly the score-based variant trained with MSE loss. It significantly improves LLM performance across diverse evaluation metrics in offline alignment settings. The consistent outperformance across diverse benchmarks suggests that this approach may offer a more robust and generalizable alignment strategy compared to other methods.

\begin{table*}[htbp]
\centering
\caption{Performance of Online Alignment Methods (RLHF vs. UNA) on New Open LLM Leaderboard}
\label{tab:online_new_leaderboard}
\small
\begin{tabular}{lccccccc}
\toprule
\textbf{Method} & \textbf{BBH} & \textbf{GPQA} & \textbf{MMLU-Pro} & \textbf{MUSR} & \textbf{IFEval} & \textbf{MATH-Hard} & \textbf{Average} \\
\midrule
Mistral-INST (Baseline) & 42.46 & 29.05 & 24.53 & 38.30 & 38.46 & 2.02 & 29.14 \\
RLHF & 42.50 & 28.99 & 24.60 & 38.29 & 38.53 & 1.79 & 29.12 \\
UNA (Online) & \textbf{42.78} & 28.32 & \textbf{24.87} & 38.03 & \textbf{39.17} & 1.75 & \textbf{29.15} \\
\bottomrule
\end{tabular}
\end{table*}

\begin{table*}[htbp]
\centering
\caption{Performance of Online Alignment Methods (RLHF vs. UNA) on Old Open LLM Leaderboard}
\label{tab:online_old_leaderboard}
\small
\begin{tabular}{lccccccc}
\toprule
\textbf{Method} & \textbf{GSM8K} & \textbf{TruthfulQA} & \textbf{Winograde} & \textbf{ARC} & \textbf{HellaSwag} & \textbf{MMLU} & \textbf{Average} \\
\midrule
Mistral-INST (Baseline) & 35.14 & 55.94 & 73.72 & 55.29 & 75.99 & 53.94 & 58.34 \\
RLHF & 34.42 & 55.88 & 73.56 & 55.20 & 76.03 & 54.03 & 58.19 \\
UNA (Online) & \textbf{35.67} & 55.88 & \textbf{74.03} & 55.20 & \textbf{76.61} & 54.02 & \textbf{58.57} \\
\bottomrule
\end{tabular}
\end{table*}

Tables \ref{tab:online_new_leaderboard} and \ref{tab:online_old_leaderboard} shift the focus to online alignment methods. We compare traditional PPO-based RLHF with the online variant of UNA, using a Mistral-INST \cite{jiang2023mistral7b} model as the baseline. Mistral-INST serves as the baseline because it is an instruction-tuned version of the Mistral 7B model that has not been further aligned using preference-based or reinforcement learning methods. The tables \ref{tab:online_new_leaderboard} and \ref{tab:online_old_leaderboard} offer insight into how different alignment techniques perform in an online learning setup, where the policy is continuously refined with new feedback.

Table \ref{tab:online_new_leaderboard} examines performance on the new Open LLM Leaderboard. In this setting, the online UNA method attains a marginally higher average score of 29.15, edging out RLHF which has a score of 29.12 and the Mistral-INST baseline which has a score 29.14. While the overall improvement is modest, online UNA demonstrates notable gains on certain benchmarks. It achieves 42.78 on BBH (vs. RLHF's 42.50), 24.87 on MMLU-Pro (vs. RLHF's 24.60), and 39.17 on IFEval (vs. RLHF's 38.53). These results suggest that UNA may be more effective at enhancing instruction-following capabilities (IFEval) and certain types of reasoning tasks (BBH, MMLU-Pro). The distinct advantage on these metrics can be attributed to the variance reduction inherent in UNA's loss function. While standard PPO updates are notoriously noisy due to the stochastic nature of advantage estimation and trajectory sampling, UNA casts the online update as a supervised regression problem. This stabilizes the gradient descent trajectory, allowing the model to converge on subtle instruction-following nuances without the forgetting or instability often introduced by the high-variance updates of traditional reinforcement learning.

Similarly, Table \ref{tab:online_old_leaderboard} compares performance on the old Open LLM Leaderboard. Here again, online UNA marginally outperforms RLHF with an average score of 58.57 compared to RLHF's score of 58.19 and the baseline's score of 58.34. Online UNA demonstrates better performance on GSM8K (35.67 vs. RLHF's 34.42), Winograde (74.03 vs. RLHF's 73.56), and HellaSwag (76.61 vs. RLHF's 76.03). The improvement on GSM8K is particularly noteworthy, as it suggests enhanced mathematical reasoning capabilities. Therefore, mathematically intense tasks typically benefit from the exploration inherent in online learning, allowing the model to self-correct reasoning paths rather than merely imitating a static dataset. Furthermore, the specific superiority of UNA over standard PPO in this domain is likely linked to the limitations of the latter's trust-region constraints. When the model serendipitously discovers a novel, high-reward reasoning chain during exploration, the probability ratio between the new and old policies often diverges significantly. PPO's clipping mechanism can dampen the learning signal from these outlier successes to maintain stability. In contrast, UNA's objective function allows for more aggressive assimilation of these rare, high-value trajectories, thereby accelerating the acquisition of complex logical patterns.

While the performance improvements in these online learning experiments are modest, the UNA framework offers potential advantages beyond raw performance metrics. As discussed by Wang et al.  \cite{wang2024unifying}, UNA can simplify the RLHF pipeline by transforming it into a supervised learning problem. Such a formulation can reduce memory consumption and shorten training time. This operational efficiency, combined with competitive performance, makes UNA a promising approach for online alignment scenarios.

\subsubsection{Model-Level Comparisons Across Reasoning, Coding, Truthfulness, and Instruction Following}
\label{sec:model-level-comparison}

\textcolor{black}{We next move from method-specific comparisons to model-level comparisons. This view is useful because deployed LLMs often combine multiple post-training stages, making it difficult to attribute performance to a single algorithm in isolation. The following tables therefore compare representative open-weight models by their reported RL technique and benchmark behavior across reasoning, coding, truthfulness, and instruction-following settings.}

\begin{table*}[htbp]
\centering
\caption{Performance of prominent open-weight LLMs on core reasoning/knowledge tasks, with their post-training RL alignment method}
\label{tab:rl_reasoning}
\small
\begin{tabular}{lcccc}
\toprule
\textbf{Model (size)} & \textbf{Reinforcement Learning Technique} & \textbf{MMLU} & \textbf{GSM8K} & \textbf{ARC-Ch.} \\
\midrule
Llama 3.1 70B Instruct & RLHF (PPO + DPO) & 79.5 & 89.1 & 93.0 \\
Llama 3.1 405B Instruct & RLHF (PPO) & 85.2 & 96.4 & 95.3 \\
Mixtral 8×22B Instruct & DPO & 77.8 & 74.1 & 70.5 \\
DeepSeek-V2 & GRPO & 86.4 & 84.0 & 92.4 \\
Qwen2-72B Instruct & RLHF (PPO / RLAIF) & 84.0 & 88.3 & 71.6 \\
Gemma 2 27B Instruct & RLHF (PPO) & 75.2 & 74.0 & 71.4 \\
Phi-3-mini-4k Instruct & Break-Fix RL & 70.9 & 85.7 & 86.3 \\
\bottomrule
\end{tabular}
\end{table*}

Table \ref{tab:rl_reasoning} provides a comprehensive overview of how prominent open-weight LLMs perform on core reasoning and knowledge tasks, along with the specific RL techniques employed in their training. The models, along with their respective reinforcement learning techniques, are evaluated on three key benchmarks: 5-shot MMLU, Maj@8 GSM8K, and 25-shot ARC-Challenge. Here, MMLU, GSM8K, and ARC-Challenge are datasets, and 5-shot, Maj@8, and 25-shot refer to the evaluation protocols applied to them. The 5-shot evaluation protocol means that the model is given five example question–answer pairs (shots) as context before answering each test question. Similarly, 25-shot ARC-Challenge presents the model with 25 such examples. For GSM8K, Maj@8 refers to majority voting over eight sampled completions, where the final answer is chosen as the most frequently occurring prediction among those eight.

Table \ref{tab:rl_reasoning} reveals several interesting patterns. Llama 3.1 405B Instruct, aligned with traditional RLHF (PPO), demonstrates exceptional performance across all three benchmarks. It achieves 85.2\% on MMLU, 96.4\% on GSM8K, and 95.3\% on ARC-Challenge. This suggests that traditional RLHF, when applied to very large models, can yield outstanding results. The smaller Llama 3.1 70B Instruct, which uses a combination of RLHF (PPO) and DPO, also performs admirably. It scores 79.5\% on MMLU, 89.1\% on GSM8K, and 93.0\% on ARC-Challenge. The exceptional performance of the 405B model with pure PPO highlights a scaling phenomenon in reinforcement learning. Larger models possess more resilient representations, enabling them to endure the high variance of online RL updates without experiencing the policy collapse that often afflicts smaller architectures. In contrast, the success of the hybrid approach on the 70B model suggests that moderately sized models benefit from a two-stage regime. PPO is essential for discovering the complex reasoning chains required for math tasks (GSM8K), while the subsequent DPO stage is necessary to anneal these behaviors, preventing the overfitting and reward hacking that typically degrade general knowledge scores (MMLU).

DeepSeek-V2 \cite{liu2024deepseekv2}, which employs GRPO, achieves the highest MMLU score of 86.4\% among all models listed. This suggests that this RL technique may be particularly effective for enhancing knowledge-intensive capabilities. However, its GSM8K performance of 84.0\% is lower than both Llama models. This divergence in performance profiles can be traced to the architectural distinction of GRPO: the removal of the critic model. By estimating the baseline from the group mean rather than a learned value function, GRPO eliminates the approximation noise and memory overhead of the critic, thereby minimizing the alignment tax that typically degrades factual knowledge (MMLU) during RL fine-tuning. However, the absence of a parametric value function may be a liability for complex reasoning tasks like GSM8K. Without a learned critic to reduce variance by predicting the expected return of specific states, the gradient signals for long, multi-step reasoning chains become noisier, making it harder to reinforce precise logical sequences.

Interestingly, Phi-3-mini-4k Instruct \cite{abdin2024phi}, despite being the smallest model in the table with just 3.8 billion parameters, achieves impressive results on GSM8K with a score of 85.7\% and on ARC-Challenge with a score of 86.3\% using the Break-Fix RL approach. Break-Fix RL is a reinforcement learning approach where the model is penalized for harmful or incorrect behaviors (“breaks”) and rewarded for corrected or safe behaviors (“fixes”), enabling iterative safety and reliability improvements. The outsized performance of this compact model is likely driven by the high information content of the \textit{break-fix} paradigm. Unlike standard RL, which often relies on scalar rewards for whole trajectories, this approach explicitly models the transition from an incorrect state to a correct one. This effectively functions as a form of contrastive training on hard negatives, allowing the model to focus its limited capacity on learning the precise error-correction mechanisms required for robust reasoning, rather than wasting parameters on broad, undirected exploration. These results demonstrate that specialized alignment techniques can enable smaller models to compete effectively on targeted reasoning and problem-solving tasks.

Mixtral 8×22B Instruct \cite{jiang2024mixtral}, aligned solely with DPO, shows relatively lower performance compared to PPO-based models, with scores of 77.8\% on MMLU, 74.1\% on GSM8K, and 70.5\% on ARC-Challenge. The reason for this comparative underperformance likely stems from the structural limitation of purely offline alignment for cognitive tasks. Reasoning benchmarks often require the model to generalize logic to novel scenarios. This capability is best fostered by the active exploration inherent in PPO, where the model generates and reinforces its own successful trajectories. In contrast, DPO is mathematically constrained to the static support of the preference dataset, effectively penalizing deviation from the provided demonstrations even if those deviations represent valid, alternative reasoning paths.

Qwen2-72B Instruct \cite{yang2024qwen2}, which uses a combination of RLHF (PPO) and RLAIF, performs strongly on MMLU with a score of 84.0\% and on GSM8K with 88.3\%. However, its performance on ARC-Challenge is comparatively lower at 71.6\%. This divergence is likely due to the susceptibility of RLAIF to the capabilities of the teacher model. MMLU and GSM8K represent domains where strong teacher models excel, allowing RLAIF to effectively distill this competence into the student via dense feedback. Conversely, ARC-Challenge consists of problems specifically designed to break standard LLM heuristics. If the AI labeler itself struggles with these adversarial common-sense reasoning tasks, the resulting reward signal becomes noisy, preventing the policy from learning the nuanced physical intuition required to solve them.

Overall, this table illustrates the complex relationship between model size, alignment technique, and performance across different reasoning tasks. It shows that although larger models generally perform better, the choice of alignment method can significantly influence the specific strengths and weaknesses of a model.

\begin{table*}[htbp]
\centering
\caption{Performance of popular open-weight code-capable LLMs on standard coding benchmarks, with their RL Technique}
\label{tab:rl_coding}
\footnotesize
\begin{tabular}{lccc}
\toprule
\textbf{Model (size)} & \textbf{Reinforcement Learning Technique} & \textbf{HumanEval (pass@1)} & \textbf{MBPP (pass@1)} \\
\midrule
Llama 3.3 70B Versatile & RLHF (PPO + DPO) & 88.4 & 87.6 \\
Qwen2-72B Instruct & RLHF (PPO / RLAIF) & 86.0 & 80.2 \\
WizardCoder-Python-34B V1.1 & Evol-Instruct (RLEIF) & 79.9 & 78.9 \\
DeepSeek-Coder-V2 & GRPO & 57.3 & 45.8 \\
StarCoder2-15B Instruct V0.1 & DPO (SelfCodeAlign) & 72.6 & 75.2 \\
Gemma 2 27B Instruct & RLHF (PPO) & 51.8 & 62.6 \\
Mixtral 8×22B Instruct & DPO & 76.2 & 64.3 \\
\bottomrule
\end{tabular}
\end{table*}

Table \ref{tab:rl_coding} shifts the focus to coding capabilities, presenting the performance of popular open-weight code-capable LLMs on two standard benchmarks: HumanEval  \cite{chen2021evaluating} and MBPP (Mostly Basic Python Problems)  \cite{austin2021program}. The metric reported is pass@1, which measures the percentage of problems correctly solved by the model on the first attempt. This table provides valuable insights into how different reinforcement learning techniques influence coding proficiency and generalization across programming tasks.

Llama 3.3 70B Versatile \cite{meta2024llama3_3}, which employs a combination of PPO-based RLHF and DPO, demonstrates exceptional coding capabilities, achieving 88.4\% pass@1 on HumanEval and 87.6\% on MBPP. Qwen2-72B Instruct \cite{yang2024qwen2}, using RLHF (PPO/RLAIF), also performs admirably with 86.0\% on HumanEval and 80.2\% on MBPP. The synergy observed in the Llama model likely results from the complementary nature of the two stages. PPO excels at exploring the combinatorial search space to find functionally correct logic paths by maximizing unit test passing. The subsequent DPO phase refines stylistic attributes such as readability and variable naming, where human preferences are most distinct. Furthermore, the success of RLAIF in Qwen2 highlights the unique amenability of programming to automated oversight. Unlike subjective creative writing, code quality is largely objective, allowing AI supervisors to generate high-fidelity preference labels with minimal noise, thereby scaling the alignment process without diluting signal quality.

WizardCoder-Python-34B V1.1 is trained using Reinforcement Learning from Evol‑Instruct Feedback (RLEIF) \cite{luo-etal-2024-wizardcoder}, a technique that combines evolutionary instruction refinement with reward-based learning to improve coding performance. This method enables the model to iteratively learn from enhanced prompts, reinforcing effective code generation strategies. As a result, WizardCoder-Python-34B V1.1 achieves strong performance, scoring 79.9\% on HumanEval and 78.9\% on MBPP. The disproportionate effectiveness of this technique relative to model scale can be ascribed to the synthetic difficulty injection provided by the Evol-Instruct framework. Standard code corpora often suffer from a simplicity bias, containing a preponderance of trivial or repetitive problems. By mutating these into increasingly convoluted and constrained variants, RLEIF forces the policy to transcend rote memorization and develop deeper, more abstract syntactic generalizations to satisfy the evolved complexity, thereby extracting greater reasoning density from fewer parameters.

In contrast, DeepSeek-Coder-V2 \cite{zhu2024deepseek}, which employs GRPO, shows relatively lower performance with 57.3\% on HumanEval and 45.8\% on MBPP. This is particularly interesting given that DeepSeek-V2 \cite{liu2024deepseekv2} performed very well on reasoning tasks (as seen in Table \ref{tab:rl_reasoning}), suggesting that GRPO may be more effective for general reasoning than for specialized coding tasks. This discrepancy highlights a critical vulnerability of group-based baselines in domains with all-or-nothing reward landscapes. Unlike multiple-choice reasoning (MMLU) where random guessing provides a baseline of correct answers within a batch, complex coding tasks often lead to scenarios where every sampled candidate fails the unit tests. In these all-failure regimes, the intra-group variance collapses to zero, effectively silencing the relative gradient signal and stalling optimization, whereas methods backed by a learned value function (like PPO) can still derive a learning signal from the global expectation of failure.

StarCoder2-15B Instruct V0.1, which uses DPO with SelfCodeAlign \cite{wei-etal-2024-selfcodealign}, achieves respectable results with 72.6\% on HumanEval and 75.2\% on MBPP, despite being the smallest model in the table. SelfCodeAlign is a fully transparent, self-alignment pipeline that enhances code models without human labels by having the model generate and validate its own instruction–response pairs using test suites. This approach succeeds because it exploits the deterministic nature of code execution to solve the noise accumulation problem common in self-training. While self-generated data in natural language can lead to model collapse by reinforcing hallucinations, the use of unit tests as an objective oracle allows SelfCodeAlign to strictly prune incorrect solutions. This ensures that the DPO phase optimizes against a dataset of mathematically verified correctness, providing a gradient signal that is far purer and more consistent than subjective human preferences.

Gemma 2 27B Instruct \cite{team2024gemma}, aligned with RLHF using PPO, scores 51.8\% on HumanEval and 62.6\% on MBPP, suggesting that the model may struggle with certain types of coding problems. Similarly, Mixtral 8×22B Instruct \cite{jiang2024mixtral}, aligned with DPO, achieves 76.2\% on HumanEval and 64.3\% on MBPP, indicating a performance gap in the opposite direction. This contrast in performance profiles highlights the distinct regularization effects of the two algorithms. Gemma’s underperformance on HumanEval under PPO is likely due to alignment tax, where the KL-divergence penalties and broad safety objectives inherent in general-purpose RLHF constrain the model from exploring the precise, often non-conversational syntactic paths required for complex coding tasks. Conversely, Mixtral’s bias toward HumanEval while dropping significantly on MBPP is symptomatic of DPO’s susceptibility to dataset-specific overfitting. Lacking the exploratory noise of online RL, the policy tightly adheres to the distribution of the preference data and struggles to generalize to the unseen problem formulations in MBPP.

These results collectively illustrate that coding proficiency is influenced by both model architecture and alignment technique. Models specifically designed for code generation (like WizardCoder and StarCoder) can achieve competitive performance even at smaller scales when aligned with appropriate techniques. Table \ref{tab:rl_coding} also suggests that hybrid approaches combining multiple alignment methods (like PPO+DPO) may offer advantages for enhancing coding capabilities.

\begin{table*}[htbp]
\centering
\caption{Truthfulness performance of large open-weight LLMs on the TruthfulQA benchmark, with their RL Technique}
\label{tab:rl_truthfulness}
\small
\begin{tabular}{lcc}
\toprule
\textbf{Model (size)} & \textbf{RL Technique} & \textbf{TruthfulQA 0-shot (\%)} \\
\midrule
Meta Llama 3.1 70B Instruct & RLHF (PPO + DPO) & 62.9 \\ 
Qwen2-72B Instruct & RLHF (PPO) + RLAIF & 67.0 \\ 
Mixtral 8×22B Instruct & Direct Preference Optimisation (DPO) & 51.1 \\ 
DeepSeek-V2 Chat (RL) & Group Relative Policy Optimisation (GRPO) & 57.7 \\ 
Gemma 2 27B Instruction-tuned & RLHF (PPO) & 51.6 \\ 
Phi-3-mini-4k Instruct & "Break-Fix" safety RL cycle & 38.5 \\ 
TÜLU 3 70B & Trust-Region DPO + RL-VR & 63.8 \\ 
\bottomrule
\end{tabular}
\end{table*}

Table \ref{tab:rl_truthfulness} focuses specifically on truthfulness alignment, presenting the performance of large open-weight LLMs on the TruthfulQA \cite{lin-etal-2022-truthfulqa} benchmark. This benchmark evaluates a model’s ability to avoid generating misleading or factually incorrect statements. The reported scores use a 0-shot evaluation setting, meaning the models respond to each prompt without any task-specific examples or prior fine-tuning, thereby assessing their default ability to generate truthful outputs.

The results reveal significant variations in truthfulness across different models and alignment techniques. Qwen2-72B Instruct \cite{yang2024qwen2}, which employs a combination of RLHF (PPO) and RLAIF, achieves the highest TruthfulQA score of 67.0\%. This suggests that incorporating AI feedback alongside human feedback may be particularly effective for enhancing factual accuracy and reducing hallucinations. The strong performance of RLAIF in this context may be due to the ability of AI systems to systematically verify factual claims against large knowledge bases, a concept known as \textit{Scalable Oversight} where AI tools assist in grading tasks too complex or tedious for unaided humans.

T\"ulu 3-70B model \cite{lambert2025tulu} uses a two-stage alignment process. First, the Llama-3.1-70B base model is aligned using length-normalized DPO. This is a KL-constrained method that keeps the new policy close to the reference model. Then, the DPO-aligned model is further refined using RLVR. RLVR gives positive rewards only when task-specific verifiers approve the output. It achieves the second-highest score of 63.8\%. The multi-stage alignment approach used for this model appears to be effective for enhancing truthfulness, potentially because the verifiable rewards component directly incentivizes factual accuracy.

Meta Llama 3.1 70B Instruct, aligned with a combination of RLHF (PPO) and DPO, achieves a respectable score of 62.9\%. This hybrid approach seems to strike a good balance between instruction following and maintaining factual accuracy. This robust performance is likely due to DPO acting as a regularizer against the reward hacking often induced by PPO. While PPO encourages the model to maximize a reward score, potentially by generating plausible-sounding but fabricated details to appear more helpful. The subsequent DPO stage effectively re-grounds the policy. By strictly optimizing against the preference dataset in the final phase, the model is pulled back toward the empirical distribution of truthful answers. This suppresses the tendency to confabulate that can emerge from the unconstrained exploration of value-based reinforcement learning.

DeepSeek-V2 Chat (RL) \cite{liu2024deepseekv2}, which employs Group Relative Policy Optimisation (GRPO), achieves a moderate score of 57.7\%. While this is lower than the top-performing models, it still represents a substantial improvement over completely unaligned models (which typically score much lower on TruthfulQA). This intermediate ranking highlights a potential blind spot in relative optimization schemes when dealing with factual integrity. Because GRPO derives its baseline solely from the average of the sampled group, it lacks an absolute grounding signal. In challenging query regimes where all generated candidates might be factually flawed (common in zero-shot hallucination triggers), the algorithm is forced to reinforce the least incorrect hallucination rather than suppressing the behavior entirely. Unlike PPO or RLVR, which utilize a critic or verifier to assign low absolute scores to plausible lies, GRPO optimizes for relative plausibility, potentially entrenching mimetic falsehoods if they appear marginally superior to their peers.

Mixtral 8×22B Instruct \cite{jiang2024mixtral} and Gemma 2 27B Instruction-tuned \cite{team2024gemma} achieve similar scores of 51.1\% and 51.6\% respectively, despite using different alignment techniques (DPO vs. RLHF with PPO). The similarity in performance implies that, for truthfulness, the specific implementation details and training data may be as important as the choice of alignment algorithm. The parity in these results underscores a shared pathology in general-purpose alignment known as \textit{sycophancy}. Whether minimizing a negative log-likelihood loss (DPO) or maximizing a scalar reward (PPO), both methods are ultimately tethered to the quality of the underlying human preference data. Standard preference datasets often prioritize helpfulness and conversational fluidity over rigorous fact-checking, effectively training models to validate user misconceptions rather than correct them. Consequently, regardless of the optimization mechanics, both models converge on a local optimum of agreeable plausibility, where the policy learns to mimic the common errors found in the training distribution rather than challenging the false premises embedded in the TruthfulQA prompts.

Phi-3-mini-4k Instruct \cite{abdin2024phi}, which uses the ``Break-Fix" safety RL cycle, achieves the lowest score of 38.5\%. This reduced efficacy is likely a byproduct of the over-refusal pathology common in safety-first alignment. The Break-Fix algorithm optimizes heavily against bad behaviors defined by safety guidelines. Consequently, the model learns a decision boundary that is overly conservative. When confronted with the mimetic traps of TruthfulQA, the compact model lacks the semantic resolution to distinguish between a malicious prompt and a tricky factual query, causing it to default to safe, non-informative abstentions which are penalized in the scoring metric, rather than risking the nuance required for a truthful rebuttal.

These results highlight the complex relationship between alignment techniques and truthfulness. While all alignment methods aim to improve model behavior, their effectiveness for enhancing factual accuracy varies considerably. The table suggests that hybrid approaches combining multiple alignment methods, particularly those incorporating AI feedback or verifiable rewards, may be most effective for enhancing truthfulness.

\begin{table*}[htbp]
\centering
\caption{Instruction-following performance of popular open-weight LLMs on standard dialogue benchmarks, with their RL Technique}
\label{tab:rl_instruction}
\footnotesize
\begin{tabular}{lccc}
\toprule
\textbf{Model (size)} & \textbf{RL Technique} & \textbf{AlpacaEval 2.0 LC Win Rate (\%)} & \textbf{MT-Bench (0–10)} \\
\midrule
Meta Llama 3.1 70B Instruct & RLHF (PPO + DPO) & 34.4 & 8.15 \\ 
Qwen2-72B Instruct & RLHF (PPO / RLAIF) & 36.6 & 9.10 \\ 
Mixtral 8×22B Instruct & DPO & 30.9 & 8.66 \\ 
DeepSeek-V2 Chat (RL) & GRPO & 38.9 & 8.97 \\ 
Gemma 2 27B Instruction-tuned & RLHF (PPO) & 57.5 & 8.62 \\ 
Phi-3-mini-4k Instruct & Break-Fix RL & 23.1 & 8.12 \\ 
T\"ULU 3 70B & Trust-Region DPO + RL-VR & 49.8 & 8.60 \\ 
\bottomrule
\end{tabular}
\end{table*}

Table \ref{tab:rl_instruction} evaluates instruction-following quality across two widely used benchmarks: AlpacaEval 2.0 and MT-Bench. AlpacaEval 2.0 \cite{alpaca_eval} evaluates a model’s single-turn helpfulness by comparing its responses using GPT-4 as an automated judge. The scores are expressed as win rates, where higher values reflect better performance. MT-Bench \cite{zheng2023judging} assesses multi-turn conversational ability, assigning a score between 0 and 10 based on criteria like coherence, helpfulness, and consistency. Together, these benchmarks provide a comprehensive view of how different RL alignment techniques influence a model’s capacity to follow instructions and sustain effective dialogue.

The results reveal some surprising patterns. Gemma 2 27B Instruction-tuned \cite{team2024gemma}, which uses traditional PPO-based RLHF, achieves the highest AlpacaEval 2.0 win rate of 57.5\%, substantially outperforming larger models. This disproportionate success is likely a manifestation of proxy optimization inherent to online RL methods. AlpacaEval relies on an automated LLM judge (typically GPT-4) to determine win rates. Because PPO is an active maximization process, it allows the policy to aggressively hill-climb the reward landscape. If the reward model used during training serves as a high-fidelity proxy for the evaluation judge, PPO enables the model to discover and exploit specific stylistic nuances. These can include preferred structural templates or tonal patterns that statistically maximize the judge’s approval. This can allow the model to outperform larger models that are constrained by the static and potentially less-optimized distributions of offline DPO datasets.

T\"ULU 3 70B  \cite{lambert2025tulu}, with its multi-stage alignment approach (Trust-Region DPO followed by RL-VR), achieves the second-highest AlpacaEval win rate of 49.8\%. The efficacy of this tiered strategy can be attributed to the decoupling of stylistic alignment from verifiable reasoning. The initial Trust-Region DPO phase acts as a stable anchor, instilling the preferred conversational tone and formatting while rigorously penalizing the KL-divergence to prevent the policy collapse often observed in unconstrained DPO. The subsequent RL-VR (Reinforcement Learning with Verifiable Rewards) phase then optimizes the model against ground-truth correctness rather than human preference proxies. This ensures that the final policy produces responses that are not only superficially fluent (pleasing the judge's stylistic bias) but also substantively robust, avoiding the vacuous verbosity trap where models generate long but empty content to game the evaluation metrics.

DeepSeek-V2 Chat (RL) \cite{liu2024deepseekv2}, which employs GRPO, achieves a respectable AlpacaEval win rate of 38.9\% and the second-highest MT-Bench score of 8.97. This divergence between a moderate win rate and a top-tier benchmark score underscores the conservative nature of the GRPO update rule. MT-Bench evaluates sustained, multi-turn capabilities across diverse categories (reasoning, roleplay, coding), requiring the preservation of the base model's deep semantic structures. GRPO, by removing the parametric critic and reducing variance through group averaging, minimizes the alignment tax that often degrades these generalist abilities during fine-tuning. Conversely, AlpacaEval is highly sensitive to stylistic optimization (such as verbosity and formatting). The fact that GRPO scores lower here than aggressive PPO-based models suggests it is less prone to reward hacking, preferring to maintain robust core competency rather than overfitting to the superficial preferences of a single-turn auto-evaluator.

Qwen2-72B Instruct \cite{yang2024qwen2}, which uses a combination of PPO-based RLHF and RLAIF, achieves the highest MT-Bench score of 9.10 and a solid AlpacaEval win rate of 36.6\%. The supremacy of this model on the multi-turn MT-Bench metric can be ascribed to the superior \textit{semantic resolution} of AI-generated feedback. While human annotators frequently succumb to fatigue or cognitive bias when evaluating long, complex dialogue chains, an AI supervisor (RLAIF) provides consistent, granular critique across the entire conversation history. This dense reward signal allows the PPO algorithm to optimize for deep coherence and logical continuity, traits essential for multi-turn interactions. It moves beyond merely perfecting the single-turn stylistic flourishes that often dominate the AlpacaEval leaderboard.

Meta Llama 3.1 70B Instruct, aligned with a combination of PPO-based RLHF and DPO, achieves a moderate AlpacaEval win rate of 34.4\% and an MT-Bench score of 8.15. While these scores are respectable, they are lower than might be expected given the model's strong performance on reasoning tasks (as seen in Table \ref{tab:rl_reasoning}). This incongruence between reasoning prowess and dialogue metrics is frequently indicative of the alignment tax associated with rigorous safety tuning. Llama models are characterized by conservative refusal boundaries. While this minimizes harmful outputs, it often triggers false positive refusals on benign but ambiguous prompts, severely penalizing the model on open-ended chat benchmarks. Furthermore, the optimization for precise, concise reasoning (beneficial for GSM8K) often conflicts with the verbosity bias inherent in AlpacaEval judges, where longer, more decorative responses are statistically preferred over shorter, factually dense ones.

Mixtral 8×22B Instruct \cite{jiang2024mixtral}, aligned solely with DPO, achieves an AlpacaEval win rate of 30.9\% and an MT-Bench score of 8.66. The relatively high MT-Bench score suggests that DPO may be effective for enhancing conversational capabilities, even if it doesn't match the instruction-following performance of more complex alignment approaches. This characteristic split reflects competent multi-turn dialogue alongside average win-rates. It is emblematic of the mode-seeking behavior of DPO. Unlike PPO, which allows the policy to drift significantly from its initialization to find high-reward "utliers (often exploiting the judge's bias for length or structure), DPO is mathematically anchored to the support of the offline dataset. It excels at smoothing out the base model's rough edges to produce consistent, high-quality chat (MT-Bench), but it lacks the active exploration mechanism required to discover the specific, often artificial, stylistic peaks that maximize relative win-rates in a competitive pairwise evaluation.

Phi-3-mini-4k Instruct \cite{abdin2024phi}, which employs the Break-Fix RL approach, records the lowest AlpacaEval win rate at 23.1\% but achieves a respectable MT-Bench score of 8.12. This bifurcation in performance metrics serves as a stark illustration of the trade-off between semantic validity and stylistic elaboration in small-scale models. The respectable MT-Bench score indicates that the "Break-Fix" cycle effectively instilled core instruction-following logic and safety adherence. However, the limited parameter count constrains the model's capacity for the creative embellishment and extended context handling often rewarded by AlpacaEval. Consequently, while the model functions correctly as a logical engine, it lacks the generative bandwidth to produce the verbose, highly nuanced prose necessary to secure wins against larger opponents in subjective preference rankings.

These results highlight the complex relationship between alignment techniques and instruction-following capabilities. Different benchmarks may capture different aspects of instruction-following, and models may excel in some areas while underperforming in others. Finally, Table \ref{tab:rl_instruction} suggests that sophisticated multi-stage alignment approaches and hybrid methods incorporating multiple feedback sources may offer advantages for enhancing instruction-following capabilities.

\subsection{Synthesis of Comparative Trends}
\label{sec:comparison-synthesis}

The comparative analysis presented underscores several key insights into how RL techniques shape the performance of LLMs across diverse tasks. Firstly, the unified alignment framework (UNA), particularly its score-based variant trained with MSE, consistently demonstrates robust improvements across multiple benchmarks in offline scenarios. This approach notably enhances factual accuracy (TruthfulQA) and instruction-following capabilities (IFEval), outperforming traditional baselines like DPO and KTO. In online alignment settings, UNA maintains competitive performance with traditional PPO-based RLHF, delivering incremental improvements in reasoning benchmarks such as GSM8K and BBH. In short, based on the comparative study, UNA seems to be an attractive alignment method for practical applications requiring real-time updates and resource efficiency.

Secondly, the effectiveness of specific RL methods varies considerably based on the targeted task and model size. This highlights the nuanced interplay between alignment strategies and desired capabilities. For reasoning and knowledge-intensive tasks, PPO-based RLHF and hybrid methods combining PPO with DPO consistently achieve strong performance, particularly in larger models like Llama 3.1 and Qwen2-72B \cite{yang2024qwen2}. For specialized coding tasks, tailored alignment approaches like evolutionary instruction feedback (RLEIF) and self-alignment techniques (SelfCodeAlign) offer notable advantages. These methods enable even relatively smaller models to perform competitively by reinforcing domain-specific competencies. Instruction-following evaluations further reveal that complex multi-stage alignment methods, such as Trust-Region DPO followed by RL-VR, deliver well-rounded improvements across conversational benchmarks. Notably, specialized alignment methods like Break-Fix RL can empower smaller models to achieve impressive conversational performance, underscoring the importance of alignment strategy selection based on model constraints and application goals. Collectively, these findings emphasize that no single RL technique universally dominates, and optimal performance across diverse tasks typically emerges from thoughtful alignment of model architecture, RL technique, and specific operational objectives.

\section{Challenges and Limitations}
Despite the significant progress in applying RL to LLMs, several critical challenges and limitations persist, which hinder the full realization of their potential. These can be broadly categorized into research bottlenecks, technical limitations, and overarching challenges in deployment and evaluation. Addressing these issues is paramount for the continued advancement and responsible application of RL-enhanced LLMs.

Current research bottlenecks primarily revolve around the scalability and quality of feedback, and the complexity of reward modeling. While RLHF is effective, its dependence on human feedback makes it expensive and time-consuming. This reliance also poses scalability challenges, especially when aligning models across a broad range of behaviors and nuanced tasks. For RLAIF, as previously discussed, Sharma et al. \cite{sharma2024critical} highlighted the concern that AI-generated feedback can inherit or even amplify biases from the supervising model. This may lead to the emergence of behaviors that, while appearing aligned, ultimately diverge from genuine human values. Furthermore, Denison et al. \cite{denison2024sycophancy} and Fu et al. \cite{fu2025reward} demonstrated that designing reward models capable of faithfully capturing complex human preferences across diverse contexts is inherently challenging. These models are often vulnerable to reward hacking, a phenomenon where language models learn to exploit the reward function to maximize scores without genuinely completing the intended task. Such vulnerabilities highlight the difficulty of aligning models through reward-based methods alone, as even small imperfections in the reward specification can lead to unintended behaviors. The development of more sophisticated reward modeling techniques, robust evaluation metrics for alignment, and efficient methods for eliciting and aggregating diverse human (or AI) preferences are active areas of research crucial for overcoming these bottlenecks.

From a technical standpoint, applying reinforcement learning to large language models comes with several inherent limitations. One major challenge is the substantial computational cost involved in training these models, especially when using on-policy algorithms like PPO. Such training demands extensive hardware resources and long runtimes. As a result, the process can be prohibitively expensive and difficult to access for many research groups and organizations. Sample efficiency is another major concern. RL algorithms often require a vast number of interactions or feedback instances to learn effectively, which is exacerbated by the high dimensionality of the action space (i.e., text generation) in LLMs. Moreover, the stability of reinforcement learning training can be difficult to maintain. Models may sometimes suffer from catastrophic forgetting, where previously learned capabilities are lost during further training. In other cases, they can experience policy collapse, leading to a sharp decline in generation quality. Ensuring stable and efficient training, along with the development of more sample-efficient RL algorithms tailored to language tasks, remains a major technical hurdle. Equally important is the creation of robust policy update techniques to enhance the practicality and reliability of RL for LLMs.

\begingroup

\subsection{Practical Training and Systems Considerations}
\label{sec:practical-systems}

Beyond algorithmic performance, the practical adoption of RL-based post-training depends on training complexity, hyperparameter sensitivity, and system-level memory cost. These considerations are especially important because standard supervised fine-tuning (SFT) and RL-based post-training differ not only in objective function, but also in the number of active models, the need for rollout generation, the presence of reward or verifier calls, and the size of the tuning surface. To make these costs explicit, let \(N\) denote the number of parameters in the trainable policy model, \(L_p\) the prompt length, \(L_r\) the generated response length, \(L=L_p+L_r\), \(B\) the number of prompts per batch, \(K\) the number of sampled responses per prompt, \(E\) the number of PPO or policy-optimization epochs per rollout batch, and \(C_{\mathrm{fwd}}(N,L)\) the cost of one forward pass through the model. For dense decoder-only Transformers, a common approximation is that one forward-backward training pass over \(D\) tokens costs on the order of \(6ND\) floating-point operations. Equivalently, a forward-backward step over a batch of \(BL\) tokens scales as \(\mathcal{O}(6NBL)\) up to architecture- and hardware-dependent constants~ \cite{kaplan2020scaling,hoffmann2022training}.

Table~\ref{tab:training_complexity_systems} summarizes the relative training complexity of major post-training pipelines. SFT has the simplest structure. It optimizes a dense token-level loss over fixed demonstrations and therefore requires no online sampling, reward inference, or policy-rollback mechanism. Direct preference methods such as DPO, KTO, and ORPO remain close to this supervised regime because they operate on offline preference pairs or labeled responses \cite{rafailov2023direct}. In contrast, PPO-style RLHF introduces a substantially more complex loop: the policy must generate responses, a reward model must score them, a value model must estimate advantages, and the policy is then updated over multiple epochs under a KL constraint \cite{ouyang2022training}. The authors of \cite{chakraborty2023parl} demonstrated that online RLHF is modeled as bi-level RL, and this approach is further studied in \cite{gaur2025sample,li2026oracle,wu2026selfimproving}. GRPO removes the learned critic but still requires multiple sampled completions per prompt \cite{shao2024deepseekmath}, while RLVR and program-synthesis RL replace learned rewards with external verifiers whose runtime may dominate training cost when execution is slow, or test suites are large \cite{lambert2025tulu}.

\begin{table*}[htbp]
\centering
\caption{Training complexity of major LLM post-training pipelines. The expressions are schematic and emphasize the dominant operations relative to SFT.}
\label{tab:training_complexity_systems}
\scriptsize
\renewcommand{\arraystretch}{1.15}
\begin{tabularx}{\textwidth}{
@{}
>{\RaggedRight\arraybackslash}p{0.17\textwidth}
>{\RaggedRight\arraybackslash}p{0.21\textwidth}
>{\RaggedRight\arraybackslash}p{0.25\textwidth}
>{\RaggedRight\arraybackslash}X
@{}
}
\toprule
\textbf{Pipeline} 
& \textbf{Dominant training loop} 
& \textbf{Approximate per-batch cost} 
& \textbf{Practical interpretation} \\
\midrule

\textbf{SFT} 
& One forward-backward pass on demonstrations 
& \(\mathcal{O}(6NBL)\) 
& Lowest-complexity baseline: dense token-level supervision, no rollout generation, no reward model, and no policy-stability constraint beyond standard optimization. \\

\textbf{DPO / KTO / ORPO} 
& Forward passes on chosen/rejected or labeled responses, followed by supervised preference loss 
& \(\mathcal{O}(6NBL)\) for the trainable policy, plus optional frozen-reference forward cost 
& Similar to SFT in implementation complexity; avoids online rollouts, learned reward-model inference, and critic training, but remains limited by the coverage and quality of the offline preference data~ \cite{rafailov2023direct}. \\

\textbf{PPO-based RLHF} 
& Generate rollouts, score with reward model, estimate advantages with value model, then run \(E\) policy updates 
& Rollout/scoring cost \(+\ \mathcal{O}(E \cdot 6N B(L_p+L_r))\), plus reward/reference/value-model passes 
& Highest training complexity among standard alignment methods: the loop couples generation, reward inference, value fitting, KL control, and repeated policy optimization~ \cite{ouyang2022training,yao2023deepspeed}. \\

\textbf{RLAIF / Constitutional AI} 
& RLHF-style loop, but feedback is generated by an AI evaluator, critique model, or constitutional judge 
& PPO-style cost \(+\) evaluator or critique-generation cost 
& Reduces human-labeling cost but adds evaluator inference and prompt/rubric engineering; system complexity depends heavily on the evaluator architecture~ \cite{lee2023rlaif,bai2022constitutional}. \\

\textbf{GRPO} 
& Sample \(K\) responses per prompt, compute group-normalized rewards, update policy without a learned critic 
& \(\mathcal{O}(K \cdot B \cdot C_{\mathrm{fwd}}(N,L_r)) + \mathcal{O}(6N B K L)\) 
& Removes the value-model bottleneck of PPO, but converts part of the cost into generation throughput because each prompt requires multiple completions~ \cite{shao2024deepseekmath}. \\

\textbf{RLVR / Program-Synthesis RL} 
& Generate candidates and evaluate them using deterministic verifiers, compilers, unit tests, solvers, or formal checkers 
& \(\mathcal{O}(K \cdot B \cdot C_{\mathrm{fwd}}(N,L_r)) + \mathcal{O}(K B C_{\mathrm{verify}})\) 
& Avoids learned reward-model bias, but training cost depends on verifier latency, test-suite size, execution timeouts, and the sparsity of successful samples~ \cite{lambert2025tulu}. \\
\bottomrule
\end{tabularx}
\end{table*}

The second practical distinction concerns hyperparameter tuning. In SFT, the dominant hyperparameters are the familiar supervised-learning choices: learning rate, batch size, sequence length, number of epochs, and regularization. RL-based methods add additional parameters that directly govern exploration, policy drift, reward scaling, and variance reduction. Table~\ref{tab:hyperparameter_tuning_systems} highlights this expansion. PPO-style RLHF is especially sensitive because the KL coefficient, clipping threshold, reward normalization, value-loss weight, generation temperature, rollout length, and number of PPO epochs interact with one another. For example, increasing temperature or rollout length can improve exploration but also increases reward variance and memory pressure; weakening the KL penalty can increase reward but also increases the risk of reward hacking or language-quality degradation. GRPO reduces the critic-related tuning burden, but introduces group size \(K\) and group-normalization stability as central choices. RLVR simplifies reward design when a deterministic verifier exists, but shifts tuning pressure toward sampling budget, verifier timeout, and sparse-reward mitigation.

\begin{table*}[htbp]
\centering
\caption{Hyperparameter tuning burden across SFT and RL-based LLM post-training pipelines}
\label{tab:hyperparameter_tuning_systems}
\scriptsize
\renewcommand{\arraystretch}{1.15}
\begin{tabularx}{\textwidth}{
@{}
>{\RaggedRight\arraybackslash}p{0.16\textwidth}
>{\RaggedRight\arraybackslash}p{0.19\textwidth}
>{\RaggedRight\arraybackslash}p{0.23\textwidth}
>{\RaggedRight\arraybackslash}X
@{}
}
\toprule
\textbf{Pipeline} 
& \textbf{Approximate tuning surface} 
& \textbf{Most sensitive hyperparameters} 
& \textbf{Tuning implication} \\
\midrule

\textbf{SFT} 
& Low: typically \(5\)--\(7\) core choices 
& Learning rate, batch size, sequence length, number of epochs, weight decay, warmup ratio 
& Usually stable once the learning rate and batch size are selected; failures are often standard supervised-learning failures such as overfitting or catastrophic forgetting. \\

\textbf{DPO / KTO / ORPO} 
& Moderate: SFT choices plus \(2\)--\(4\) preference-objective choices 
& Preference temperature \(\beta\), margin or odds-ratio coefficient, reference-model strength, length normalization, preference batch construction 
& Easier to tune than PPO because there is no rollout loop or critic, but performance is sensitive to preference-label quality, length bias, and the strength of the reference constraint~ \cite{rafailov2023direct}. \\

\textbf{PPO-based RLHF} 
& High: often \(12+\) interacting choices 
& KL coefficient, PPO clip range, value-loss coefficient, reward scaling/whitening, entropy bonus, rollout batch size, response length, sampling temperature, number of PPO epochs, GAE parameters \((\gamma,\lambda)\) 
& Most tuning-intensive standard pipeline; small changes can alter the balance between reward maximization, policy stability, and language quality. TRL's PPO trainer, for example, exposes defaults such as \( \texttt{kl\_coef}=0.05\), \( \texttt{cliprange}=0.2\), \( \texttt{vf\_coef}=0.1\), \( \gamma=1.0\), and \( \lambda=0.95\)~ \cite{vonwerra2020trl}. \\

\textbf{RLAIF / Constitutional AI} 
& High: PPO/DPO choices plus evaluator-design choices 
& Evaluator model, critique prompt, constitutional rules, preference-generation temperature, judge calibration, plus PPO/DPO parameters 
& Reduces reliance on human annotators but introduces a second tuning layer: the AI feedback generator must itself be calibrated so that its preferences do not amplify systematic bias. \\

\textbf{GRPO} 
& Moderate-to-high: PPO-like policy choices without value-model choices 
& Group size \(K\), reward normalization, KL coefficient, response length, sampling temperature, group-level reward variance 
& Removes value-loss and critic-stability tuning, but requires careful control of group diversity; if all sampled responses receive identical rewards, the normalized advantage can vanish or become unstable~ \cite{shao2024deepseekmath}. \\

\textbf{RLVR / Program-Synthesis RL} 
& Moderate: policy choices plus verifier/execution choices 
& Sampling budget \(K\), verifier timeout, unit-test coverage, pass/fail reward shaping, curriculum difficulty, KL coefficient, response length 
& The reward is objective, but the training process can still be brittle because sparse binary rewards make the useful-signal rate highly dependent on sampling budget and verifier coverage~ \cite{lambert2025tulu}. \\
\bottomrule
\end{tabularx}
\end{table*}

The third systems-level distinction is memory footprint. Table~\ref{tab:gpu_memory_systems} gives a normalized estimate using a full fine-tuning setting with mixed precision and AdamW. The estimate excludes activations, KV cache, communication buffers, ZeRO/FSDP sharding effects, and parameter-efficient adapters; these omissions are deliberate because they vary by implementation. Under the standard mixed-precision AdamW accounting, a trainable model requires approximately \(18N\) bytes for weights, gradients, and optimizer states, before activation memory is added~ \cite{wolf2020transformersmemory}. Therefore, even before considering activations, a 7B-parameter model requires roughly \(18 \times 7 = 126\) GB of parameter-state memory for full SFT. Frozen reference or reward models are cheaper because they do not require gradients or optimizer states; when stored in half precision, their weight-only cost is approximately \(2N\) bytes. These estimates explain why practical RLHF systems rely heavily on model sharding, CPU/NVMe offload, tensor parallelism, vLLM-style generation engines, and careful scheduling across actor, critic, reward, and reference models~ \cite{yao2023deepspeed,hu2024openrlhf}.

\begin{table*}[htbp]
\centering
\caption{Approximate GPU memory footprint relative to full SFT. Estimates use \(18N\) bytes for each trainable AdamW model and \(2N\) bytes for each frozen half-precision model, excluding activations, KV cache, temporary buffers, and sharding/offload. The 7B column gives the corresponding parameter-state memory lower bound.}
\label{tab:gpu_memory_systems}
\scriptsize
\renewcommand{\arraystretch}{1.15}
\begin{tabularx}{\textwidth}{
@{}
>{\RaggedRight\arraybackslash}p{0.16\textwidth}
>{\RaggedRight\arraybackslash}p{0.24\textwidth}
>{\RaggedRight\arraybackslash}p{0.17\textwidth}
>{\RaggedRight\arraybackslash}p{0.18\textwidth}
>{\RaggedRight\arraybackslash}X
@{}
}
\toprule
\textbf{Pipeline} 
& \textbf{Model components kept during training} 
& \textbf{Approx. parameter-state memory} 
& \textbf{7B example} 
& \textbf{System-level implication} \\
\midrule

\textbf{SFT} 
& One trainable policy 
& \(18N\) bytes \((1.0\times)\) 
& \(\approx 126\) GB 
& Baseline full fine-tuning cost before activations; feasible only with sharding/offload for large models unless using PEFT. \\

\textbf{DPO / KTO / ORPO} 
& Trainable policy \(+\) frozen reference model 
& \(18N + 2N = 20N\) bytes \((1.11\times)\) 
& \(\approx 140\) GB 
& Slightly above SFT if the reference model is resident; can be reduced if reference log-probabilities are precomputed or if reference-free variants are used. \\

\textbf{PPO-based RLHF} 
& Trainable policy \(+\) trainable critic/value model \(+\) frozen reward model \(+\) frozen reference model 
& \(18N + 18N + 2N + 2N = 40N\) bytes \((2.22\times)\) 
& \(\approx 280\) GB 
& Memory pressure is substantially higher than SFT because PPO maintains multiple model roles; large-scale systems distribute actor, critic, reward, and reference models across GPUs~ \cite{hu2024openrlhf}. \\

\textbf{RLAIF / Constitutional AI} 
& PPO/DPO-style policy pipeline \(+\) AI evaluator or critique model 
& \(20N\)--\(40N\) bytes \(+\) evaluator memory 
& \(\approx 140\)--\(280\) GB \(+\) evaluator 
& If optimized with PPO, the cost approaches RLHF; if optimized with DPO-like losses, it is closer to preference fine-tuning, but evaluator inference adds additional memory or serving cost. \\

\textbf{GRPO} 
& Trainable policy \(+\) frozen reference model; no learned critic 
& \(18N + 2N = 20N\) bytes \((1.11\times)\) 
& \(\approx 140\) GB 
& Removes the critic memory cost of PPO, but increases generation memory/throughput pressure because \(K\) completions are sampled per prompt~ \cite{shao2024deepseekmath}. \\

\textbf{RLVR / Program-Synthesis RL} 
& Trainable policy \(+\) optional frozen reference model \(+\) external verifier 
& \(18N\)--\(20N\) bytes \((1.0\)--\(1.11\times)\) \(+\ C_{\mathrm{verify}}\) 
& \(\approx 126\)--\(140\) GB \(+\) verifier 
& If the verifier is external, memory can remain close to SFT; however, wall-clock cost can be dominated by execution, compiler calls, solver latency, or unit-test throughput. \\
\bottomrule
\end{tabularx}
\end{table*}

Taken together, Tables~\ref{tab:training_complexity_systems}--\ref{tab:gpu_memory_systems} show that practical post-training choices are governed by a three-way trade-off among optimization flexibility, tuning burden, and system cost. SFT is the least expensive and most stable pipeline, but it can only imitate behaviors present in the supervised data. Direct preference methods preserve much of SFT's simplicity while incorporating preference information, making them attractive when high-quality offline preference data is available. PPO-based RLHF provides the richest online optimization loop, but it introduces the largest hyperparameter surface and memory footprint. GRPO and RLVR represent two different attempts to reduce this burden: GRPO removes the learned critic and replaces absolute value estimation with group-relative normalization, whereas RLVR removes learned reward uncertainty by shifting supervision to deterministic verifiers. Thus, benchmark comparisons should ideally be reported alongside rollout budget, number of active model components, group size, verifier cost, sequence length, and hardware configuration, because these system variables strongly affect whether a method is practical beyond its final score.
\endgroup

Beyond specific research and technical hurdles, broader challenges exist in the evaluation, safety, and ethical deployment of RL-aligned LLMs. Evaluating the true alignment and safety of large language models remains a formidable challenge, as existing benchmarks often fail to capture the full spectrum of failure modes and adversarial behaviors. This concern has been underscored by Abeysinghe et al. \cite{abeysinghe2024challenges} and Lee et al. \cite{lee-etal-2025-evaluating}, who emphasize the limitations of current evaluation frameworks in reliably measuring model robustness and safety. Ensuring that models are not only helpful and harmless on average but also robust against misuse, manipulation, or the generation of subtle misinformation is an ongoing struggle. The problem of “alignment faking” or sycophancy, as discussed by Wang et al. \cite{wang-etal-2024-fake} and Greenblatt et al. \cite{greenblatt2024alignment}, adds another layer of complexity to evaluation. In such cases, models may outwardly appear aligned while concealing underlying misaligned behaviors. Ethical concerns in alignment involve the values instilled during training and the potential for biased feedback to produce inequitable models. These risks raise serious questions about fairness, inclusivity, and unintended harm. Additionally, the broader societal impact of deploying powerful, RL-aligned LLMs calls for robust governance frameworks and oversight. These multifaceted challenges underscore the need for interdisciplinary collaboration and a continued focus on building trustworthy and beneficial AI systems. However, even with robust governance, a fundamental epistemological gap remains in how we measure success itself.

\textbf{Evaluation Crisis:} Perhaps the most significant open gap facing the field is the \textit{Proxy-Objective Mismatch}. Current optimization frameworks often maximize a reward score (the Proxy) under the implicit assumption that it remains strictly correlated with human intent (Alignment) or logical validity (Reasoning).  However, as models scale in capacity and optimization pressure increases, they can enter a Goodharting regime. \textcolor{black}{Reward-model overoptimization provides a concrete instance of this effect. Continued optimization against a learned proxy can keep increasing proxy reward even after held-out gold reward plateaus or deteriorates~ \cite{gao2023scaling}. A related safety failure appears as exaggerated refusal, where models reject benign prompts that superficially resemble unsafe requests~ \cite{rottger2024xstest}.} In this regime, they optimize the proxy metric to extreme levels while decoupling from the underlying objective. In alignment tasks, this manifests as models generating verbose, confident, but vacuous responses to satisfy length-biased reward models. Simultaneously, in reasoning domains, this mismatch creates shortcut learning, where agents exploit spurious correlations, such as memorizing specific solution templates or gaming the coverage gaps of a unit test, to maximize outcome-based rewards without developing robust, generalizable logic. Unlike simple overfitting, this represents a structural failure of measurement. The metric itself ceases to be a valid indicator of quality once it becomes the target of optimization. This crisis suggests that static, scalar reward functions are insufficient for both super-human alignment and rigorous problem-solving. Bridging this gap requires a paradigm shift toward dynamic and interactive evaluation frameworks. Examples include verifier-in-the-loop training or recursive oversight that can adaptively validate both semantic integrity and logical soundness against ground truth.

\section{Emerging Trends and Future Directions}
The field of RL for LLMs is rapidly advancing, with several emerging trends poised to shape its future trajectory. One significant trend is the shift toward more advanced and efficient RL algorithms beyond PPO. This includes the adoption of offline RL methods, which reduce the need for costly online data collection. Another development is the integration of alignment techniques more closely with the LLM architecture, such as DPO and its variants  \cite{rafailov2023direct, wang2024offline, snell2206offline}. These approaches aim to reduce the computational burden and sample complexity associated with traditional RLHF, making alignment more accessible and efficient. Another key direction is the increasing sophistication of AI-driven feedback (RLAIF) and self-improvement mechanisms, where models learn to critique and refine their own outputs or learn from other AI systems. Furthermore, there is an increasing emphasis on enhancing the reasoning capabilities of LLMs through RL, moving beyond simple preference alignment to instill complex, multi-step problem-solving skills, often involving verifiable rewards or process-based supervision. This includes developing RL techniques that can explicitly train models to generate coherent intermediate reasoning steps, crucial for tasks requiring deep logical inference and planning.

Looking ahead, future research is likely to focus on more robust and interpretable alignment techniques. One direction is moving beyond black-box reward models. Researchers aim to understand what values and preferences are being learned. They also want to know how these influence model behavior. Another area of interest is multi-objective reinforcement learning. This helps balance conflicting goals like helpfulness, harmlessness, honesty, and fairness. Personalized alignment is also a growing focus. Here, LLMs would adapt to individual user preferences in a safe and controlled way. The integration of reinforcement learning with other machine learning paradigms presents exciting avenues. Causal inference can help better understand model behavior. Unsupervised and self-supervised methods may help discover reward signals without human labels. As LLMs gain the ability to interact with external tools and environments, multi-agent RL will become increasingly important. These frameworks can enable training for collaboration and social awareness. Ultimately, the grand challenge is to build LLMs that are not only powerful but also safe and aligned with human values. This will require ongoing advances in RL techniques, evaluation strategies, and ethical safeguards.

\section{Conclusion}
This survey provides a comprehensive exploration of reinforcement learning techniques for large language models. It shows how RL has grown from a simple fine-tuning method to a central approach in LLM development. The field has progressed from RLHF to advanced methods like RLAIF, DPO, and GRPO. Each technique brings its own strengths and limitations. Our analysis highlights key trade-offs between alignment methods. Traditional PPO-based RLHF performs exceptionally well, especially with very large models. In contrast, newer approaches like UNA-score excel at improving factual accuracy and instruction-following. RL techniques also show strong potential in boosting reasoning abilities. Methods like OB-RL, CoT-RO, Verifier-Guided RL, and RLVR help improve multi-step reasoning and logical consistency. Despite these advancements, several challenges still remain. These include reward hacking, high computational costs, limited scalability of collecting high-quality feedback, and the risk of AI feedback systems reinforcing their own biases. Looking forward, several promising research directions emerge. These include developing more efficient algorithms and creating hybrid approaches that combine the strengths of different methods. Researchers are also exploring multi-objective RL to balance competing goals. Integrating RL with other learning paradigms is another important path. Advancing hierarchical RL methods for tool use and external resource integration is equally vital. Together, these efforts aim to create language models that are more helpful, harmless, and honest in serving human needs.

\section{Authors' Perspective}

In synthesising the extensive literature and empirical results surveyed in this work, we diverge from the prevailing trend that views alignment primarily as a unified optimization problem solvable by simplified, offline objectives. Instead, we argue that the field is bifurcating into two distinct modalities, which are \textit{Instruction Adherence} and \textit{Reasoning Search}. They demand fundamentally different reinforcement learning paradigms.

A dominant view in recent research suggests that implicit alignment methods, such as DPO  \cite{rafailov2023direct} and KTO  \cite{ethayarajh2024kto}, will supersede traditional actor-critic architectures (like PPO) due to their stability and computational efficiency. We disagree with this view regarding reasoning-intensive tasks. While our analysis confirms that DPO is superior for style transfer, tone, and safety constraints (Instruction Adherence), it fundamentally lacks the active exploration mechanism required for logic discovery. Implicit methods maximize the likelihood of preferred data already present in the reference distribution. However, complex reasoning often requires the model to traverse low-probability valleys. These are intermediate steps that seem statistically unlikely but are logically necessary to reach a correct solution. We posit that on-policy exploration, as found in PPO  \cite{schulman2017proximal} or GRPO  \cite{guo2025deepseek}, remains indispensable for domains like mathematics and coding. In these areas, the model must not merely mimic human preferences but actively discover novel execution paths that may not exist in the supervised training set.

We also challenge the framing of the Alignment Tax as the degradation of reasoning capabilities during safety training and argue that it is not an inevitable cost of doing business. We view it instead as a symptom of objective mismatch. Current alignment pipelines typically apply a monolithic reward model (mixing safety, style, and helpfulness) to all queries. Our stance is that the prevailing "one-size-fits-all" alignment strategy is flawed. We advocate for a Dynamic Compute paradigm, similar to the architectures seen in reasoning-specialized models  \cite{jaech2024openai, guo2025deepseek}. In this view, alignment for safety should be treated as a constraint satisfaction problem (via supervised regression or DPO), while reasoning should be treated as a tree-search optimization problem (via outcome-based RL or RLVR). The conflation of these two objectives into a single scalar reward function is, in our view, the primary driver of performance degradation in generalist models.

Finally, we agree with the growing skepticism regarding static benchmarks (e.g., AlpacaEval, MT-Bench). Our analysis of \textit{Goodhart's Law} in Section 7 suggests that the correlation between leaderboard performance and human utility is breaking down. We observe that models are increasingly optimizing for the \textit{proxy} (length, formatting, confident tone) rather than the \textit{intent}. We posit that the field must pivot from static evaluation sets to Verifier-Guided Dynamic Evaluation. Future effective RL research will likely rely not on fixed test sets, but on interactive environments where the model is evaluated on its ability to satisfy a functional verifier (e.g., a compiler, a formal proof checker, or a game engine) rather than a static text similarity metric. Without this shift, RL algorithms risk optimizing for sophistry and sounding correct rather than optimizing for truth.

\bibliography{reference}
\bibliographystyle{unsrt}

\end{document}